%% file: sample-manuscript.tex
\documentclass[acmsmall, review=false, screen=true]{acmart}
\renewcommand\footnotetextcopyrightpermission[1]{} 
\usepackage{amsmath}
\usepackage{paralist}
\usepackage{siunitx}
\usepackage{subcaption}
\usepackage{xcolor}
\usepackage{soul}
\usepackage{enumitem}
\graphicspath{{Figures/}}
\usepackage{multirow}
\usepackage{balance}

\usepackage{caption}
\usepackage{subcaption}
\usepackage{cleveref}

\usepackage{epstopdf}
\usepackage{etoolbox}
\usepackage{tikz}
\newcolumntype{C}{>{\centering\arraybackslash}p{9em}}
\newcolumntype{E}{>{\centering\arraybackslash}p{
7em}}
\newcolumntype{D}{>{\centering\arraybackslash}p{3em}}
\newrobustcmd*{\mysquare}[1]{\tikz{\filldraw[draw=#1,fill=#1] (0,0)
rectangle (0.2cm,0.2cm);}}

\newrobustcmd*{\mycircle}[1]{\tikz{\filldraw[draw=#1,fill=#1] (0,0) circle [radius=0.1cm];}}

\newrobustcmd*{\mytriangle}[1]{\tikz{\filldraw[draw=#1,fill=#1] (0,0) --
(0.2cm,0) -- (0.1cm,0.2cm);}}

\usepackage{hyperref}
\AtBeginDocument{%
  \providecommand\BibTeX{{%
    \normalfont B\kern-0.5em{\scshape i\kern-0.25em b}\kern-0.8em\TeX}}}

\acmConference[]{}{}{}
\begin{document}
\title{AIST: An Interpretable Attention-based Deep learning Model for Crime Prediction}

\author{Yeasir Rayhan}
\affiliation{
  \institution{Bangladesh University of Engineering and Technology}
  \country{Bangladesh}}
\email{yeasirrayhan.prince@gmail.com}

\author{Tanzima Hashem}
\affiliation{
  \institution{Bangladesh University of Engineering and Technology}
  \country{Bangladesh}
}
\email{tanzimahashem@cse.buet.ac.bd}


\begin{abstract}
  Accuracy and interpretability are two essential properties for a crime prediction model. Because of the adverse effects that the crimes can have on human life, economy and safety, we need a model that can predict future occurrence of crime as accurately as possible so that early steps can be taken to avoid the crime. On the other hand, an interpretable model reveals the reason behind a model's prediction, ensures its transparency and allows us to plan the crime prevention steps accordingly. The key challenge in developing the model is to capture the non-linear spatial dependency and temporal patterns of a specific crime category while keeping the underlying structure of the model interpretable. In this paper, we develop AIST, an Attention-based Interpretable Spatio Temporal Network for crime prediction. AIST models the dynamic spatio-temporal correlations for a crime category based on past crime occurrences, external features (e.g., traffic flow and point of interest (POI) information) and recurring trends of crime. Extensive experiments show the superiority of our model in terms of both accuracy and interpretability using real datasets. 
\end{abstract}



\maketitle
\pagestyle{plain}

\section{Introduction}
\label{sec:intro}
Criminal activities have become a major social problem due to their adverse effect on human life, economy and safety. The availability of crime data in recent years has enabled researchers to develop models for crime prediction. The government and responsible authorities can take preventive measures if they know about a crime event in advance. Knowing the insight behind the prediction of a crime occurrence would allow them to plan preventive measures appropriately and keep the society safe from the happening of the crime. Interpretable predictions ensure the transparency and accountability of the model. Thus, both accuracy and interpretability are two essential and desired properties for a crime prediction model. We propose an \emph{\underline{A}ttention-based \underline{I}nterpretable \underline{S}patio \underline{T}emporal Network (AIST)}, an interpretable deep learning model for crime prediction. 

\begin{figure}[htbp]
	\begin{subfigure}{0.32\columnwidth}
		\def\svgwidth{\columnwidth}
	    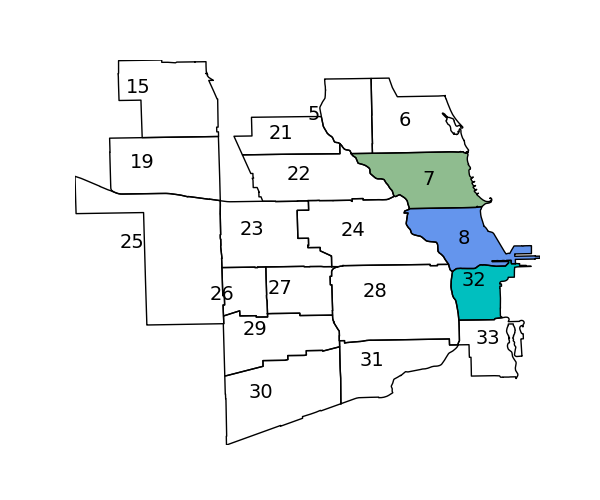
		\caption{Chicago Communities}
		\label{Fig:chicago_map}
	\end{subfigure}
	\begin{subfigure}{0.32\columnwidth}
	\def\svgwidth{\columnwidth}
	    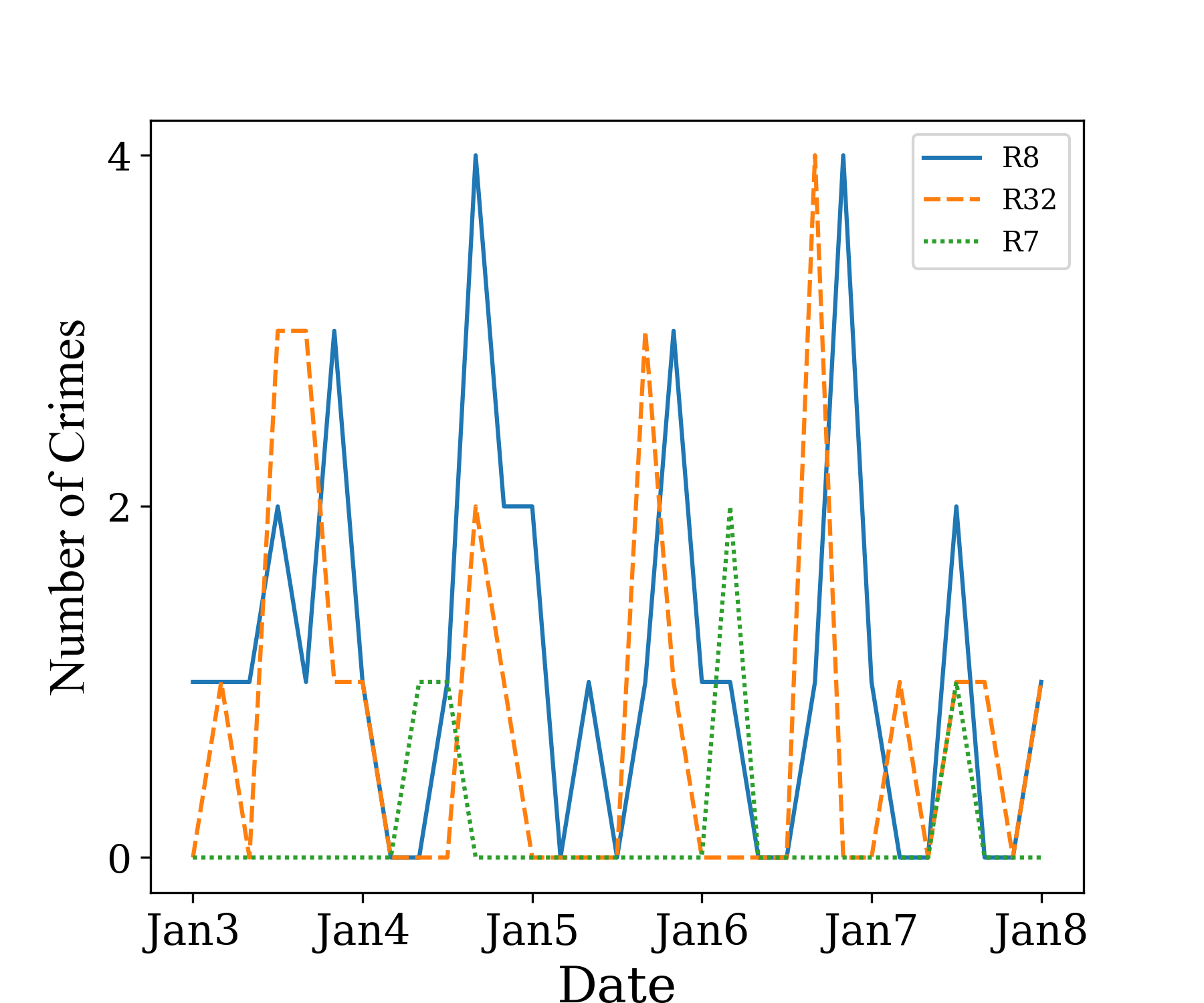
		\caption{Spatial correlation}
		\label{Fig:sp_cor}
	\end{subfigure}
	\begin{subfigure}{0.32\columnwidth}
	 \def\svgwidth{\columnwidth}
	    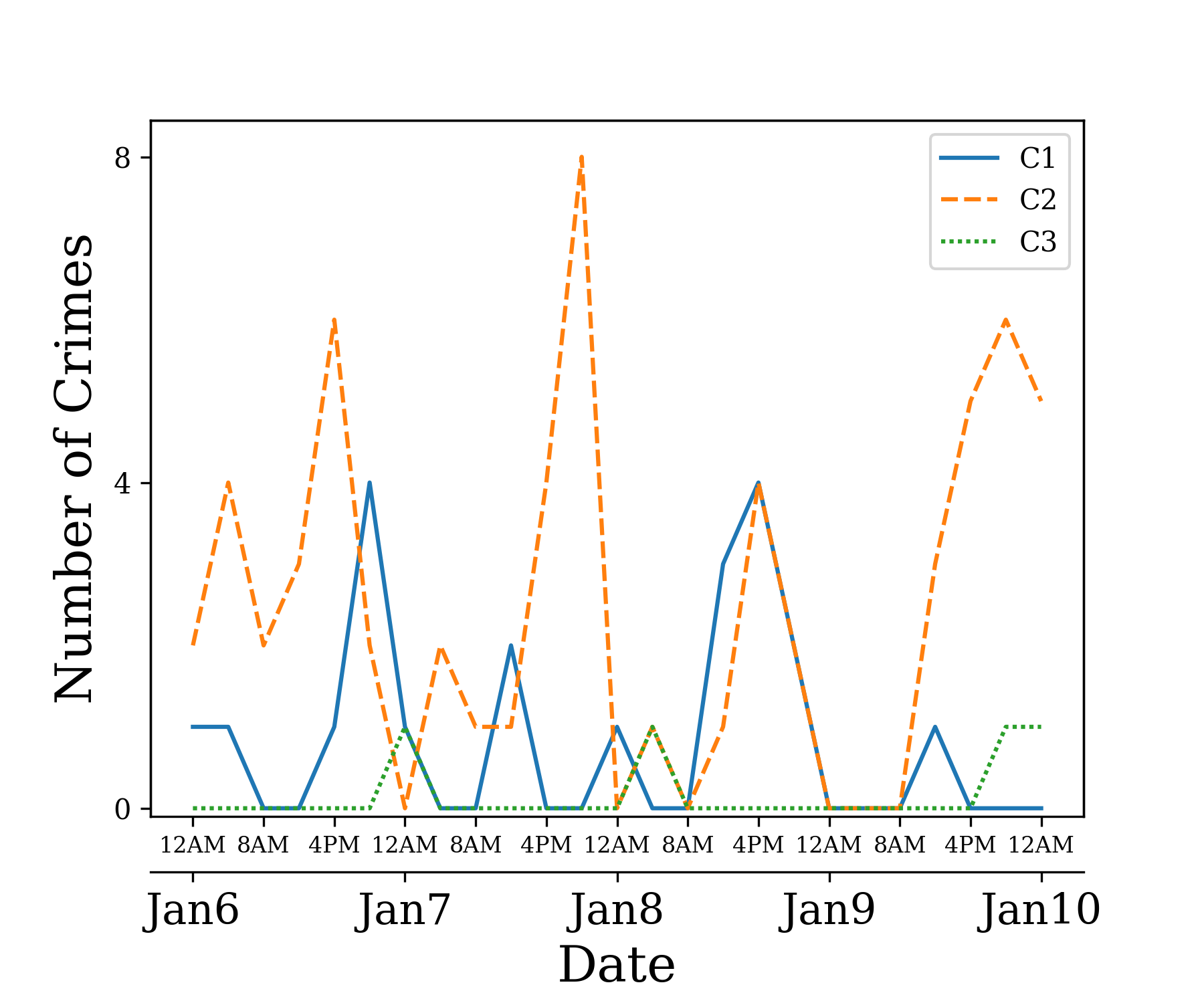
		\caption{Temporal correlation}
		\label{Fig:tem_cor}
	\end{subfigure}
	\caption{Spatio-temporal dependencies of crime distribution}
	\label{Fig:map_sp_tem_cor}
    \vspace{-1em}
\end{figure}

Crime events exhibit spatial and temporal correlations and external features (e.g., taxi flow) often have influence on the crime occurrence.

\textbf{Spatial Correlation.} Spatially nearby regions show a similar crime distribution and the extent of this similarity varies across regions and time. Figure~\ref{Fig:chicago_map} shows the communities (i.e., regions) of Chicago and Figure~\ref{Fig:sp_cor} shows an example on January, 2019 Chicago crime data. Regions 8 and 32 show strong spatial correlation while Regions 8 and 7 do not, though both of them are spatially nearby. Also, the spatial correlation between Regions 8 and 32 changes with time. 

\textbf{Temporal Correlation.} Crime occurrences of a region show both short and long term temporal correlations and these correlations vary with crime categories. Fig~\ref{Fig:tem_cor} shows an example for Region 8: deceptive practice (C0) and theft (C1) peak during mid night, whereas robbery (C5) peak during late night or early morning. There is also a significant difference of crime distributions across different crime categories: deceptive practice (C0) and theft (C1) occur at regular intervals whereas robbery is not so common for Region $8$. Besides, the crime distributions of the same category differ throughout the week. 

\vspace{-1em}
\begin{figure}[htbp]
	\begin{subfigure}{0.32\columnwidth}
	    \def\svgwidth{\columnwidth}
	    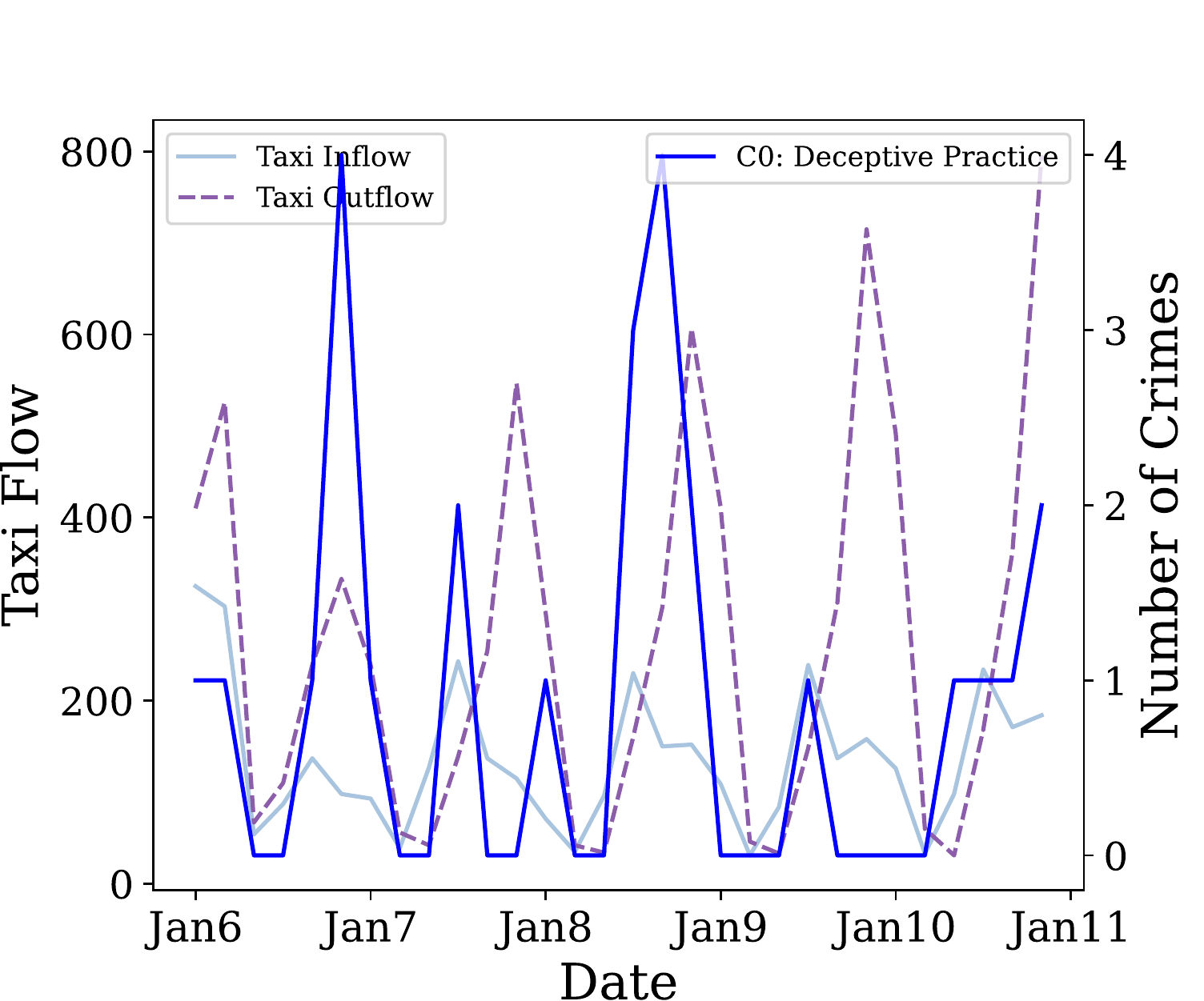
		\caption{}
		\label{Fig:ext_cor_1}
	\end{subfigure}
	\begin{subfigure}{0.32\columnwidth}
	 \def\svgwidth{\columnwidth}
	    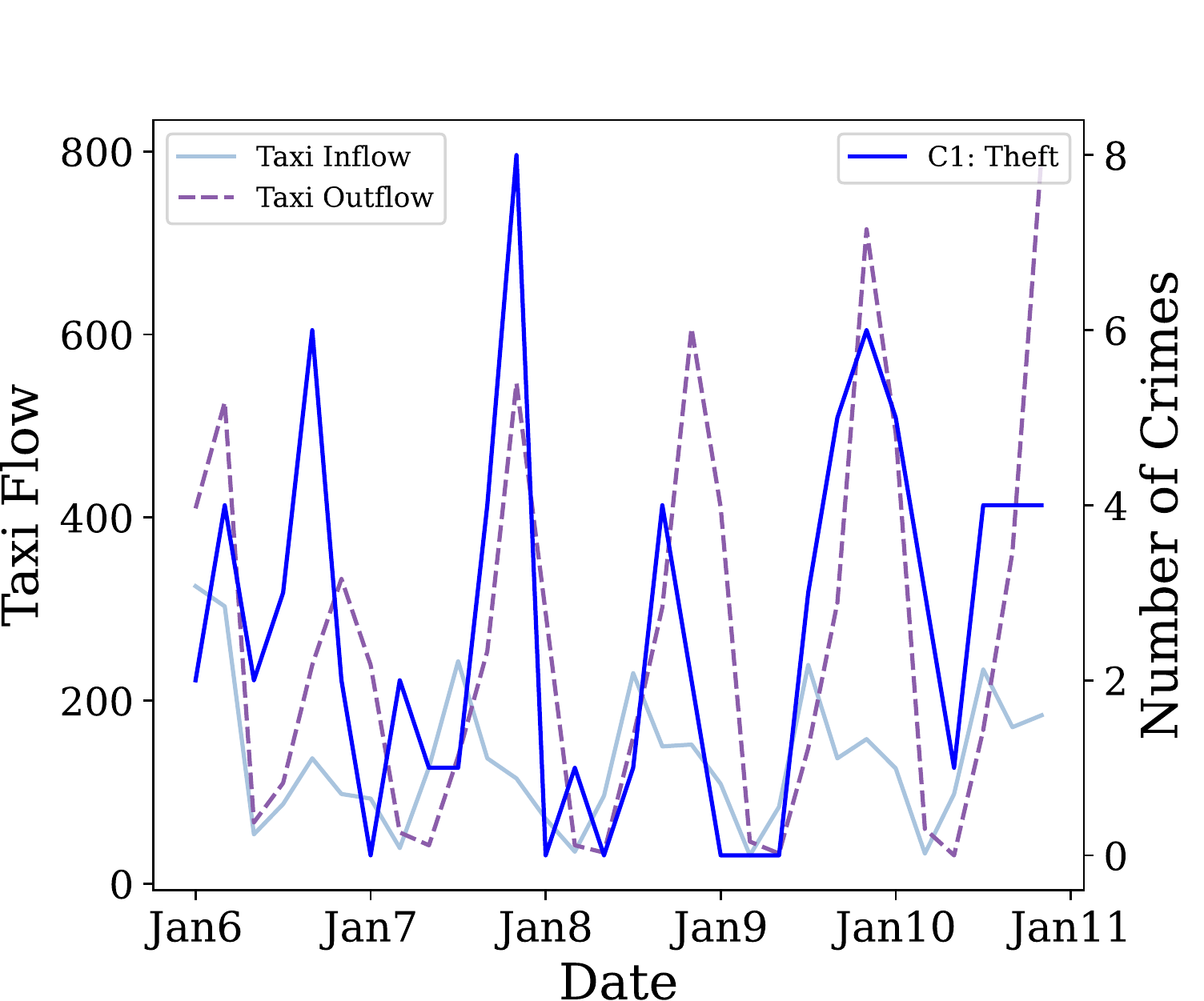
		\caption{}
		\label{Fig:ext_cor_2}
	\end{subfigure}
	\begin{subfigure}{0.32\columnwidth}
	    \def\svgwidth{\columnwidth}
	    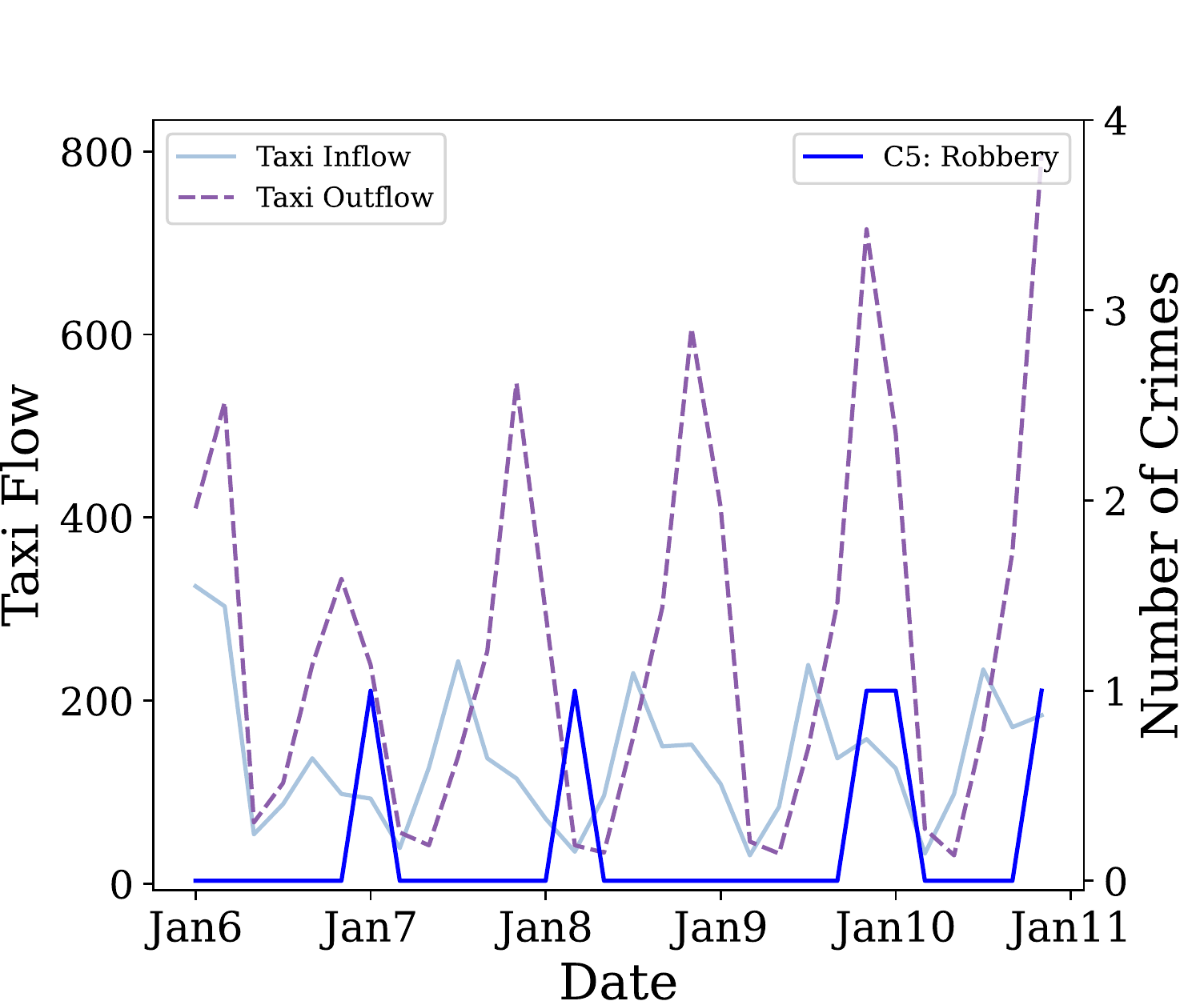
		\caption{}
		\label{Fig:ext_cor_3}
	\end{subfigure}
	\caption{Influence of Taxi Flows on the crime distribution in Chicago Communities}
	\label{Fig:ext_cor}
	\vspace{-1em}
\end{figure}

\textbf{External Features.} Functionalities and urban characteristics of a region like points of interests (POIs), traffic flow have direct influence on its crime occurrences. The influence of these external features on the crime occurrences tend to vary from time to time and region to region. Figure~\ref{Fig:ext_cor} shows such an example in Region $8$, where the distribution of deceptive practice (C0) (Figure~\ref{Fig:ext_cor_1}) and theft (C1) (Figure~\ref{Fig:ext_cor_2})  have a strong correlation with taxi flows, whereas robbery (C5) (Figure~\ref{Fig:ext_cor_3}) comparatively shows a weaker correlation with taxi flows.

Modeling these diverse spatio-temporal correlations and learning meaningful external features and their probable influence on crime are challenging tasks. Traditional interpretable machine learning and data mining methods~\cite{DBLP:conf/kdd/WangKGL16, 4666600, DBLP:conf/pakdd/Yu0CM14} cannot model these non-linear spatio-temporal correlations and thus fail to predict the crime occurrences accurately. Recent deep learning models~\cite{DBLP:conf/cikm/HuangZZC18, DBLP:conf/www/HuangZZWCY19} capture this non-linear spatial and temporal dependencies to some extent and improve the accuracy of traditional models. They still have major limitations:


\begin{compactitem}
\item The models only learn static spatial correlations. However, the correlations for two regions may vary with time.

\item The models do not address long term (e.g. daily, weekly) temporal correlations.  

\item The models do not consider the external features and hence the learned region embedding is incomplete.
\item The models lack interpretability. Both these models use LSTM based attention weights which are difficult to interpret because of the recurrence on the hidden states generated by LSTMs~\cite{DBLP:conf/nips/ChoiBSKSS16}. They are also not sparse enough to be meaningful for long sequence.

\end{compactitem}




To overcome the limitations, we develop AIST that captures dynamic spatio-temporal correlation for crime prediction and provides quantitative insights based on external features behind a prediction. We develop two novel variants of graph attention networks (GAT)~\cite{DBLP:conf/iclr/VelickovicCCRLB18}, $hGAT$ and $fGAT$ to learn the crime and feature embedding of the nodes (regions), respectively at each time step. These embedding are then fed to three sparse attention based-LSTMs (SAB-LSTMs)~\cite{DBLP:conf/nips/KeGBBMPB18} for modeling recent, daily and weekly crime trends. Finally, AIST applies a location-based attention mechanism to identify the significance of different trends to make a prediction. 

GAT does not consider the hierarchical information of nodes. However, in real-world scenarios nodes tend to form clusters and belong to different hierarchies based on similar characteristics. In urban context, nodes (regions) that belong to a same hierarchy shares similar functionalities and crime distributions. We propose $hGAT$ that incorporates this prior knowledge of hierarchical information into GAT's architecture to produce a better crime embedding of nodes.

Concatenating the feature vectors with spatial~\cite{DBLP:conf/aaai/Yao0KTJLGYL18, DBLP:conf/aaai/YaoTWZL19} or temporal view~\cite{DBLP:conf/aaai/LiZKXZ19} either directly or after a linear transformation is a common practice for incorporating the external features into the model. However, it fails to fully utilize the features and generate insights for a model's prediction. We propose $fGAT$ that replaces the additive self-attention mechanism of GAT with a novel scaled dot product self-attention mechanism~\cite{DBLP:conf/nips/VaswaniSPUJGKP17} to learn crime and region specific relevant feature embedding.

The unique challenge of a crime prediction problem that does not apply to other spatio-temporal prediction problems (e.g., traffic flow prediction, crowd flow prediction, passenger demand prediction) is the fact that crime data is spatially, temporally and categorically extremely sparse. AIST utilizes the feature embedding learned from fGAT to tackle the sparseness of crime data. On top of that, it is also necessary to keep the crime prediction  model's architecture reasonably interpretable, which makes the tasks even harder than building a spatio-temporal model that does not consider interpretability~\cite{10.1038/s42256-019-0048-x}.



AIST is interpretable because it takes transparent decisions at each prediction step based on the different attention modules used in the model architecture. To explain a prediction, we first find whether the prediction is based on recent occurrences or any recurring trend and then identify the previous time steps that are given the most importance. Our model knows why a time step is given importance as the input at each time step is an interpretable spatial embedding. Hence, if we backtrack we can find the most important regions and features for a specific time step. Even though attention as a form of explanation is not new in spatio-temporal literature~\cite{DBLP:conf/cikm/HuangZZC18, DBLP:conf/www/HuangZZWCY19, DBLP:conf/aaai/GuoLFSW19, DBLP:conf/cikm/ZhangHXX20}, simply using an attention module does not make a model interpretable~\cite{DBLP:conf/acl/SerranoS19,DBLP:conf/naacl/JainW19}.  Keeping this in mind, unlike existing spatio-temporal literature the model architecture of AIST is designed so that it complies with the conditions presented in~\cite{DBLP:conf/emnlp/WiegreffeP19} under which attentions can be regarded as faithful explanations.


Besides inherent interpretable architectures~\cite{DBLP:conf/nips/ChenLTBRS19}, recently post-hoc local explanation techniques~\cite{DBLP:conf/nips/LundbergL17} that provide approximate explanations to a model's decision making have been explored to imitate the behavior of deep learning black box models. However, they are not well received considering the fact that if these explanations had been adequate enough, there would be no need for the original model~\cite{10.1038/s42256-019-0048-x}. Hence, we keep AIST architecture inherently interpretable while ensuring its accuracy. 




In summary, the contributions of this paper are as follows:
\begin{compactitem}
    \item We propose a novel interpretable spatio-temporal deep learning model, AIST which is able to capture diverse spatio-temporal correlations based on past crime occurrences, external features and recurring trends.
    
    \item We propose $hGAT$, a novel GAT variant that allows AIST to learn more faithful node embedding.
    
    \item We propose $fGAT$, another novel GAT variant that provide insights behind the predictions of AIST.
    
    
    \item We conduct experiments on Chicago crime data. AIST achieves a higher accuracy than the state-of-the-art methods and provides useful insights for its predictions. Experiment results also validate that the explanations provided by different attention modules in hGAT, fGAT and SAB-LSTMs are faithful.
\end{compactitem}

The remaining of the paper is organized as follows. We discuss the related work in Section~\ref{Sec:relatedWork} and formulate the crime prediction problem in Section~\ref{Sec:problemFormulation}. We present our model, $AIST$ in Section~\ref{sec:model}. Section~\ref{Sec:experimentalResult} presents the experimental results and evaluates the accuracy and the interpretability of AIST. Section~\ref{sec:conclusion} concludes the paper.

\section{Related Work}
\label{Sec:relatedWork}

Data-driven crime prediction problems have received wide attention from the researchers for decades. Existing studies on crime prediction can be divided into following categories: (i) \emph{crime rate inference} that predicts the crime rate of a region, (ii) \emph{crime hotspot detection} that finds the locations where crimes are clustered, and (iii) \emph{crime occurrence prediction} that forecasts the occurrence of a crime category for a location at a future timestamp. Our work falls in the third category. In Sections~\ref{sec:crimeModels} and~\ref{sec:interpretableModels}, we elaborate existing crime prediction models and interpretable models, respectively. In Section~\ref{spatio-temporal}, we discuss the deep learning methods used for spatial-temporal prediction. 

\subsection{Crime Prediction Models}
\label{sec:crimeModels}
\textbf{Statistical and Classic Machine Learning Methods.} Recent studies~\cite{DBLP:conf/kdd/WangKGL16, DBLP:journals/tbd/WangYKGL19, DBLP:journals/corr/abs-1908-02570} used statistical and classic machine learning methods (e.g., linear regression, negative binomial regression, geographically weighted regression, random forest) for crime rate inference problem. In~\cite{DBLP:conf/kdd/WangKGL16, DBLP:journals/tbd/WangYKGL19}, the authors studied the effect of point of interest (POI) (e.g., a restaurant or a shopping mall) and taxi flow information along with the traditional demographics features of a region while in~\cite{DBLP:journals/corr/abs-1908-02570}, the authors utilized FourSquare check-in data for estimating the crime rate of a particular region. Researchers have also employed kernel density estimation (KDE)~\cite{article, DBLP:journals/Hart, eck2005mapping, DBLP:conf/sibgrapi/NetoSV16} for predicting hot-spot maps. However, these works only take spatial features and dependencies into account ignoring the temporal dynamics of crime.

To address the temporal dynamics, time-series models such as autoregressive integrated moving average (ARIMA)~\cite{4666600} have been proposed for one-week ahead crime occurrence prediction. In~\cite{doi:10.1198/jasa.2011.ap09546}, the authors implemented a self-exciting point process similar to one used by the seismologists in the context of urban crime to understand the temporal trends of burglary. Even though these models acknowledge the temporal dynamics, they do not incorporate the spatial context of crimes.

Both spatial and temporal information have been also explicitly modeled in the literature. In~\cite{DBLP:conf/pakdd/Yu0CM14}, the authors proposed an algorithm that constructs a global crime pattern from local crime cluster distributions, and employed it for predicting residential burglary. In~\cite{DBLP:journals/tgis/NakayaY10}, the authors employed STKDE, a variant of KDE for mapping transient and stable crime clusters. The work in~\cite{DBLP:journals/tist/TooleEP11} used analytic and statistical techniques to identify the spatio-temporal crime patterns. In~\cite{DBLP:conf/cikm/ZhaoT17}, spatio-temporal correlations like intra-region temporal correlation and inter-region spatial correlation have been considered for crime occurrence prediction. However, all of these methods cannot fully model the complex non-linear relation of space and time and  the dynamicity of spatial-temporal correlation.

Besides spatio-temporal features, incorporating additional data (e.g. Twitter, demographics data) improve the accuracy of existing crime prediction models. The authors in ~\cite{DBLP:journals/dss/Gerber14} added Twitter-based features extracted from topic based modeling for improving the prediction of models. In~\cite{10.1145/1938606.1938608}, the authors used fuzzy association rule mining to find consistent crime patterns using population demographics information of communities. Another line of work~\cite{DBLP:conf/gis/XiongSKDPS19, DBLP:conf/www/WangJWWL19} explores the heterogeneous and task-specific division of spatial regions over traditional grid and community based division which helps improve the accuracy of the crime prediction.


\textbf{Deep Learning Methods.} Deep learning models have recently been shown to be very effective in domains like computer vision, speech recognition and natural language processing. Recent deep learning models have also attempted to capture the non-linear spatio-temporal dependencies of crime. DeepCrime~\cite{DBLP:conf/cikm/HuangZZC18}, a hierarchical recurrent framework with attention mechanism, considers temporal correlation, its inter-relation with ubiquitous data and category dependencies for future crime prediction. However, DeepCrime does not consider spatial correlations of crimes. In~\cite{DBLP:journals/corr/WangZZBB17}, the authors applied ST-ResNet architecture~\cite{DBLP:conf/aaai/ZhangZQ17} for crime intensity prediction while in~\cite{DBLP:conf/www/HuangZZWCY19}, the authors developed MiST, a LSTM based neural network architecture with attention mechanism to model spatio-temporal and cross-categorical correlation for crime prediction. None of these models can capture dynamic spatial correlation and identify the impact of external features on crime predictions. Besides, these models are not interpretable. DeepCrime and MiST employ attention based RNNs which lack interpretability because of the recurrence on the hidden states generated by RNNs and their non-sparse attention weights for longer sequences. ST-ResNet uses deep residual units with hundreds and thousands of CNNs stacked altogether which makes it harder to interpret the model's prediction.

\subsection{Interpretable Models}
\label{sec:interpretableModels}
The statistical and classic machine learning models have an advantage over deep learning models in terms of interpretability. However, they cannot model the complex non-linearity of space and time and thus lacks accuracy. On the other hand, though neural networks can capture the spatial-temporal non-linear relationship, they are not interpretable. 

Attention-based models focus on the most relevant information while performing a certain task. These models have become very popular in image processing~\cite{DBLP:conf/icml/XuBKCCSZB15, DBLP:conf/nips/MnihHGK14, DBLP:journals/corr/BaMK14, DBLP:conf/cvpr/FuZM17} and natural language processing~\cite{DBLP:journals/corr/BahdanauCB14, DBLP:conf/nips/VaswaniSPUJGKP17}, and health-care predictions~\cite{DBLP:conf/nips/ChoiBSKSS16, DBLP:conf/kdd/ChoiBSSS17, DBLP:conf/kdd/BaiZEV18, DBLP:conf/kdd/MaCZYSG17} for ensuring interpretability. Similar to the statistical and classic machine learning models, despite of having a self-explanatory structure, simply using an attention based model does not make an explanation of a prediction faithful. The model architecture of AIST provides faithful explanations, which is also validated by our experiment results.

The other category of interpretable models is post-hoc models, where a separate model is used for explanation. Examples of post-hoc models include LIME~\cite{DBLP:conf/kdd/Ribeiro0G16}, SHAP~\cite{DBLP:conf/nips/LundbergL17}, rule-based learning~\cite{DBLP:journals/corr/SuWVM16} and saliency visualizations~\cite{DBLP:conf/nips/DabkowskiG17}.


\subsection{Deep Learning for Spatio-temporal Prediction}
\label{spatio-temporal}

Deep learning methods have become popular in recent years in the domain of spatial temporal prediction. A common approach is to use the convolution based architecture (CNN)~\cite{DBLP:conf/gis/ZhangZQLY16, DBLP:journals/tits/ChenYL18, DBLP:conf/icdm/ChenLTCZYVFZ18} for finding the spatial correlation and the recurrent based architecture~\cite{DBLP:conf/kdd/RongXYM18, DBLP:journals/corr/abs-1801-02143} for  finding the temporal correlation. In~\cite{DBLP:conf/aaai/YaoTWZL19}, the authors used both CNN and attention-based LSTM to capture the dynamic spatio-temporal dependencies for traffic prediction. In~\cite{DBLP:conf/ijcai/LiangKZYZ18}, the authors proposed a multi-level attention mechanism along with a recurrent layer and a fusion module to incorporate the external features for geo-sensory time prediction.

Recent literature~\cite{DBLP:conf/aaai/GuoLFSW19, DBLP:conf/ijcai/YuYZ18, DBLP:journals/TKDE/9139357, DBLP:conf/cikm/ZhangHXX20, DBLP:conf/cikm/XieG0X0Z20, DBLP:conf/www/Wang0WJWTJY20, DBLP:conf/kdd/HongLYLFWQY20} has also started exploring the graph neural networks for such prediction. In~\cite{DBLP:conf/aaai/GuoLFSW19, DBLP:conf/ijcai/YuYZ18, DBLP:conf/kdd/HongLYLFWQY20}, the authors proposed a pure convolutional structure in the form of a graph convolution in the spatial dimension and a general convolution in the temporal dimension to model the traffic flows. In~\cite{DBLP:journals/TKDE/9139357}, several temporal views are fed to their respective graph convolution layers and then fused altogether along with semantic views to model the crowd flows. In~\cite{DBLP:conf/www/Wang0WJWTJY20}, the authors modeled traffic flows with a graph convolutional network (GCN) followed by a recurrent layer and a transformer to capture the local and global temporal correlation, respectively. Both these models~\cite{DBLP:journals/TKDE/9139357, DBLP:conf/www/Wang0WJWTJY20} incorporate geospatial position of nodes into the GCN to better model the spatial dependencies. In~\cite{DBLP:conf/cikm/ZhangHXX20}, the authors proposed STCGA that combines multiple self-attention, graph attention, and convolutional residual networks to predict the traffic flow.~\cite{DBLP:conf/cikm/XieG0X0Z20} proposes DIGC and a pre-trained binary classifier, both of which consists of a GCN followed by an LSTM to extract the spatio-temporal and latent incident crime features, respectively for traffic speed prediction. 

None of these spatio-temporal prediction models are designed to handle the sparseness of crime data. The finer the spatial, temporal or categorical resolution gets, the sparser the crime data becomes; which makes it even harder to model the crime. Hence, the spatio-temporal prediction models fail to perform well for crime prediction tasks. Unlike existing spatio-temporal literature, AIST chooses a handful of region and crime category specific external features, and applies fGAT to learn a more stable and faithful feature embedding of the target region as a substitute of the sparse crime data. This learned feature embedding along with the crime embedding learned by hGAT allows AIST to capture the slightest of changes in the feature or crime embedding of the target region over time and make predictions accordingly. Our experiment results also show that AIST outperforms the high-performance spatio-temporal prediction models.

\begin{small}
\begin{table}[htbp]
	\caption{Notations and their meanings} 
	\centering 
	\begin{tabular}{|c|p{9cm}|}
		\hline \hline 
		Notation & Symbol \\ 
		\hline
		$N, T, K, J$ & Number of regions, time steps, crime categories, external features\\  
		\hline
		$\mathcal{N}_i$ & First-order neighbors of region $r_i$ (including itself)\\  
		\hline
		$\tau$ & Length of a time step\\  
		\hline
		$x_{i, t}^k$ & Crime occurrences of $k$-th category at region $r_i$ during $t$-th time step\\ 
		\hline
		$\textbf{x}_{i, t}$ & Crime occurrences of all categories at region $r_i$ during $t$-th time step\\ 
		\hline
		$\textbf{X}_{t}$ & Crime occurrences of all categories at all regions during $t$-th time step\\ 
		\hline
		$f_{i, t}^j$ & $j$-th external feature of region $r_i$ during $t$-th time interval\\ 
		\hline
		$\textbf{f}_{i, t}$ & All external features of region $r_i$ during $t$-th time step\\ 
		\hline
		$\textbf{F}_{t}$ & All external features of all regions during $t$-th time step\\ 
		\hline
		$\hat{\textbf{Y}}_{T+1}$ & Predicted crime occurrences of all regions and categories of the city at $(T+1)$-th time step\\
		\hline
	\end{tabular}
	\label{symbols}
\end{table}
\end{small}

\section{Problem Formulation}
\label{Sec:problemFormulation}

In this section, we introduce some notations\footnote{Bold letters, e.g. \textbf{$A, a$} denote matrices and vectors respectively and small letters, e.g. $a$ denote scalars} and formulate crime prediction problem as a regression task. Table~\ref{symbols} summarizes the notations used in the paper.

\textbf{Region.} We model a city with an undirected graph $G = (V, E)$, where $V$ represents a set of $N$ regions $\{r_1, r_2, r_3,\ldots, r_N\}$ and $E$ represents a set of edges connecting them. In this study, a region denotes a community area: a pre-defined administrative boundary that serves various planning and statistical purposes. For a region $r_{i}$ to be connected to region $r_{i'}$ they must share a common boundary. 

\textbf{Crime Occurrence.} Let $x_{i, t}^k \in \mathbb{R}$ represent the number of crimes reported of category $k$ (e.g. theft) at region $r_i$ during $t$-th time step\footnote{We use time step and time interval synonymously.}. If $\mathbf{x}_{i, t} = [x_{i, t}^1, x_{i, t}^2, x_{i, t}^3, \ldots, x_{i, t}^K] \in \mathbb{R}^{K}$ denotes the reported crimes of all $K$ categories at region $r_i$ during $t$-th time step and $\mathbf{X}_{t} = (\mathbf{x}_{1, t}, \mathbf{x}_{2, t}, \mathbf{x}_{3, t}, \ldots, \mathbf{x}_{N, t}) \in \mathbb{R}^{K\times N}$ denotes the reported crimes of all categories at all $N$ regions during $t$-th time step, then the crime occurrences of the whole city for $T$ time steps can be denoted as $\mathcal{X} = (\mathbf{X}_1, \mathbf{X}_2, \ldots, \mathbf{X}_T) \in \mathbb{R}^{K\times N\times T}$.

\textbf{External Feature.} We use POI information, traffic inflow and traffic outflow as external features for improving the model's prediction accuracy. The external features of the city during $T$ time steps are denoted as  $\mathcal{F} = (\mathbf{F}_1, \mathbf{F}_2, \ldots, \mathbf{F}_T) \in \mathbb{R}^{J\times N\times T}$ for $J$ external features, where $\mathbf{F}_{t} = (\mathbf{f}_{1, t}, \mathbf{f}_{2, t}, \mathbf{f}_{3, t}, \ldots, \mathbf{f}_{N, t}) \in \mathbb{R}^{J\times N}$ denotes the external features of the city during $t$-th time step, $\mathbf{f}_{i, t} = [f_{i, t}^1, f_{i, t}^2, f_{i, t}^3, \ldots, f_{i, t}^J] \in \mathbb{R}^{J}$ denotes the external features of a region $r_i$ during time step $t$ and $f_{i, t}^j \in \mathbb{R}$ denotes the $j$-th external feature of a region $r_i$ during time step $t$. 



\textbf{Problem Definition.} Given past crime occurrences $\mathcal{X}$ and external features $\mathcal{F}$ for last $T$ time steps, predict $\hat{\textbf{Y}}_{T+1}$, the crime occurrences of the city during $(T+1)$-th time step.

\begin{figure*}[tbp]
	\centering
	\def\svgwidth{\textwidth}
	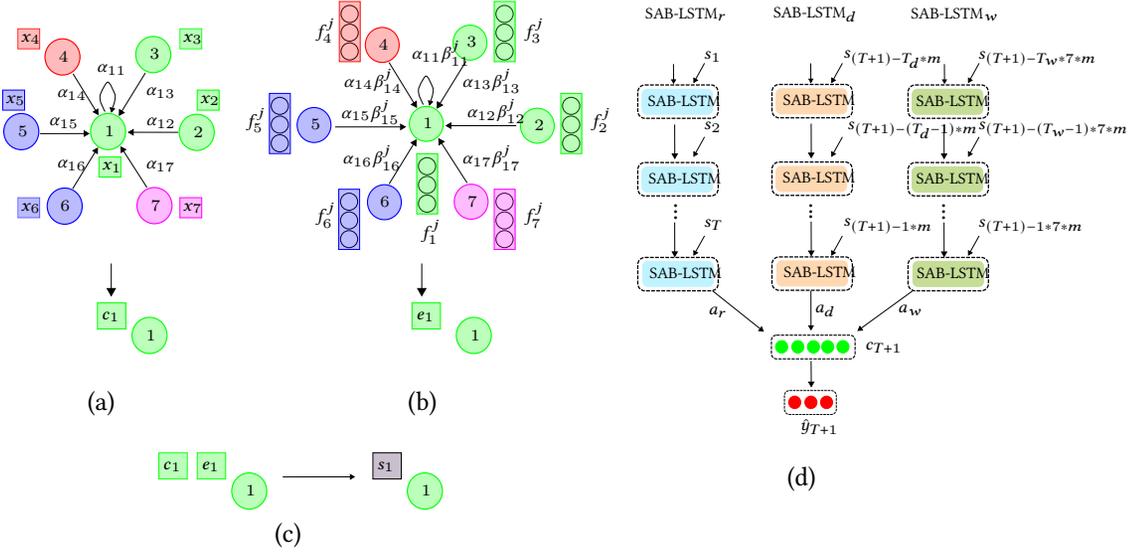
\caption{Model Overview of AIST: (a) hGAT is applied to calculate the crime embedding $c_1$ of target region $r_1$, (b) fGAT is applied to calculate the feature embedding $e_1$ of $r_1$, (c) both $c_1$ and $e_1$ are concatenated to produce spatial representation $s_{1}$ of $r_1$ at time step $t$, (d) the spatial representations generated at different time steps are then fed to three SAB-LSTMs to capture recent, daily and weekly crime trends at $r_1$ and a location-based attention is applied to predict the crime occurrence of $r_1$ at $(T+1)$-th time step, $\hat{y}_{T+1}$.}
	\label{Fig:overview}
	\vspace{0em}
\end{figure*}

\section{Model Description}
\label{sec:model}
The key idea behind our model's high prediction accuracy is that we exploit (i) hierarchical information of regions, (ii) external features, and (iii) short, long term crime patterns to capture the dynamic spatio-temporal dependencies while keeping the model's architecture reasonably interpretable. Given the crime occurrences of category $k$ at region $r_i$ during time steps $[1..T]$, we find the crime embedding $\mathbf{c_{i, t}^k}$ and feature embedding $\mathbf{e_{i, t}^k}$ of $r_i$ using our proposed hGAT and fGAT, respectively, and concatenate them to produce spatial representation $\mathbf{s_{i, t}^k}$ for each of these time step. These embedding are then fed to three SAB-LSTMs for capturing recent, daily and weekly trends which outputs the hidden states $\mathbf{h_{T+1}^r}, \mathbf{h_{T+1}^d}, \mathbf{h_{T+1}^w}$, respectively. After applying an attention mechanism on these hidden states, a context vector, $\mathbf{c_{T+1}}$ is generated to predict the crime occurrence at $(T+1)$-th time step, $\hat{y}_{i, T+1}^k$. Figure~\ref{Fig:overview} gives an overview of the model. In Section~\ref{spatial}, we elaborate on how we generate crime embedding using hGAT and feature embedding using fGAT to produce spatial representation (Figures~\ref{Fig:overview}a--~\ref{Fig:overview}c). In Sections~\ref{temporal} and~\ref{prediction}, we discuss our crime trend generation and prediction steps, respectively (Figure~\ref{Fig:overview}d). 
\subsection{Spatial View}
\label{spatial}
Convolutional neural networks (CNNs)~\cite{DBLP:conf/gis/ZhangZQLY16} and its variants~\cite{DBLP:conf/aaai/YaoTWZL19} have been applied to model spatial correlation between regions in spatio-temporal prediction. Though CNNs learn meaningful features on regular grid structured data, they do not perform well on irregular graph data because the number of nodes in a graph and their neighbor counts are variables. Urban crime data exhibit a clear graph structure considering the correlation between regions and other external features. Modeling them as grid structured data results in incomplete information and  makes it hard to learn meaningful information. To address this issue, graph convolutional networks (GCNs)~\cite{DBLP:conf/iclr/KipfW17} have gained popularity in recent times. GCNs learn a node's embedding as an aggregation of its neighbor's features and calculate their contribution with predefined Laplacian Matrix, which is the difference of the degree matrix and the adjacency matrix of the graph. Since the contributions of neighbor nodes are static, GCNs can not capture the dynamic spatial correlation between regions. 

Based on these observations, we choose GAT as the base architecture to capture the spatial dependencies for crime prediction. Similar to GCN, GAT learns a node's embedding as an aggregation of its neighbor's features but uses a self-attention mechanism to learn their contributions instead. GAT does not require any costly matrix operation and knowledge about the graph structure upfront, which allows GAT to learn dynamic spatial correlation between regions. 

We use two GAT variants: hGAT and fGAT to learn the crime and feature embedding of a target region as follows.

\textbf{\emph{Crime Embedding.}}
The city of Chicago is divided into $77$ communities (regions) and the communities are grouped into $9$ districts or sides forming a containment hierarchy. Communities under the same side tend to share similar socio-economic, demographic and urban features, which result in similar crime distribution than those under different sides. Hence, while aggregating the node (community) information in GAT, prioritizing the nodes that fall under the same side with the target node over others may help to learn better crime representation of a target node. From this intuition, we propose $hGAT$ to amplify the signals of the nodes that fall under the same side with the target node than those which do not. Since almost every city can be divided into multiple partitions at different spatial resolutions, this idea can be generalized to other cities as well.
\begin{figure}[tbp]
	\begin{subfigure}{3.5 cm}
		\def\svgwidth{\columnwidth}
		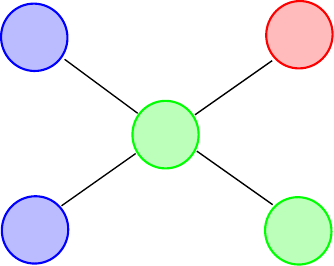
		\caption{First-hop neighbors}
	\end{subfigure}
	\hspace{4 mm }
	\begin{subfigure}{4.5 cm}
		\def\svgwidth{\columnwidth}
		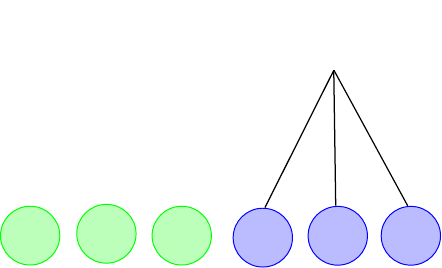
		\caption{Hierarchical Structure}
	\end{subfigure}
	\caption{Complex spatial interaction between regions}
	\label{Fig:region_structure}
	\vspace{-1em}
\end{figure}

\begin{figure}[htbp]
	    \centering
	    \def\svgwidth{0.75\textwidth}
	    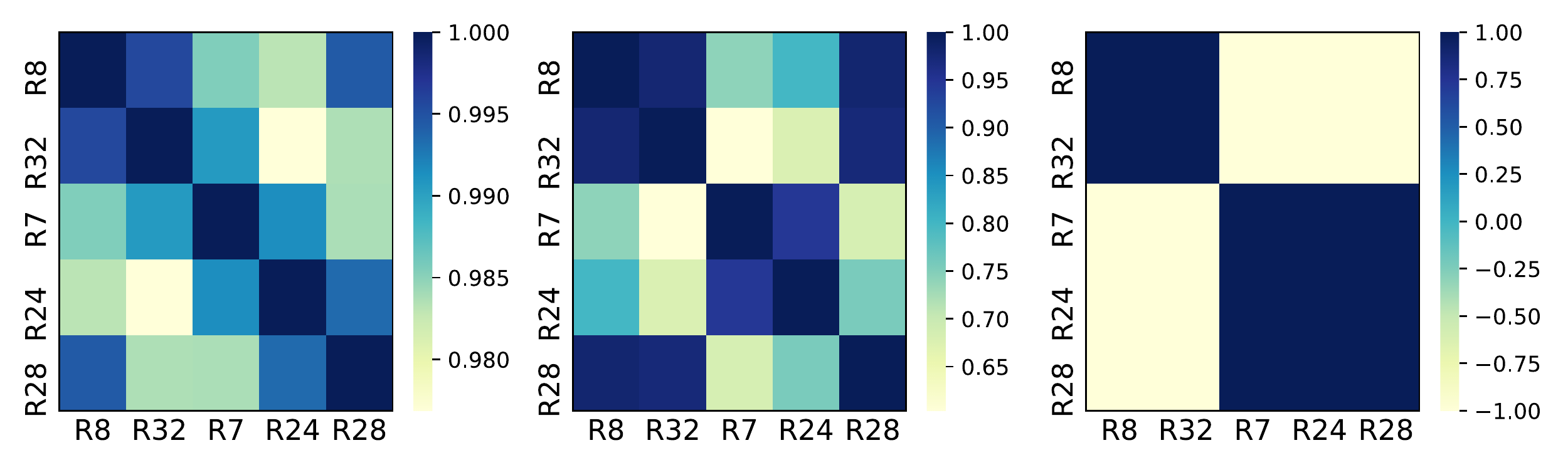
	\caption{Pearson correlation coefficient among regions of Chicago based on 2019 crime, POI and 2019 taxi flows distribution (left to right)}
	\label{Fig:hgat_cor_total}
\vspace{-1em}
\end{figure}

Figure~\ref{Fig:region_structure} represents a scenario where the first-order neighbors of target region $8$ are $\{7, 8, 24, 28, 32\}$ and target region $8$ along with region $32$ and $33$ fall under the same side (represented as grey circles in Figure~\ref{Fig:region_structure}). It is evident from Figure~\ref{Fig:hgat_cor_total} that R8 is strongly correlated to R32 than other nearby regions (R7, R24) in terms of crime distribution and external features (POI and taxi flows). Similarly, R24 shows a stronger correlation with R28 than R8. Thus, it is expected that the target region 8 is influenced more by region $32$ which is not only a first-hop neighbor but also falls under the same side, whereas regions $7, 24$ or $28$ do not. To be clear, we do not consider the influence of regions such as $33$ that falls under the same side with the target region 8, but is not a first-hop neighbor. We only want to amplify the signal from those regions which satisfy both conditions: falls under the same side and is a first-hop neighbor. 

hGAT considers two sets of features for each node ($r_i$): (i) node level features: crime occurrences at community level during time step $t$: $x_{i, t}^k$, (ii) parent level features: crime occurrences at district/side level during time step $t$: $z_{i, t}^k$ as input to GAT and an additional attention layer to capture the similarity between nodes based on parent level features. Let the parent node of a region $r_i$ be $p_j = Parent(r_i)$. Then the parent feature of node $r_i$ is calculated as $z_{i, t}^k = \sum_{\forall r_{n} Parent(r_{n})=p_j}x_{n, t}^k$. Basically, we sum the crime occurrences of category $k$ across the nodes whose parent node is $p_j$ to create the parent feature of target node $r_i$. Traditional GAT considers only node level features. Hence, it can not model hierarchical information into a node's embedding.
For hGAT, we use two transformation matrix, (i) $\mathbf{w_x}\in \mathbb{R}^{F}$ to learn the similarities between a target region and its neighbor's node level features, and (ii) $\mathbf{w_z}\in \mathbb{R}^{F'}$ to learn similarities between their parent-level features. Based on these information two separate feed-forward attention layer computes two sets of pair-wise unnormalized attention scores between the target region and its first-hop neighbors: $e_{ii'}^c$ and $e_{ii'}^p$, respectively. For clarity, we omit the indices of crime categories ($k$) and time step ($t$).
\begin{align*}
e_{ii'}^c &= \text{LeakyReLU}(\mathbf{a_x}^T[\mathbf{w_x}x_{i} \mathbin\Vert \mathbf{w_x}x_{i'}]) \\
e_{ii'}^p &= \text{LeakyReLU}(\mathbf{a_z}^T[\mathbf{w_z}z_{i} \mathbin\Vert \mathbf{w_z}z_{i'}])
\end{align*}

We perform an element-wise addition to combine these two sets of unnormalized attention scores and apply softmax over them to generate final attention scores, where $\mathcal{N}_i$ denotes first order neighbors including itself. 
$$e_{ii'} = e_{ii'}^c + e_{ii'}^p$$
$$\alpha_{ii'} = \text{softmax}_{i'}(e_{ii'}) = \frac{exp(e_{ii'})}{\sum_{i'' \in \mathcal{N}_i}exp(e_{ii''})}$$
Finally, we use this combined attention score to update the crime embedding ${c}_{i, t}^k$ of target region $r_i$. 
\begin{equation}
\label{eq:1}
\mathbf{{c}}_{i, t}^k = \sigma\left(\sum_{i' \in \mathcal{N}_i}\alpha_{ii'}\mathbf{w_x}x_{i', t}^k\right)
\end{equation}
Figure~\ref{Fig:overview}a gives an overview of generating the crime embedding. 

\textbf{\emph{Feature Embedding.}} Besides historical crime observations, external features have been shown to be useful in crime prediction problems~\cite{DBLP:conf/cikm/HuangZZC18, DBLP:conf/cikm/ZhaoT17, DBLP:conf/kdd/WangKGL16}. We propose $fGAT$ that replaces additive self-attention mechanism with a novel scaled dot product self-attention mechanism~\cite{DBLP:conf/nips/VaswaniSPUJGKP17} to learn category specific feature embedding of regions. 


The feature embedding of a target region is formulated as an aggregation of its neighbors' features based on their possible influence on the crimes of the target region. The intuition behind finding a possible influential feature is - if two regions having similar features experience similar crime occurrences at a specific time step then the features might influence crimes or serve as proxies for crime prediction in addition to the crime occurrences.
We compute the query vector $\mathbf{q}_{ii'}$ by multiplying the concatenated crime occurrences of a target region ($r_i$) and its neighbor region ($r_{i'}$) with weight matrix $\mathbf{W_q} \in \mathbb{R}^{d_q\times 2}$ to learn their crime distribution similarities. For preparing the key vector $\mathbf{k}_{ii'}^j$ for feature $j$, we multiply the concatenated features of $r_i$ and $r_{i'}$ with weight matrix $\mathbf{W_k} \in \mathbb{R}^{d_k\times 2}$ to learn their feature similarities. Here, $d_q$ and $d_k$ represents the dimension of the query and key vector, respectively.
\begin{align*}
    \mathbf{q}_{ii'} = \mathbf{W_q}([x_{i, t}^k \mathbin\Vert x_{i', t}^k])\\
    \mathbf{k}_{ii'}^j = \mathbf{W_k}([f_{i, t}^j \mathbin\Vert f_{i', t}^j])
\end{align*}
Then, the attention weight of $j$-th feature of $r_{i'}$ is calculated using the dot-product attention mechanism.
$$\beta_{ii'}^{j} =\text{softmax}_{j}(\frac{\mathbf{q}_{ii'}{\mathbf{k}_{ii'}^j}^T}{\sqrt{d_k}}) $$
Once the attention weights of individual features are found, the feature embedding ${e}_{i, t}^k$ of $r_i$ is formulated as follows. 
\begin{equation}
\label{eq:2}
\mathbf{{e}}_{i, t}^k = \sigma\left(\sum_{i' \in \mathcal{N}_i}\left( \alpha_{ii'} \sum_{j=1}^{J}\beta_{ii'}^{j} \mathbf{w_v}f_{i', t}^j\right)\right)
\end{equation}

Here $ \beta_{ii'}^{j}\mathbf{w_v} f_{i', t}^j$ represents the contribution of $j$-th feature of $r_i'$ on the feature embedding of $r_i$ and $\mathbf{w_v} \in \mathbb{R}^{F}$. Figure~\ref{Fig:overview}b gives an overview of generating the feature embedding.

Finally, We concatenate the crime embedding, $\mathbf{{c}}_{i, t}^k$ and feature embedding, $\mathbf{{e}}_{i, t}^k$ to find spatial embedding $\mathbf{s}_{i, t}^k$ of target region $r_i$ at $t$-th time step for crime category $k$. This spatial embedding $\mathbf{s}_{i, t}^k$ is fed as input to a SAB-LSTM cell for time step $t$ as shown in Figure~\ref{Fig:overview}c.
$$\mathbf{s}_{i, t}^k = [\mathbf{{c}}_{i, t}^k \mathbin\Vert \mathbf{{e}}_{i, t}^k]$$

\subsection{Temporal View}
\label{temporal}
LSTM and GRU are two popular recurrent neural networks (RNNs) that capture temporal correlations. However, besides being non-interpretable they suffer from vanishing gradient problem for long sequences. To address these issues, attention-based RNNs are proposed that use attention mechanism to focus on relevant hidden states. These attention weights are difficult to interpret because of the recurrence on the hidden states generated by LSTMs~\cite{DBLP:conf/nips/ChoiBSKSS16}. They are also not sparse enough to be meaningful for long sequence.

SAB-LSTM~\cite{DBLP:conf/nips/KeGBBMPB18} back-propagates across only a selected small subset instead of all hidden states, which are selected using a sparse and hard attention mechanism. Thus it mitigates the gradient vanishing problem and is also interpretable. In this section, first we give a brief overview of a SAB-LSTM cell and then discuss how we apply them in predicting crimes.


At each time step $t$ the underlying LSTM of SAB-LSTM takes the spatial embedding of region $r_i$, $\mathbf{s}_{i, t}^k$ and the previous hidden state $\mathbf{h_{t-1}}$ as inputs for crime category $k$. It produces a new cell state $\mathbf{c_t}$ along with a provisional hidden state $\mathbf{\hat{h}_t}$. 
$$\mathbf{\hat{h}}_{t}, \mathbf{c_{t}} = \text{LSTM}(\mathbf{s}_{i, t}^k, \mathbf{h_{t-1}})$$
The provisional hidden state, $\mathbf{\hat{h}_t}$ is concatenated with all the vectors stored in memory $\mathcal{M} = [\mathbf{h}_1^{mem}, \mathbf{h}_2^{mem}, \ldots, \mathbf{h}_{|\mathcal{M}|}^{mem}]$ and passed through a feed-forward neural network to generate unnormalized attention weights $(e_m)$ for each vector stored in the memory. Memory $\mathcal{M}$ contains a set of hidden states selected arbitrarily (after each $k_{att}$ time step) for comparison with the generated provisional hidden state. 
$$e_m = \mathbf{W_m}tanh(\mathbf{\hat{h}}_{t} \mathbin\Vert \mathbf{h}_m^{mem})$$
Then, SAB-LSTM subtracts the $(k_{top}+1)$-th highest attention score from all the attention scores and use normalization to generate $k_{top}$ sparse attention weights to select only $k_{top}$ memory cells.
$$\alpha_{m} =\frac{e_{m}-e_{k_{top}+1}}{\sum_{m^{''} \in \mathcal{M}} (e_{m^{''}}-e_{k_{top}+1})}$$


Once the attention weights ($\alpha_{m}$) are obtained, it calculates a summary vector $\mathbf{sum_t}$ by summing over the $k_{top}$ memories. This summary vector is then concatenated to the previously generated hidden provisional state $\mathbf{\hat{h}_t}$ to get the final hidden state, $\mathbf{h_t}$.
\begin{align*}
\mathbf{sum_t} &= \sum_{m \in \mathcal{M}}\alpha_{m}\mathbf{h}_m^{mem}\\ 
\mathbf{h_t} &= \mathbf{\hat{h}}_t + \mathbf{sum_t}
\end{align*}

The hidden state generated at time step $t$ has two contributing factors. First, the provisional hidden vector $(\mathbf{\hat{h}_t})$ which is the output of a traditional LSTM at time $t$ and non-interpretable. Second, the summary vector $\mathbf{sum_t}$ which is the summation of dynamic, sparse hidden states aligned with current state and interpretable. For crime prediction task we omit the first contributing factor and only use the summary vector $\mathbf{sum_t}$ as our output hidden state $\mathbf{h_t}$. Even though the accuracy is slightly compromised but this makes SAB-LSTM more interpretable.
\begin{equation}
\label{eq:3}
\mathbf{h_t} = \mathbf{sum_t} = \sum_{m \in \mathcal{M}}\alpha_{m}\mathbf{h}_m^{mem}
\end{equation}
We use three SAB-LSTMs for our crime prediction task (Figure~\ref{Fig:overview}d). $\text{SAB-LSTM}_r$ captures recent crime trends based on a target region's spatial embedding during past $T$ time steps. $\text{SAB-LSTM}_d$ captures daily trends based on spatial embedding at the same time step as the predicted time step but on previous days . Finally, $\text{SAB-LSTM}_w$ captures weekly trends based on the  the spatial embedding at the same time step as the predicted time step but on previous weeks. We formulate them as follows. For simplicity of representation, we omit region-index $i$ and crime category-index $k$.
\begin{align*}
\mathbf{h}_{T+1}^r &= \text{SAB-LSTM}_r(\mathbf{s}_{t}) \\
\mathbf{h}_{T+1}^d &= \text{SAB-LSTM}_d(\mathbf{s}_{(T+1)-t_{d}*m}) \\
\mathbf{h}_{T+1}^w &= \text{SAB-LSTM}_w(\mathbf{s}_{(T+1)-t_{w}*7*m})
\end{align*}
Here, $t = [1..T], t_d = [1..T_{d}], t_w = [1..T_{w}], m = 24/\tau$, $T_d = T/m$, $T_w = T/(m*7), \tau = \text{length of each time step}$. $\mathbf{h}_{T+1}^r, \mathbf{h}_{T+1}^d, \mathbf{h}_{T+1}^w \in \mathbb{R}^H$, $H = \text{hidden state dimension}$.
\vspace{2mm}

\subsection{Prediction}
\label{prediction}
After calculating the final hidden states of all three SAB-LSTMs we use location-based attention mechanism~\cite{DBLP:conf/emnlp/LuongPM15} to capture the contribution ($\alpha_a$) of the recent, daily and weekly trends. Then, a context vector is calculated using the generated attention weights. Here, $\mathbf{W_h} \in \mathbb{R}^{H\times A}, \mathbf{b_h} \in \mathbb{R}^A $ are learnable parameters, $A$ = attention dimension and $a=\{r, d, w\}$.
$$\alpha_a = \text{softmax}_{a} (\text{tanh}(\mathbf{W_h}^{T}\mathbf{h}_{T+1}^a + \mathbf{b_h})$$
\begin{equation}
\label{eq:4}
    \mathbf{c}_{i, T+1}^k = \sum_a \alpha_a \mathbf{h}_{T+1}^a
\end{equation}
Finally, the context vector is fed to a fully connected layer for predicting the crime occurrence at time step ($T+1$) for region $r_i$ and crime category $k$ where, $\mathbf{w} \in \mathbb{R}^{H}, \mathbf{b} \in \mathbb{R}$ are learnable parameters. We add the previously omitted region-index $i$ and crime category-index $k$ below. 
\begin{equation}
\label{eq:5}
 \hat{y}_{i, T+1}^k = \text{tanh}(\mathbf{w}\mathbf{c}_{i, T+1}^k  + b)
\end{equation}
Figure~\ref{Fig:overview}d gives an overview of generating the context vector and prediction.
\section{Experiment} 
\label{Sec:experimentalResult} 

\subsection{Experimental Settings}
\subsubsection{Data-sets} 
We evaluate our model on publicly available 2019 Chicago crime data~\cite{chicago_crime2019}, following the state-of the-art~\cite{DBLP:conf/kdd/WangKGL16, DBLP:journals/tbd/WangYKGL19}. Chicago is one of the most violent cities of United States and the crime concentration of Chicago is very diverse; it has both some of the safest and some of the most crime prone neighborhoods. We use 2019 Chicago taxi trip data~\cite{chicago_taxi2019} and POI information as external features. We collect POI information from FourSquare API while Chicago crime and taxi data are publicly available.   
\begin{itemize}
    \item \textbf{Chicago-Crime} (2019). We select \num[group-separator={,}]{152720} crime records of four crime  categories: theft, criminal damage, battery, narcotics from $1/1/2019$ to $31/12/2019$ and extract these information of each record: timestamp, primary category of crime, community area where it occurred.
    \item \textbf{Chicago-POI} We select \num[group-separator={,}]{89324} POIs of $10$ categories: food, residence, travel, arts \& entertainment, outdoors \& recreation, education, nightlife, professional, shops and event.
    
    \item \textbf{Chicago-Taxi} (2019). We select \num[group-separator={,}]{29110097} taxi trips from $1/1/2019$ to $31/12/2019$ and extract these information of each record: pickup timestamp, drop-off timestamp, pickup community area, drop-off community area.
\end{itemize}

\subsubsection{Data Preprocessing} We consider Chicago crime data of first 8 months as training set and 10\% and 90\% of the remaining last 4 months as validation set and test set, respectively. We use taxi inflow (F1), outflow (F2) and POI category: food (F3), residence (F4), travel (F5), arts \& entertainment (F6), outdoors \& recreation (F7), education (F8), nightlife (F9), professional (F10), shops (F11) and event (F12) as external features. We use Min-Max normalization to scale the crime events to [-1, 1] and later denormalize the prediction to get the actual number of crime events. Following~\cite{DBLP:journals/tbd/WangYKGL19}, we do not scale the external features. 

\subsubsection{Parameter Settings} We optimize the hyperparameters of AIST using a grid search strategy. The search space for every hyperparameter is presented in Table~\ref{table:exp_hyperparameter} and the selected value in the search space is shown in bold. For simplicity, we use the same parameter settings across all crime categories and regions. AIST is trained using the Adam optimizer with batch size = 42 and initial learning rate = 0.001. We set the duration of each time step, $\tau = 4$ hours. We set the number of recent ($T$), daily ($T_d$) and weekly ($T_w$) time steps to 20, 20 and 3, respectively.

\emph{hGAT \& fGAT settings.} Both hGAT and fGAT are single layer GATs consisting of single attention head for computational efficiency. 
We set the output size ($F$) of both hGAT and fGAT to 8. For fGAT, we set $d_q, d_k=40$. Dropout with $p=0.5$ is applied to  unnormalized node-level ($e_{ii'}^c$), unnormalized parent-level ($e_{ii'}^p$), normalized combined attention weights ($\alpha_{ii'}$) in hGAT, and dropout with $p=0.5$ is applied to normalized dot-product attention weights ($\beta_{ii'}^j$) in fGAT. 

\emph{SAB-LSTM settings.} All 3 SAB-LSTMs are single layered with hidden dimension $H=40$. For $SAB-LSTM_{r}$ and $SAB-LSTM_{d}$, we set $k_{att}=5, k_{top}=5, trunc_{length} = 5$. For $SAB-LSTM_w$ we set $k_{att}=1, k_{top}=5, trunc_{length} = 1$. Dropout with $p=0.2$ is applied to each output of 3 SAB-LSTMs. Finally, we set the attention dimension of location based attention as $A=30$.
\begin{small}
\begin{table*}[htbp]
	\caption{Hyperparameter Settings of AIST}
	\centering
		\begin{tabular}[t]{|l|l|c|}
			\hline
			Hyperparameters & Search Space\\
			\hline
			
			Number of recent time step ($T$) & [16, \textbf{20}, 24, 28]\\
			\hline
			
			Number of daily time step ($T_d$) & [12, 16, \textbf{20}, 24] \\
			\hline
			
			Number of weekly time step ($T_w$) & [2, \textbf{3}, 4, 5]\\
			\hline
			
			Output size of both hGAT and fGAT ($F$) & [6, \textbf{8}, 10, 12]\\
			\hline
			
			Dimension of query and key vector of fGAT ($d_q, d_v$) & [36, \textbf{40}, 44, 48]\\
			\hline
			
			Dimension of hidden states of SAB-LSTM ($H$) & [24, 32, \textbf{40}, 48] \\
			\hline
			
			Dimension of Location Attention ($H$) & [22, \textbf{30}, 38, 46]\\
			\hline
		\end{tabular}
	\label{table:exp_hyperparameter}
\end{table*}
\end{small}

\subsubsection{Evaluation Criteria}
We use mean average error (MAE) and mean square error (MSE) to evaluate AIST predictions. Here, $n$ represents the number of predictions, $y_i$ represents the predicted result and $\hat{y}_i$ represents the ground truth. 
\begin{align*}
\text{MAE} = \frac{1}{n}\sum_{i=1}^{n}|y_i - \hat{y}_i| \qquad
\text{MSE} = \frac{1}{n}\sum_{i=1}^{n}(y_i - \hat{y}_i)^2  \end{align*}
We also use Total Variation Distance (TVD) for comparing prediction scores and Jensen-Shannon Divergence (JSD) for comparing attention weight distributions to evaluate the interpretations of AIST, where $\alpha = \frac{\alpha_1 + \alpha_1}{2}$ and $\text{KL}(p || q)$ calculates the Kullback–Leibler divergence between probability distributions $p$ and $q$.
$$\text{TVD} = \frac{1}{2}\sum_{i=1}^{|\textbf{y}|}(|\hat{y}_{1i} - \hat{y}_{2i}|)$$
$$\text{JSD}(\alpha_1, \alpha_2) = \frac{1}{2} \text{KL}[\alpha_1 || \alpha] +  \frac{1}{2} \text{KL}[\alpha_2 || \alpha]$$
\subsubsection{Baselines} 
We compare AIST with the following baselines.
\begin{compactitem}
    \item ARIMA~\cite{4666600}. The most general case of models for predicting time series combining moving average and auto-regression.
    
    \item DTR~\cite{DBLP:books/wa/BreimanFOS84}. A decision tree algorithm for regression that chooses the best random split while partitioning samples in multiple subsets.
    \item Att-RNN~\cite{DBLP:journals/corr/BahdanauCB14}. It uses attention mechanism with RNN to capture the temporal correlation.
    
    \item DeepCrime~\cite{DBLP:conf/cikm/HuangZZC18}. A hierarchical recurrent framework that encodes the temporal correlation and inter-dependencies between crimes and urban anomalies.
    
    \item MiST~\cite{DBLP:conf/www/HuangZZWCY19}. It uses multiple LSTMs to encode the spatial, temporal and categorical views of crime.
    
    \item GeoMAN*. A multi-level attention network with a sequential encoder-decoder architecture that models both spatial and temporal correlation, customized to predict crimes. Unlike GeoMAN~\cite{DBLP:conf/ijcai/LiangKZYZ18}, while calculating the spatial dependencies we only consider those who share a common boundary rather than considering all the regions in the network. 
    
    
    \item STGCN~\cite{DBLP:conf/ijcai/YuYZ18}. A graph convolutional layer is placed between two gated temporal convolution layers to model the spatio-temporal correlation.
    
    \item MVGCN*. A GNN architecture with multiple graph convolution and fully connected layers to process different temporal and semantic views, respectively. MVGCN~\cite{DBLP:journals/TKDE/9139357} is customized by only considering those regions that share a common boundary for predicting crimes.
\end{compactitem}


\begin{small}
\begin{table*}[t]
	\caption{Comparison of AIST with baselines on Chicago Crime Data (2019)}
	\centering
		\begin{tabular}[t]{|c|c|c|c|c|c|}
			\hline
			Model & Criteria & Theft (C1) & 
			\begin{tabular}{@{}c@{}}Criminal \\ Damage (C2)\end{tabular}
			& Battery (C3) & Narcotics (C4) \\
			\hline
			
			\multirow{2}{*}{ARIMA~\cite{4666600}} 
			& MAE & 1.2010 & 0.5863 & 0.8840 & 0.5705\\
			& MSE & 2.8492 & 0.7238 & 1.4242 & 0.7928 \\
			\hline
			
			\multirow{2}{*}{DTR~\cite{DBLP:books/wa/BreimanFOS84}} 
			& MAE & 1.1943 & 0.5590 & 0.8983 & 0.4901\\
			& MSE & 3.3275 & 0.8123 & 1.9336 & 0.8522\\
			\hline
				
			\multirow{2}{*}{Att-RNN~\cite{DBLP:journals/corr/BahdanauCB14}} 
			& MAE & 1.0419 & 0.4096 & 0.7377 & 0.4128\\
			& MSE & 2.5443 & 0.4427 & 1.0665 & 0.6380\\
			\hline
			
			\multirow{2}{*}{DeepCrime~\cite{DBLP:conf/cikm/HuangZZC18}} 
			& MAE & 1.0022 & 0.3727  & 0.7271 & 0.3702 \\
			& MSE & 2.6279 & 0.4751 & 1.0567 & 0.6394\\
			\hline
			
			\multirow{2}{*}{MiST~\cite{DBLP:conf/www/HuangZZWCY19}} 
			& MAE & 1.0241 & 0.3727 & 0.7365 & 0.3701\\
			& MSE & 2.5153 & 0.4836 & 1.0345 & 0.6495\\
			\hline

			\multirow{2}{*} 
			{\begin{tabular}{@{}c@{}}customized \\GeoMAN~\cite{DBLP:conf/ijcai/LiangKZYZ18}\end{tabular}}
			& MAE & 0.9092 & 0.3876 & 0.7226 & 0.3450\\
			& MSE & 1.9930 & \textbf{0.4385} & 0.9871 & \textbf{0.5595}\\
			\hline
			
			\multirow{2}{*}{STGCN~\cite{DBLP:conf/ijcai/YuYZ18}} 
			& MAE & 1.0416 & 0.5130 & 1.0869 & 0.3886\\
			& MSE & 1.8121 & 0.4860 & 1.0595 & 0.6342\\
			\hline
			
			\multirow{2}{*}
			{\begin{tabular}{@{}c@{}}customized \\MVGCN~\cite{DBLP:journals/TKDE/9139357}\end{tabular}}
			& MAE & 1.5244 & 0.4641 & 0.7928 & 0.4093 \\
			& MSE & 4.2593 & 0.7019 & 1.1949 & 0.8337\\
			\hline
			
			\multirow{2}{*}{AIST} 
			& MAE & \textbf{0.8747} & \textbf{0.3615} & \textbf{0.6910} & \textbf{0.3399}\\
			& MSE & \textbf{1.6986} & 0.4837 & \textbf{0.9568} & 0.5609\\
			\hline
			
		\end{tabular}
	\label{table:exp_result}
	\vspace{-1em}
\end{table*}
\end{small}

			
				
			
			


\subsection{Prediction Performance}
\label{SubSec:pred} 
\subsubsection{Comparison with baselines. } 
The performance of the baselines and AIST is shown in Table~\ref{table:exp_result}. The baselines include both high-performance crime prediction models (e.g., DeepCrime, Mist) and high-performance spatio-temporal prediction models (e.g., customized GeoMAN, STGCN, customized MVGCN). All the baselines have been tuned optimally to produce the best prediction result. In general, AIST outperforms all baselines by achieving the lowest MAE and MSE scores across all crime categories (except the MSE scores for crime category Criminal Damage (C2) and Narcotics (C4)).

\begin{itemize}[leftmargin=*]

\item AIST learns the time varying spatial dependencies, diverse temporal correlation and crime relevant dynamic context to perform crime prediction tasks. Other competing deep learning models such as MVGCN*, STGCN, GeoMAN*,  MiST, Att-RNN do not learn the crime and region specific relevant context; DeepCrime, Att-RNN do not consider the spatial correlation and MVGCN* do not consider the temporal correlation. As a result, in general these models fail to perform better than AIST for crime prediction tasks.

\item Aside from AIST, GeoMAN* has the second best MAE and MSE scores across all crime categories. On top of that, it has better MSE scores than AIST for category Criminal Damage (C2) and Narcotics (C4). C2 lacks periodical temporal properties and the surrounding context has less influence on C4. Since, GeoMAN* does not consider daily or weekly temporal properties and the influence of region specific context on a crime category, it performs better than AIST for these crime categories in terms of MSE scores. 

\item Other attention-based neural network architectures:  DeepCrime and MiST perform considerably worse than GeoMAN* because they do not consider the spatial correlation and external features, respectively. Though DeepCrime and MiST have similar MAE scores, MiST is better than DeepCrime in terms of the MSE score. This is because DeepCrime only captures the temporal correlation and region-category dependencies, whereas MiST captures both spatio-temporal and cross-categorical correlation of crimes. 

\item STGCN has the third best MSE score across all crime categories. Only AIST and GeoMAN* have better MSE score than STGCN. Having a lower MSE score than those of other attention-based deep learning models such as Att-RNN, DeepCrime and MiST, STGCN is more likely to capture the sudden changes in crime distribution. However, in terms of the MAE score STGCN shows a poor performance in comparison with AIST and others such as GeoMAN*, MiST, DeepCrime, and Att-RNN. 

\item Att-RNN, a plain recurrent neural network architecture that only considers the recent temporal correlation is behind DeepCrime and MiST, but above STGCN in terms of the performance based on the MAE score. However, in terms of the MSE score its performance is quite similar to DeepCrime. For category Criminal Damage (C2), Att-RNN performs better than AIST supporting our claim that periodical information is not helpful for predicting this category of crime.
  
\item MVGCN*, despite being a top performing architecture in crowd flows prediction, performs worst among the competing deep learning models in the crime prediction task based on MSE scores. Because of the sparsity of crime distributions, careful exploration of available crime data and context are an absolute necessity for capturing the sudden change in the crime distribution. However, the large MSE scores of MVGCN* mean that the graph convolution and fully connected layers in MVGCN* used to model the spatial correlation and the influence of the external features lack the ability to do so. Hence, it fails to compete with others.

  
\item Traditional time series analysis and machine learning methods such as ARIMA and DTR though interpretable, lack the ability to model the non-linear and complex crime patterns. This is because ARIMA only considers a fixed temporal pattern, whereas DTR does not consider the temporal properties of crime at all. Hence, in general they show poor performance than the deep learning models across all crime categories.

\end{itemize}

\subsubsection{Effectiveness of different spatial components of AIST} 
\begin{figure}[tbp]
	\begin{subfigure}{0.245\columnwidth}
	    \def\svgwidth{\columnwidth}
	    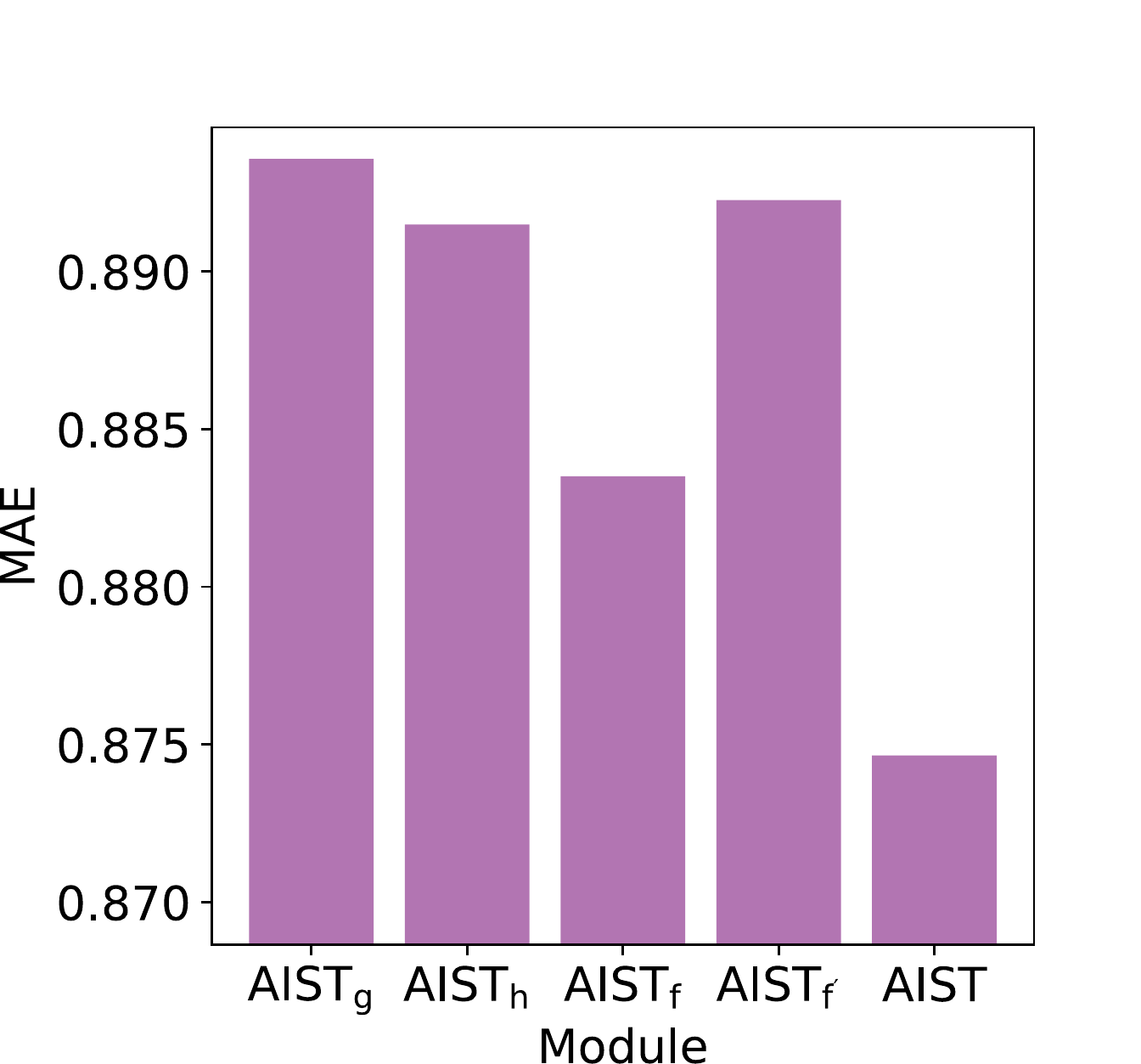
		\caption{C1: Theft}
		\label{Fig:eval_model1}
	\end{subfigure}
	\begin{subfigure}{0.245\columnwidth}
	    \def\svgwidth{\columnwidth}
	    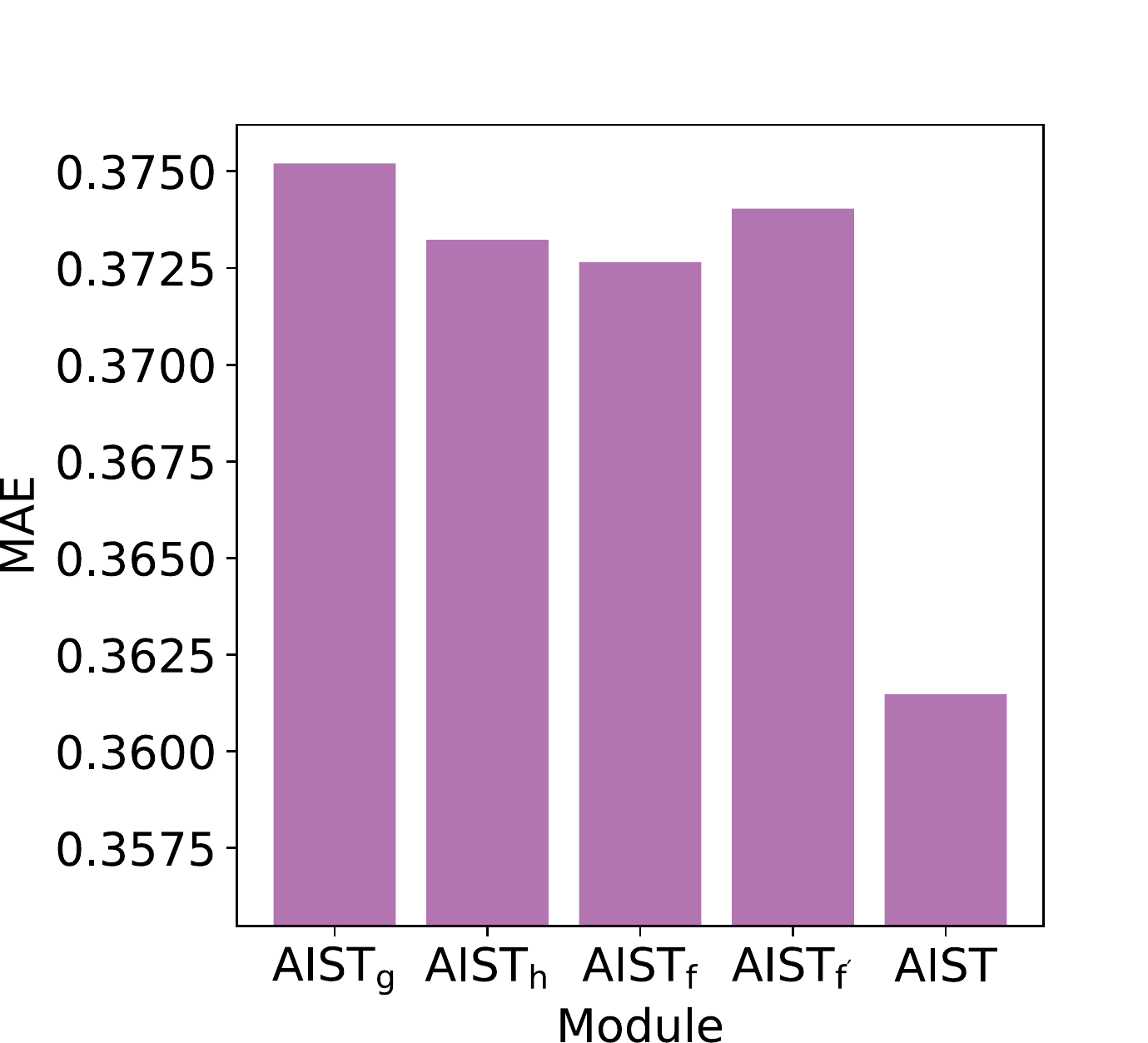
		\caption{C2: Criminal Damage}
		\label{Fig:eval_model2}
	\end{subfigure}
	\begin{subfigure}{0.245\columnwidth}
	    \def\svgwidth{\columnwidth}
	    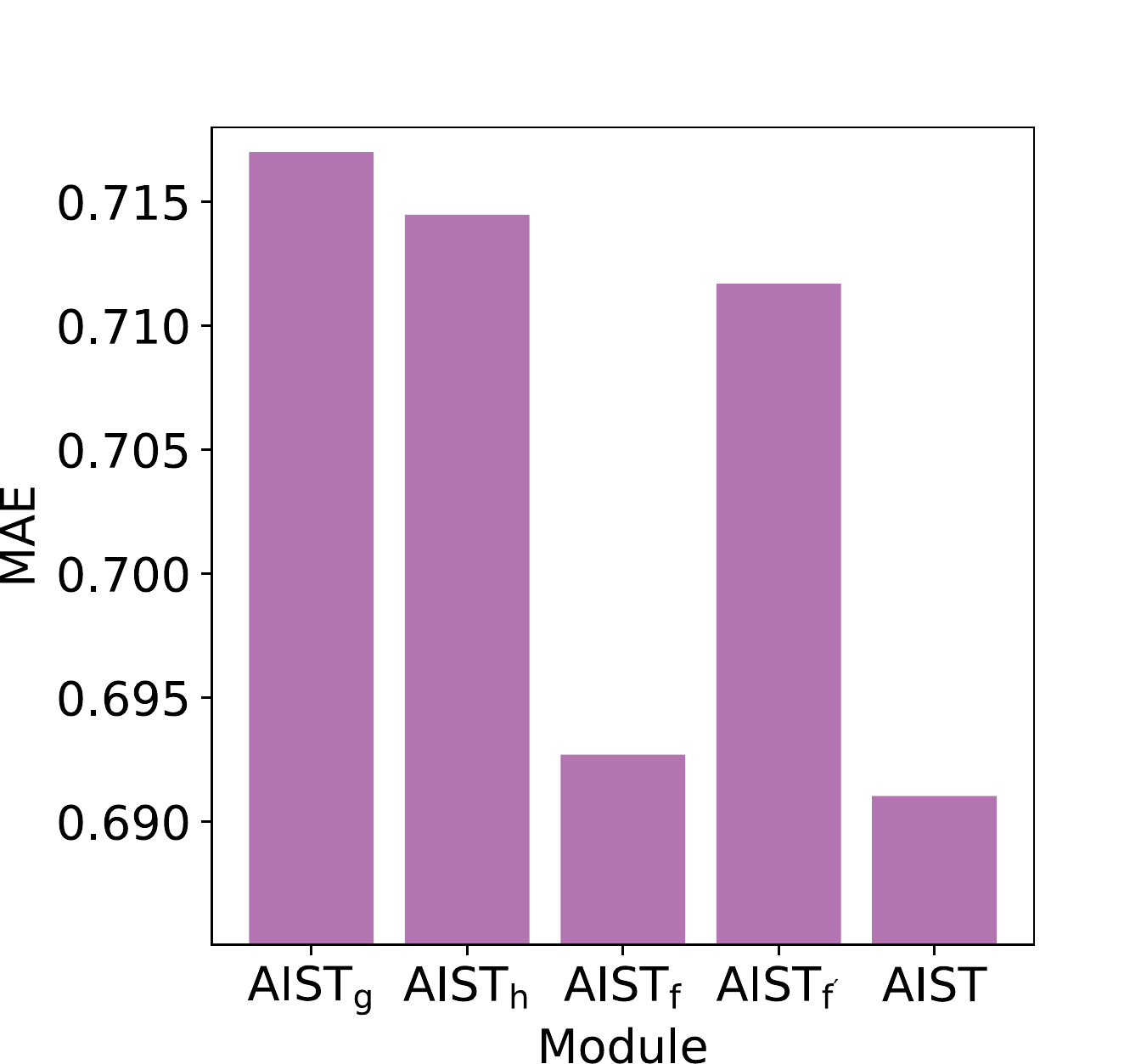
		\caption{C3: Battery}
		\label{Fig:eval_model3}
	\end{subfigure}
	\begin{subfigure}{0.245\columnwidth}
	    \def\svgwidth{\columnwidth}
	    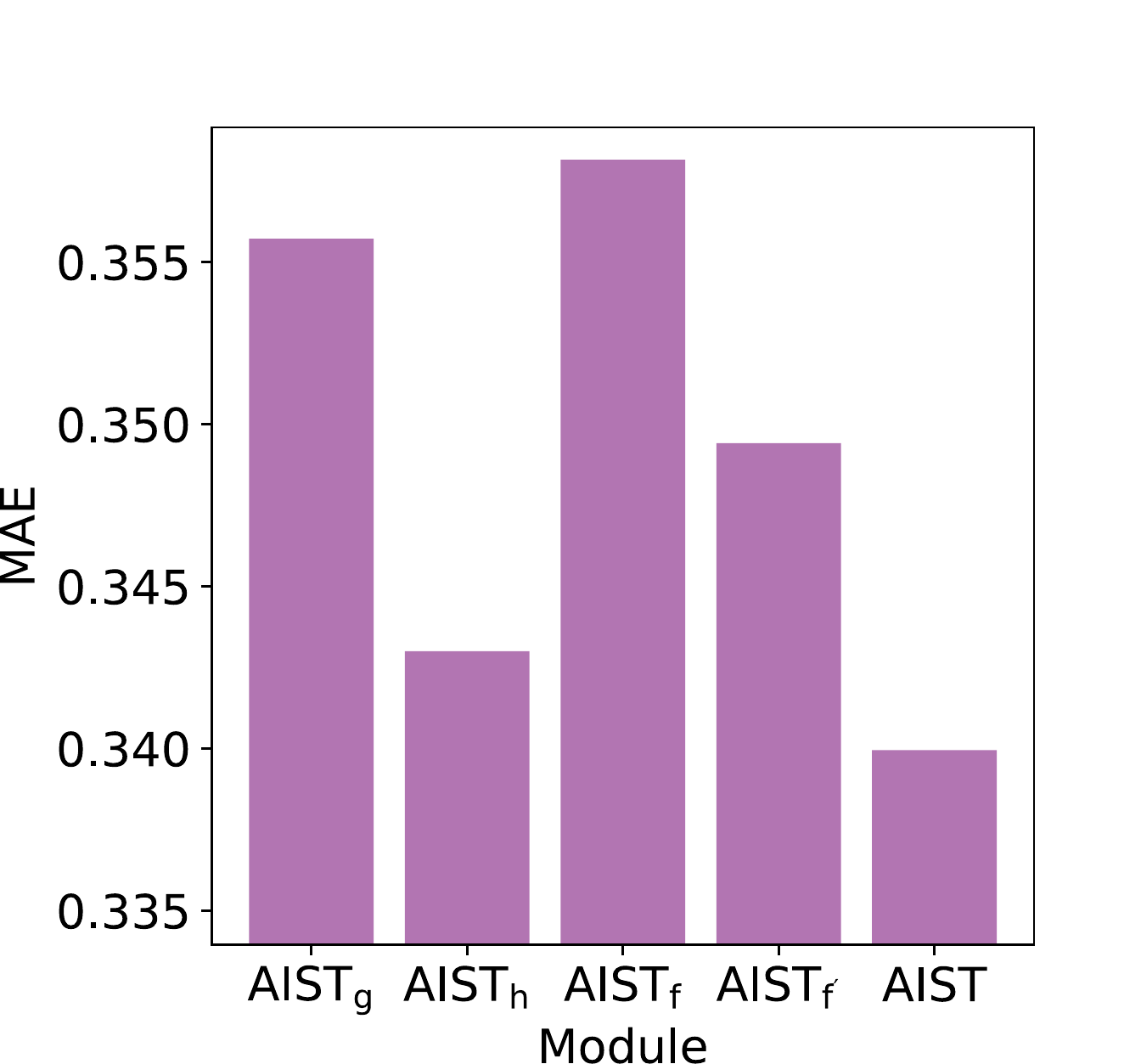
		\caption{C4: Narcotics}
		\label{Fig:eval_model7}
	\end{subfigure}
	\caption{Effect of different spatial components}
	\label{Fig:eval_self}
	\vspace{-1em}
\end{figure}

We consider the following variants of AIST to understand the influence of different spatial components on its prediction performance. Figure~\ref{Fig:eval_self} shows a comparative performance analysis of these spatial variants of AIST. 

\begin{itemize}
    \item $\text{AIST}_{g}.$ hGAT is replaced with traditional graph attention networks to learn crime embedding; fGAT is omitted from the spatial module. 
    
    \item $\text{AIST}_{h}.$ hGAT is used to learn crime embedding; fGAT is omitted from the spatial module. 
    
    \item $\text{AIST}_f.$ hGAT is used to learn crime embedding and the crime embedding is concatenated with the external features to produce the final spatial embedding; fGAT is omitted from the spatial module.
    
    \item $\text{AIST}_{f'}.$ GAT (in place of hGAT) and fGAT constitute the spatial module.
\end{itemize}

$\text{AIST}_h$ has a better MAE score across all crime categories than $\text{AIST}_g$ which justifies the selection of hGAT over GAT for learning the crime embedding. Specifically, from Fig~\ref{Fig:eval_model7} it is evident that hGAT learns a better crime embedding of category Narcotics compared to other categories, which indicates the existence of strong narcotics networks in certain parts of Chicago. Besides, the fact that AIST consistently performs better than $\text{AIST}_{f'}$ shows the superiority of the spatial embedding learned by hGAT alongside fGAT over $\text{AIST}_{f'}$.

For categories Theft (C1) and Battery (C3), the concatenation of external features improves the prediction performance of $\text{AIST}_h$, which indicates the impact of contextual information for the crime prediction task, specially for these categories. However, the prediction performance of $\text{AIST}_f$ deteriorates significantly for category Narcotics (C4) and remains almost same as $\text{AIST}_h$ for category Criminal Damage (C2), which suggest its inability to constantly differentiate the influential features from noise across all crime categories. The significant
performance improvement of AIST over $\text{AIST}_f$ in predicting crime events across all categories suggests that careful extraction and learning of crime relevant feature embedding by fGAT is a necessity while performing crime prediction tasks.

\subsubsection{Effectiveness of different temporal components of AIST}
\begin{figure}[htbp]
	\begin{subfigure}{0.245\columnwidth}
	    \def\svgwidth{\columnwidth}
	    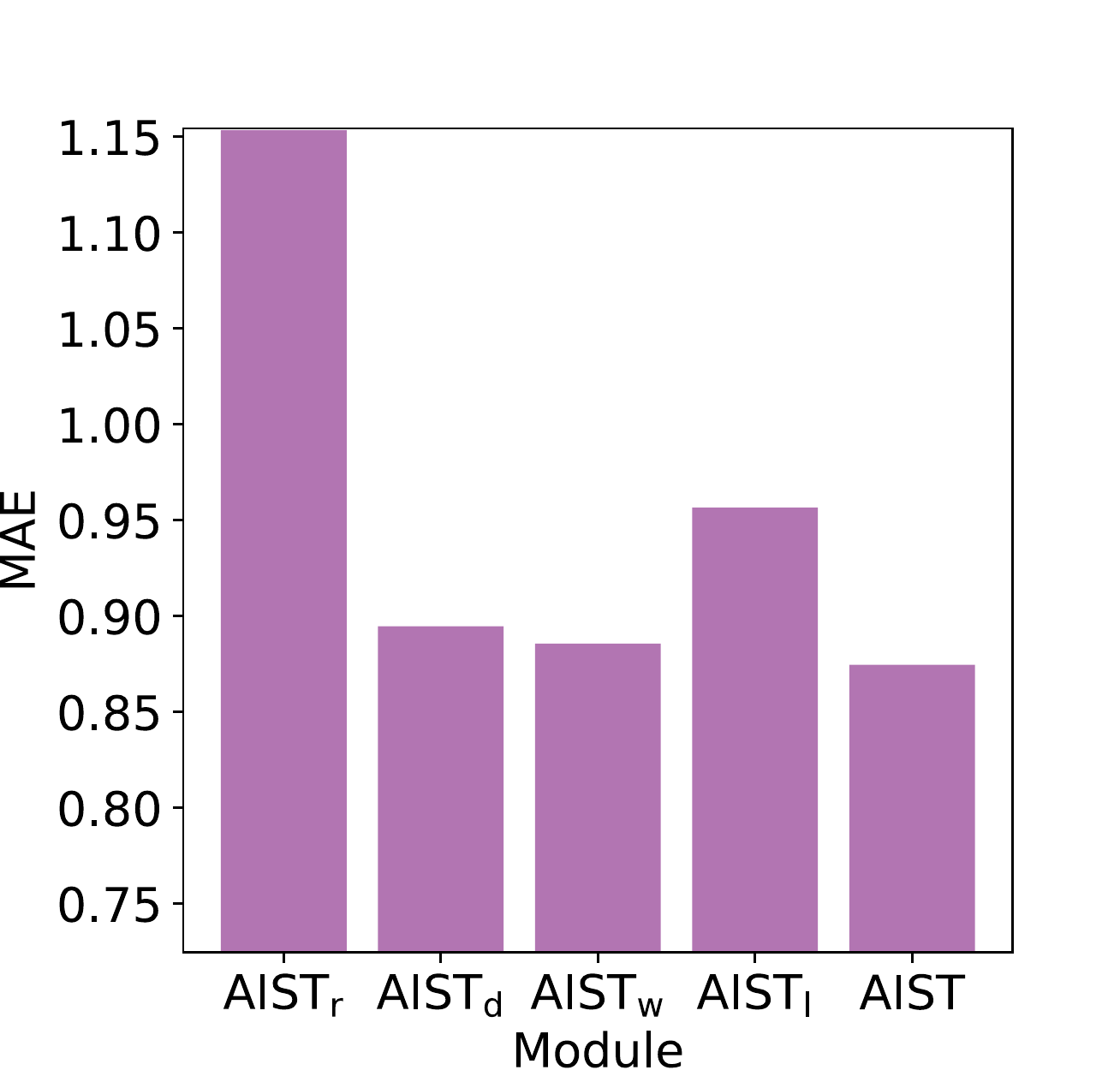
		\caption{C1: Theft}
		\label{Fig:tmodel1}
	\end{subfigure}
	\begin{subfigure}{0.245\columnwidth}
	    \def\svgwidth{\columnwidth}
	    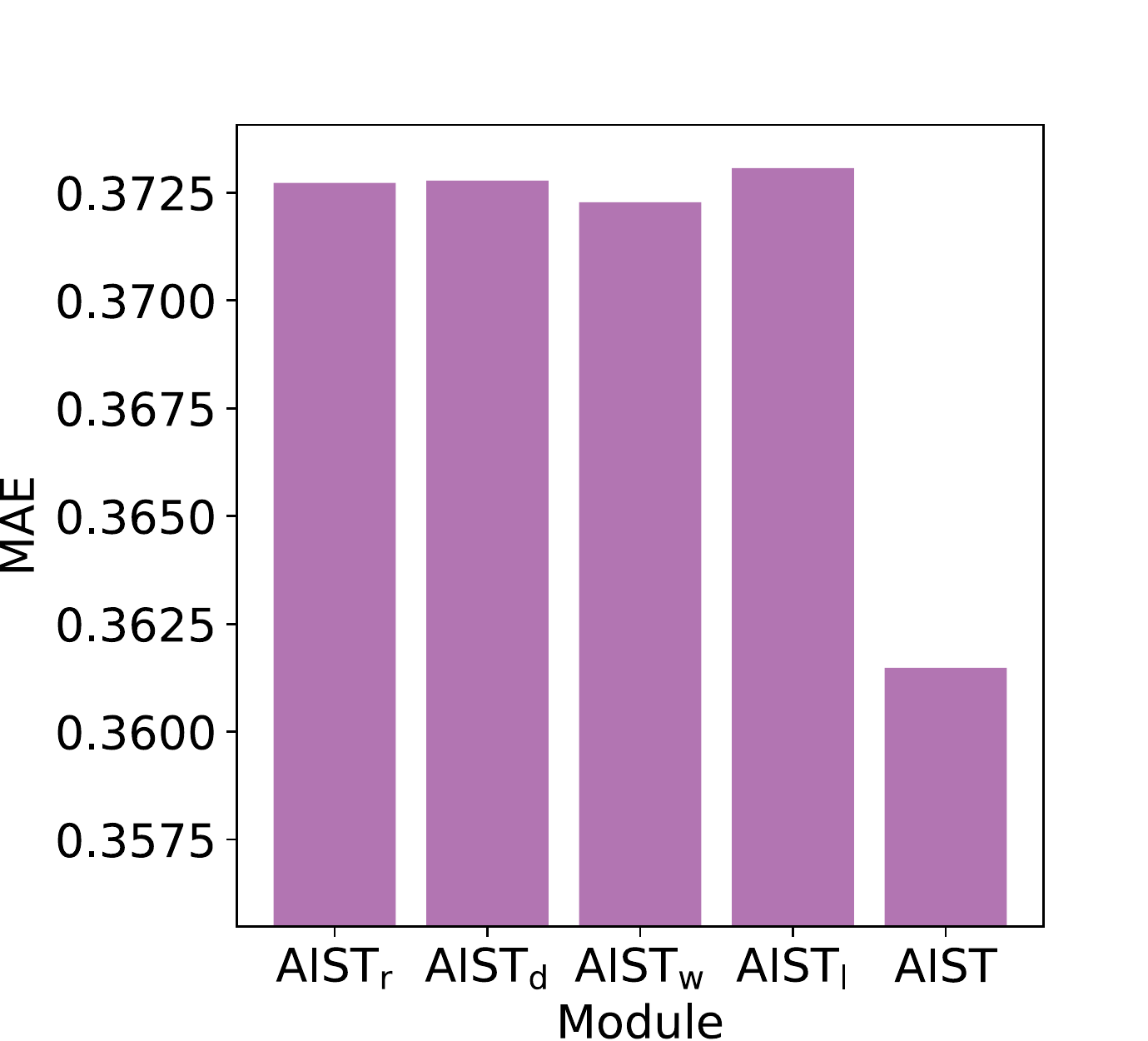
		\caption{C2: Criminal Damage}
		\label{Fig:tmodel2}
	\end{subfigure}
	\begin{subfigure}{0.245\columnwidth}
	    \def\svgwidth{\columnwidth}
	    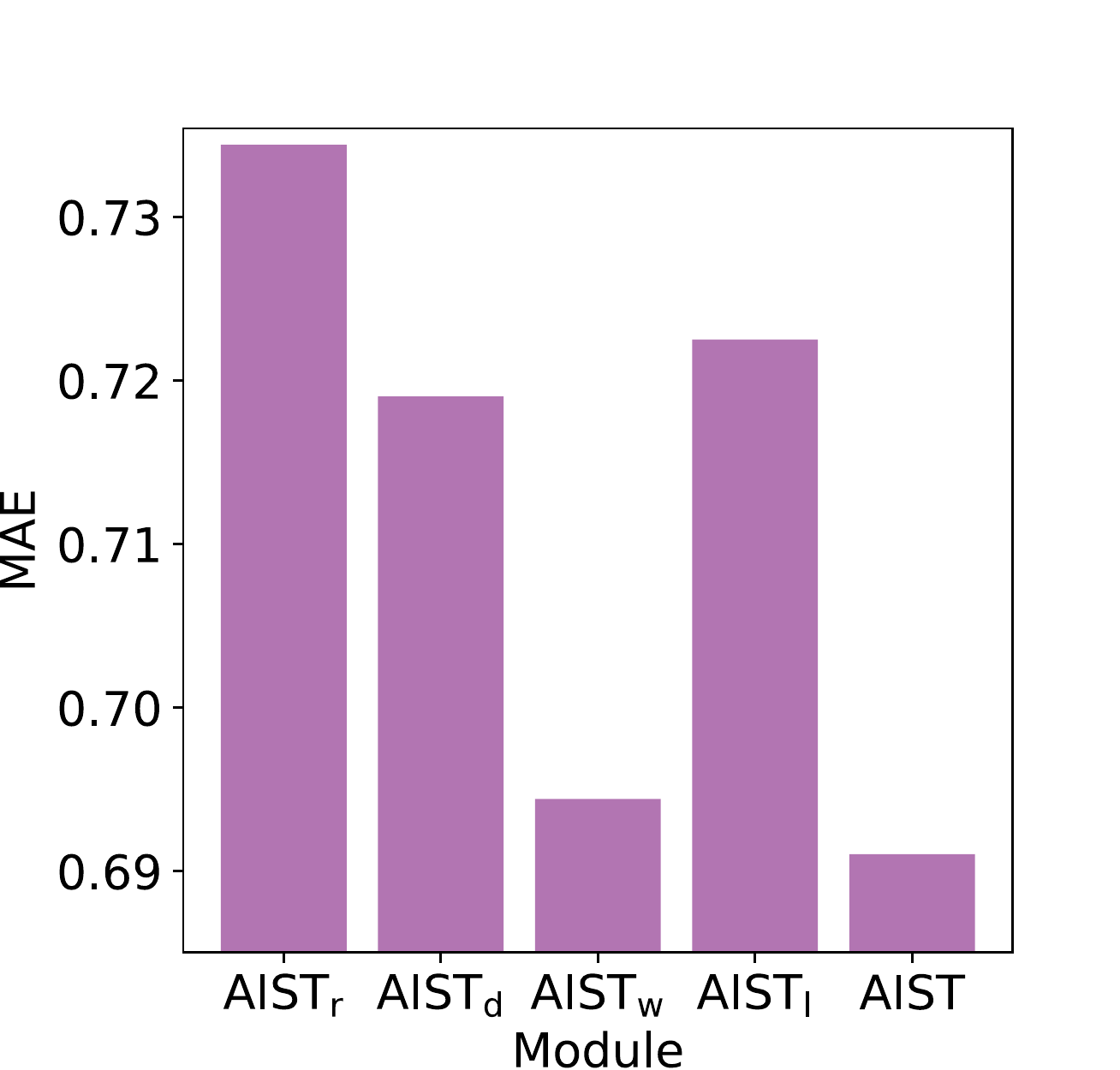
		\caption{C3: Battery}
		\label{Fig:tmodel3}
	\end{subfigure}
	\begin{subfigure}{0.245\columnwidth}
	    \def\svgwidth{\columnwidth}
	    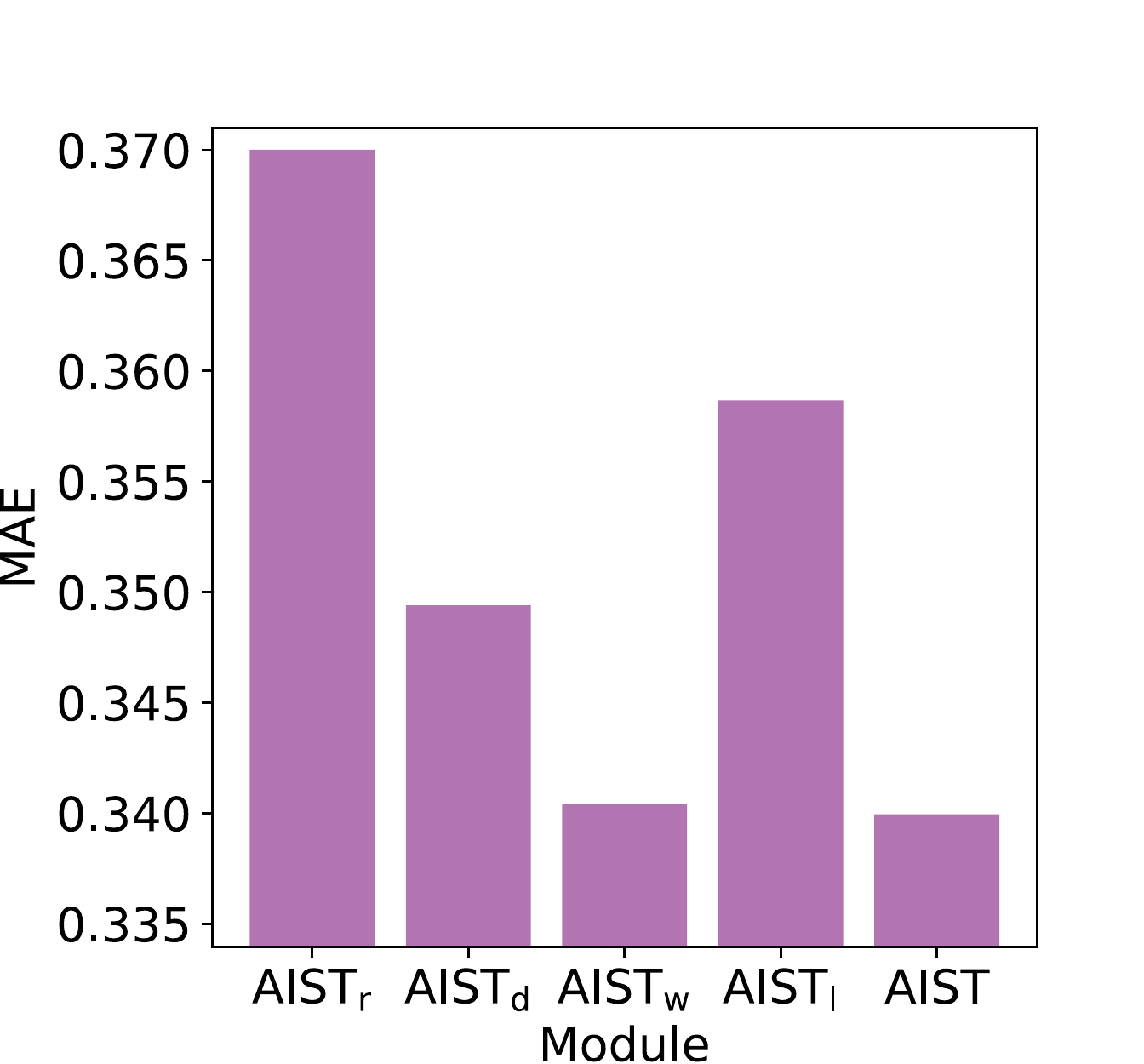
		\caption{C4: Narcotics}
		\label{Fig:tmodel7}
	\end{subfigure}
	\caption{Effect of different temporal components}
	\label{Fig:tmodel}
	\vspace{-1em}
\end{figure}

Figure~\ref{Fig:tmodel} shows a comparative performance analysis of different temporal components of AIST. 

\begin{itemize}
    \item $\text{AIST}_{r}.$ $\text{SAB-LSTM}_{d}$ and $\text{SAB-LSTM}_{w}$ are omitted from the temporal module.
    
    \item $\text{AIST}_{d}.$ $\text{SAB-LSTM}_{w}$ is omitted from the temporal module.
    
    \item $\text{AIST}_{w}.$ $\text{SAB-LSTM}_{d}$ is omitted from the temporal module.
    
    \item $\text{AIST}_{l}.$ All three $\text{SAB-LSTMs}$ are replaced by traditional LSTMs.
\end{itemize}

The significant decrease in the MAE scores of $\text{AIST}_{d}$ and $\text{AIST}_{w}$ over $\text{AIST}_{r}$ for Theft (C1), Battery (C2) and Narcotics (C4) suggest that both daily and weekly trends are instrumental in crime prediction tasks. Between these two, $\text{AIST}_{w}$ has better MAE scores over $\text{AIST}_{d}$ across all crime categories indicating the dominance of weekly trends over daily trends. Above all, the better MAE scores of AIST over $\text{AIST}_{l}$ for all crime categories justify the selection of SAB-LSTMs over traditional LSTMs for the crime prediction tasks.

Contrary to the general observation discussed above, $\text{AIST}_{r}, \text{AIST}_{d}, \text{and AIST}_{w}$ show similar performance for category Criminal Damage (C2) (Figure~\ref{Fig:tmodel2}), which suggests that it does not follow any daily or weekly trend. 

\subsubsection{Effect of Parameters}
\begin{figure}[htbp]
	\begin{subfigure}{0.245\columnwidth}
	    \def\svgwidth{\columnwidth}
	    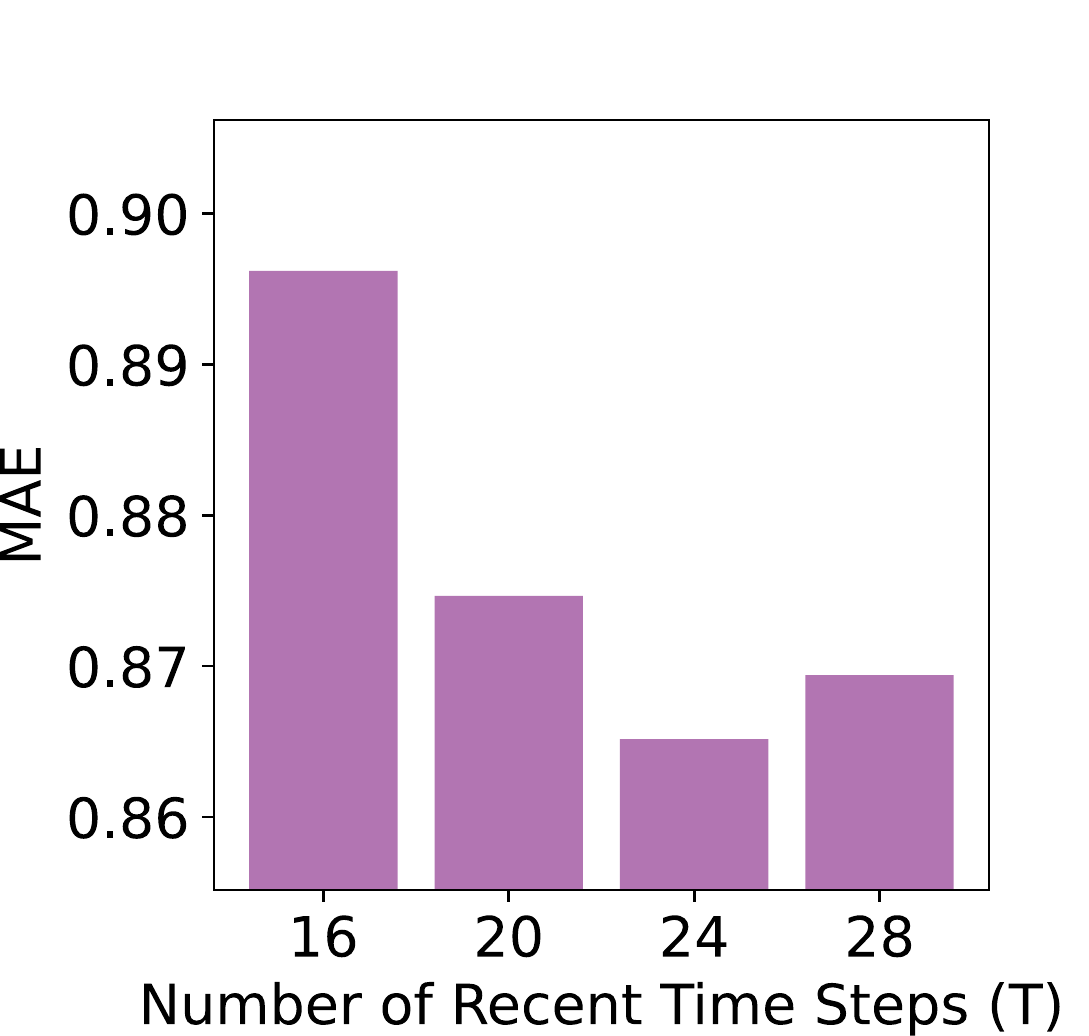
		\caption{C1: Theft}
		\label{Fig:param_T1}
	\end{subfigure}
	\begin{subfigure}{0.245\columnwidth}
	    \def\svgwidth{\columnwidth}
	    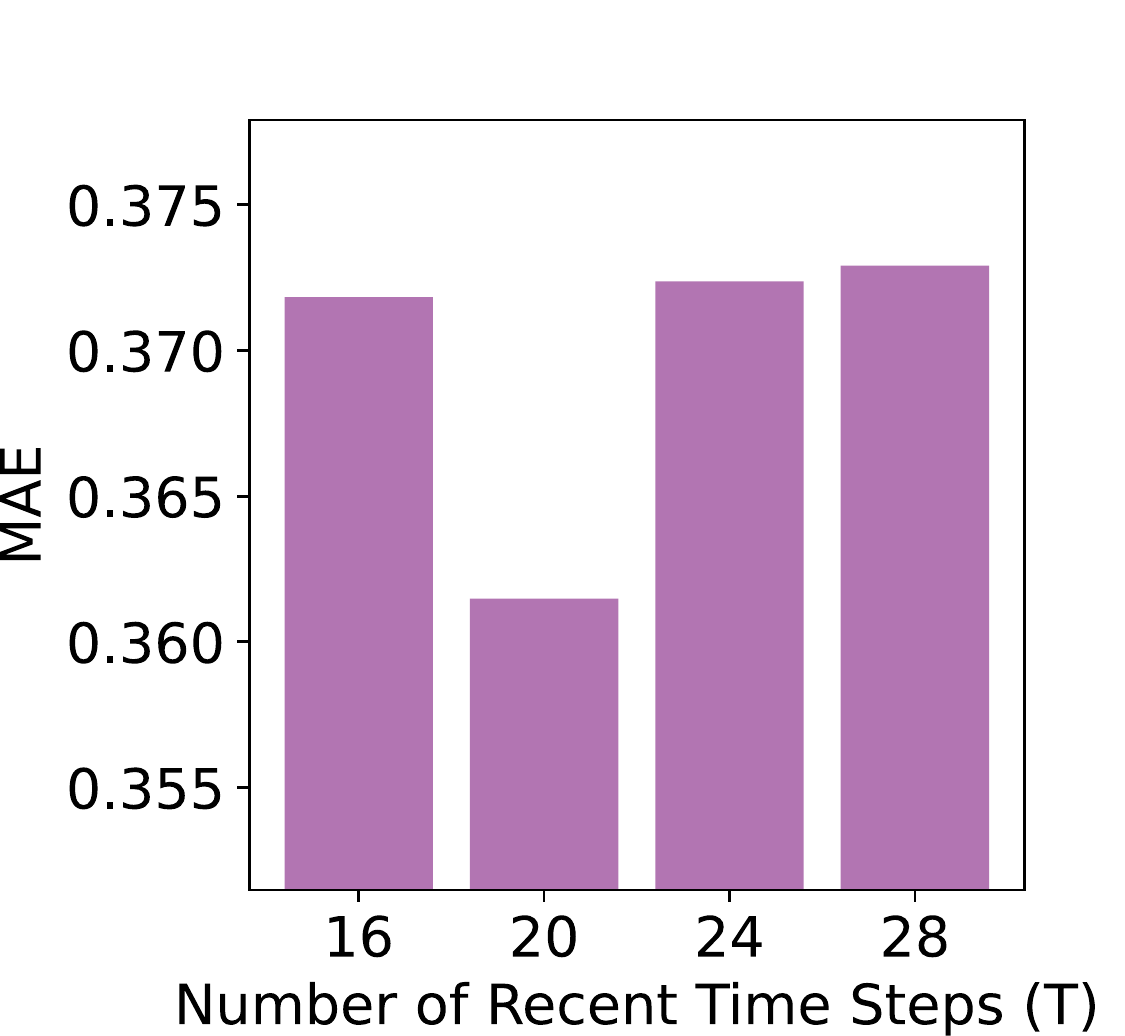
		\caption{C2: Criminal Damage}
		\label{Fig:param_T2}
	\end{subfigure}
	\begin{subfigure}{0.245\columnwidth}
	    \def\svgwidth{\columnwidth}
	    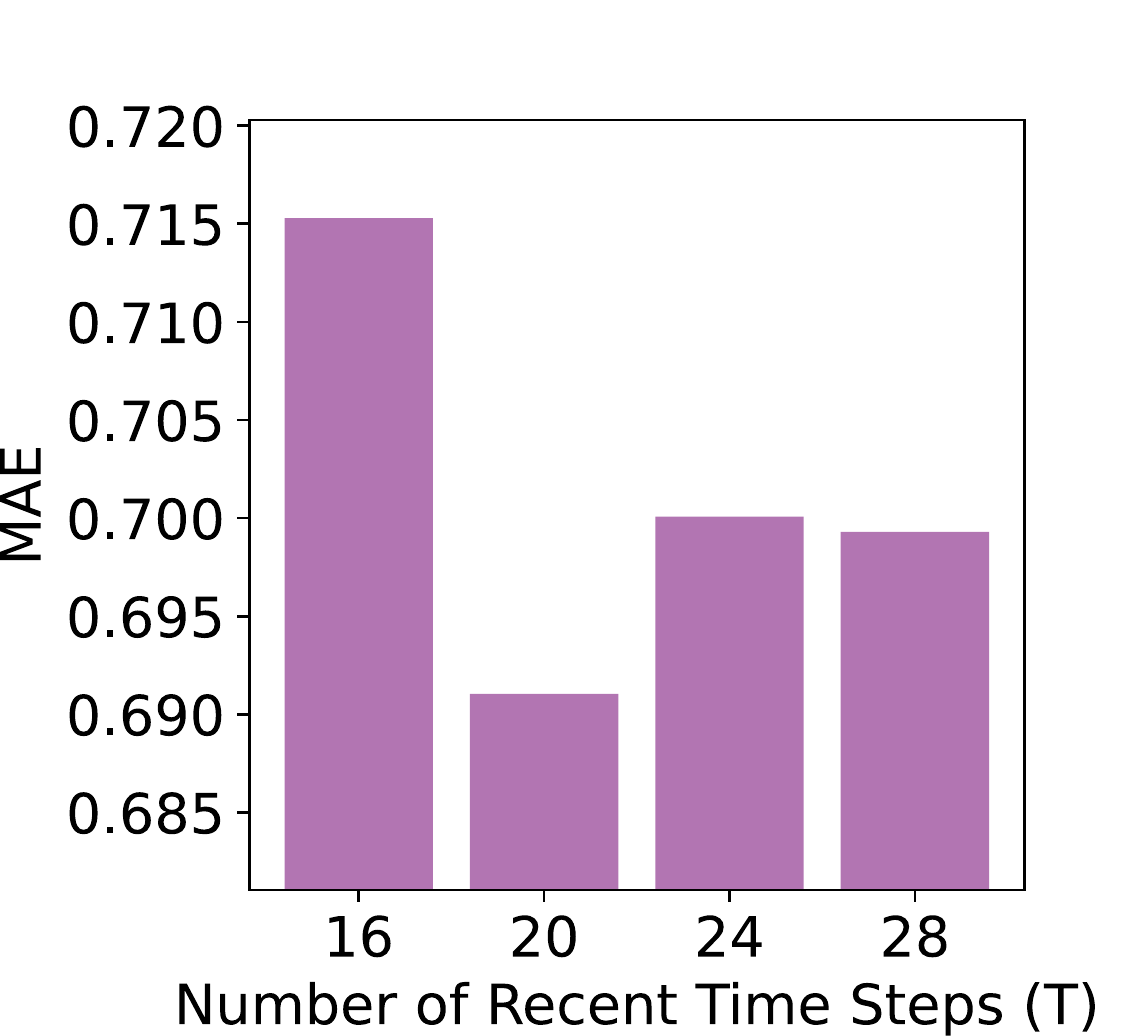
		\caption{C3: Battery}
		\label{Fig:param_T3}
	\end{subfigure}
	\begin{subfigure}{0.245\columnwidth}
	    \def\svgwidth{\columnwidth}
	    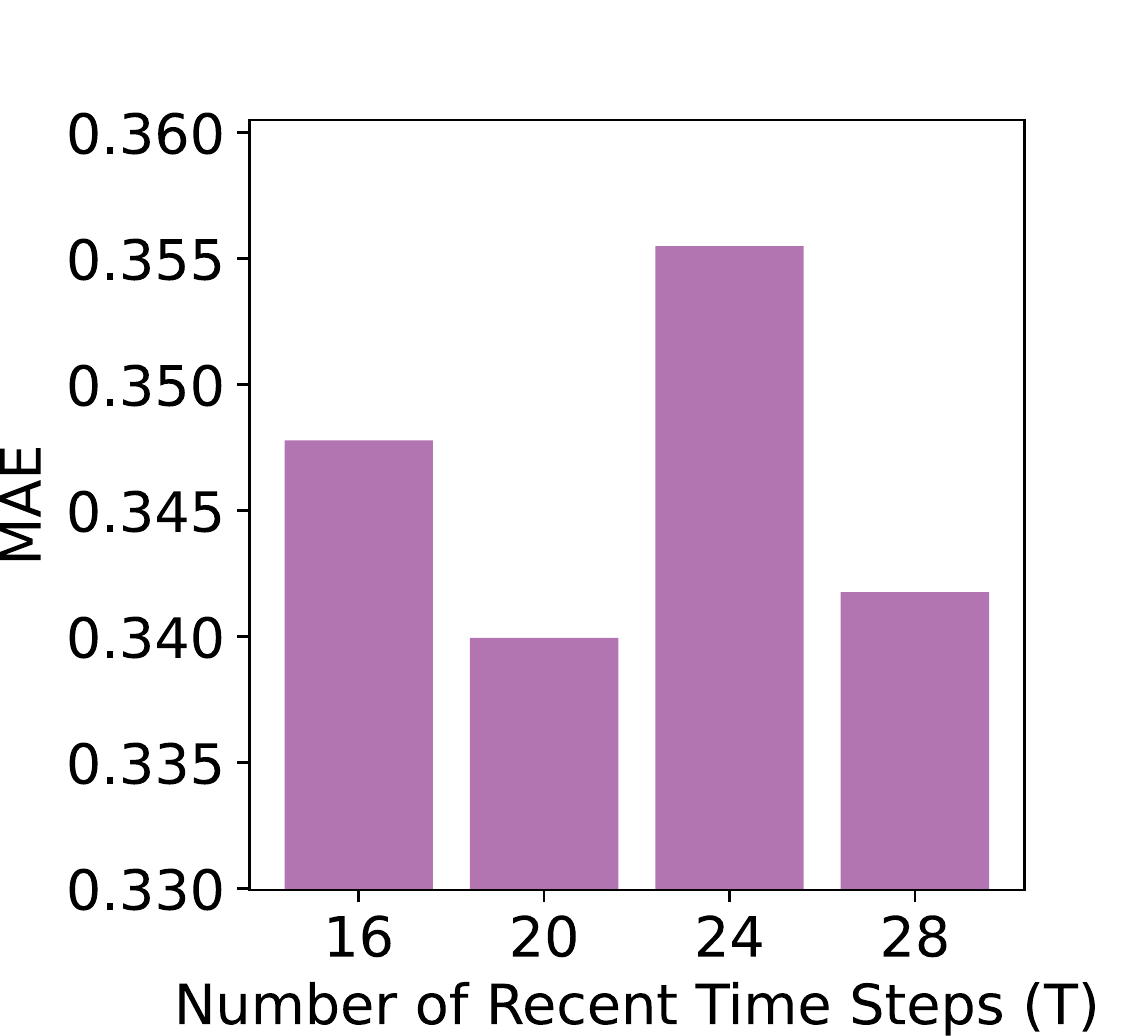
		\caption{C4: Narcotics}
		\label{Fig:param_T7}
	\end{subfigure}
	\caption{Effect of the Number of Recent Time Steps, $T$}
	\vspace{-1em}
	\label{Fig:param_T}
\end{figure}

\begin{figure}[htbp]
	\begin{subfigure}{0.245\columnwidth}
	    \def\svgwidth{\columnwidth}
	    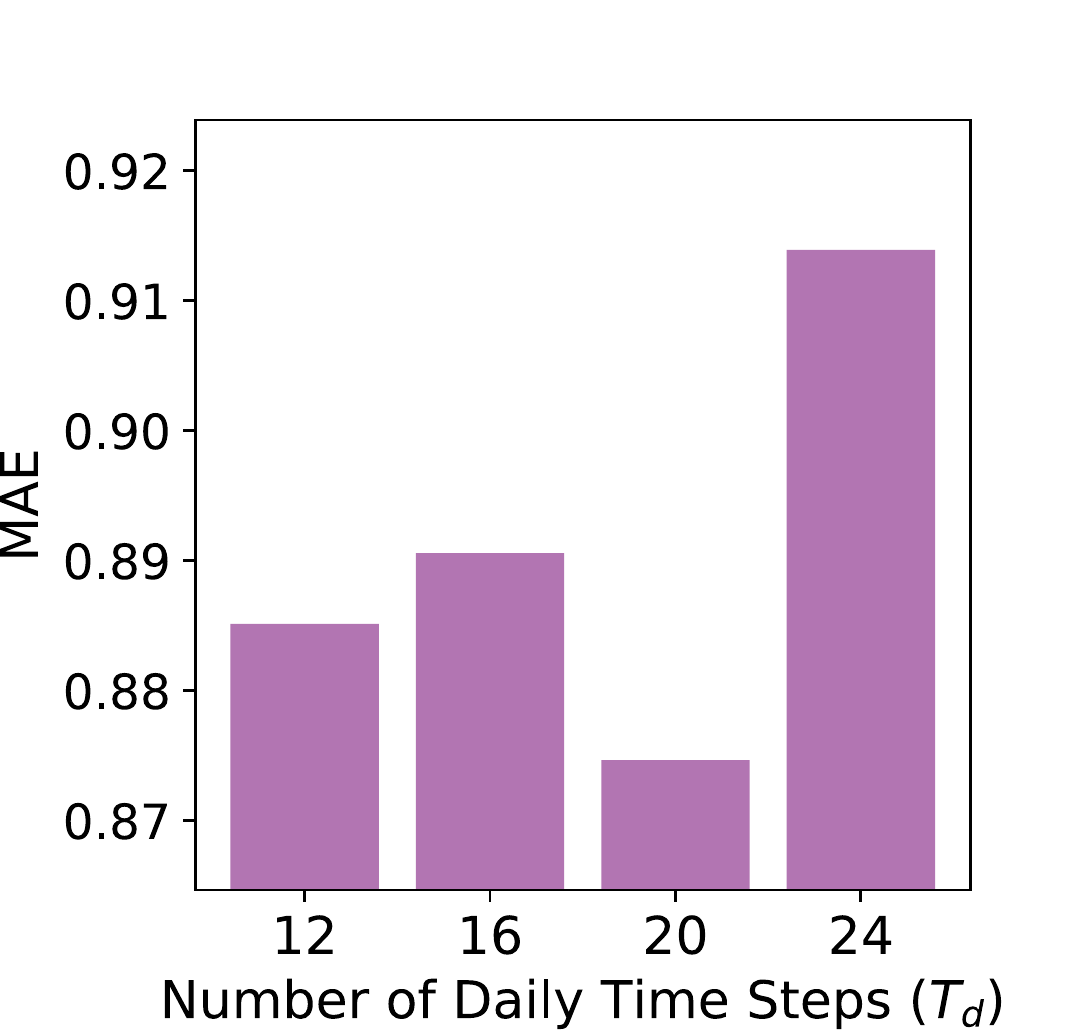
		\caption{C1: Theft}
		\label{Fig:param_TD1}
	\end{subfigure}
	\begin{subfigure}{0.245\columnwidth}
	    \def\svgwidth{\columnwidth}
	    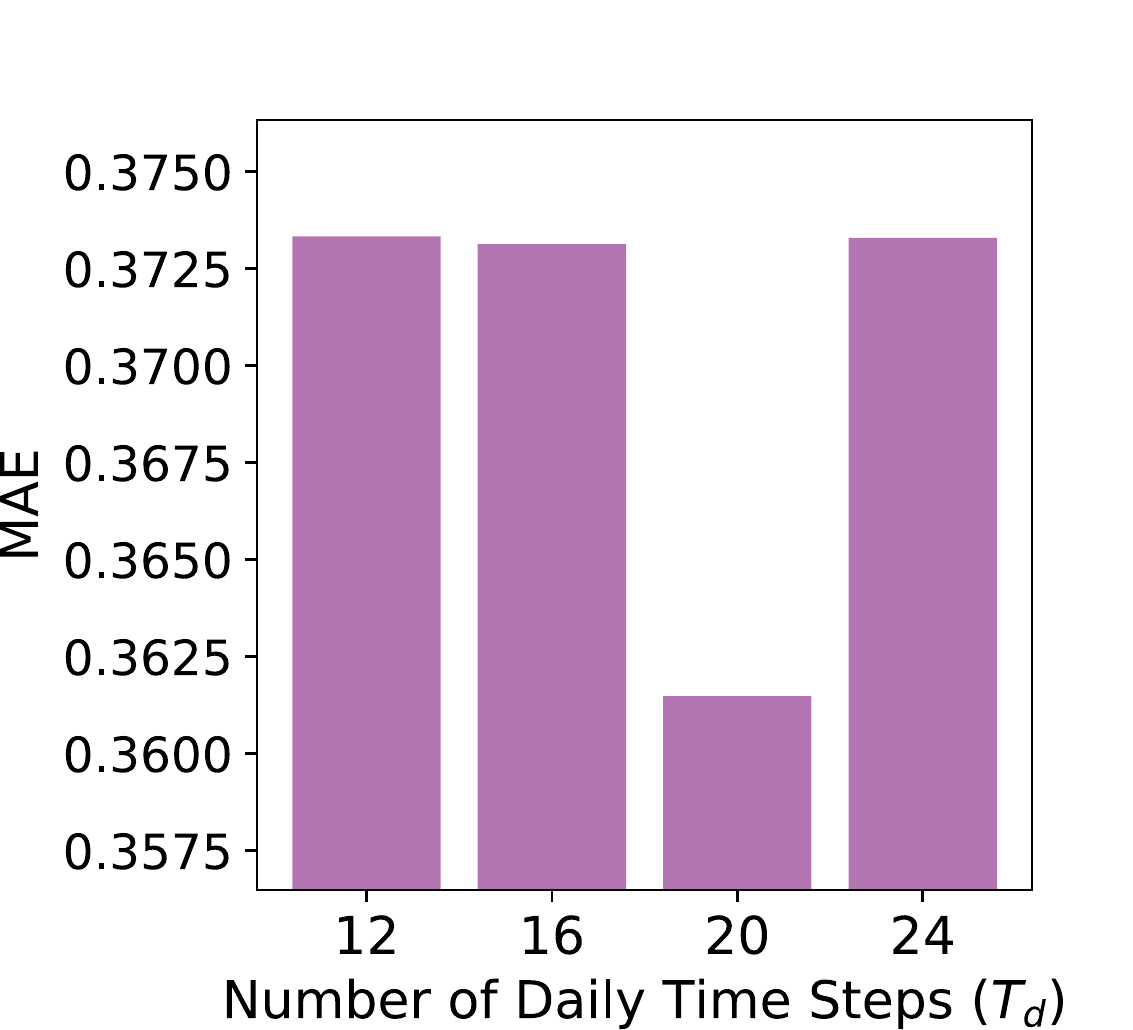
		\caption{C2: Criminal Damage}
		\label{Fig:param_TD2}
	\end{subfigure}
	\begin{subfigure}{0.245\columnwidth}
	    \def\svgwidth{\columnwidth}
	    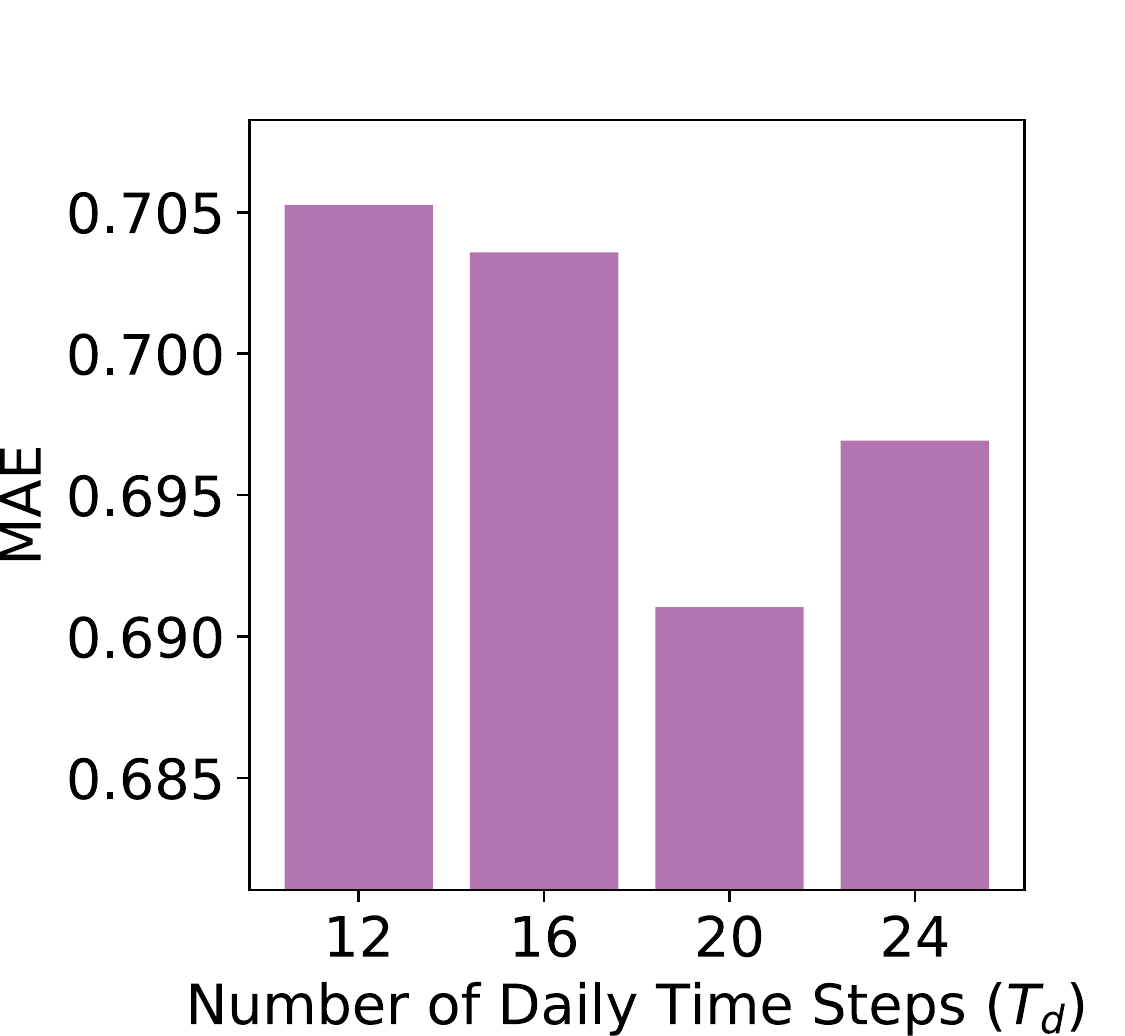
		\caption{C3: Battery}
		\label{Fig:param_TD3}
	\end{subfigure}
	\begin{subfigure}{0.245\columnwidth}
	    \def\svgwidth{\columnwidth}
	    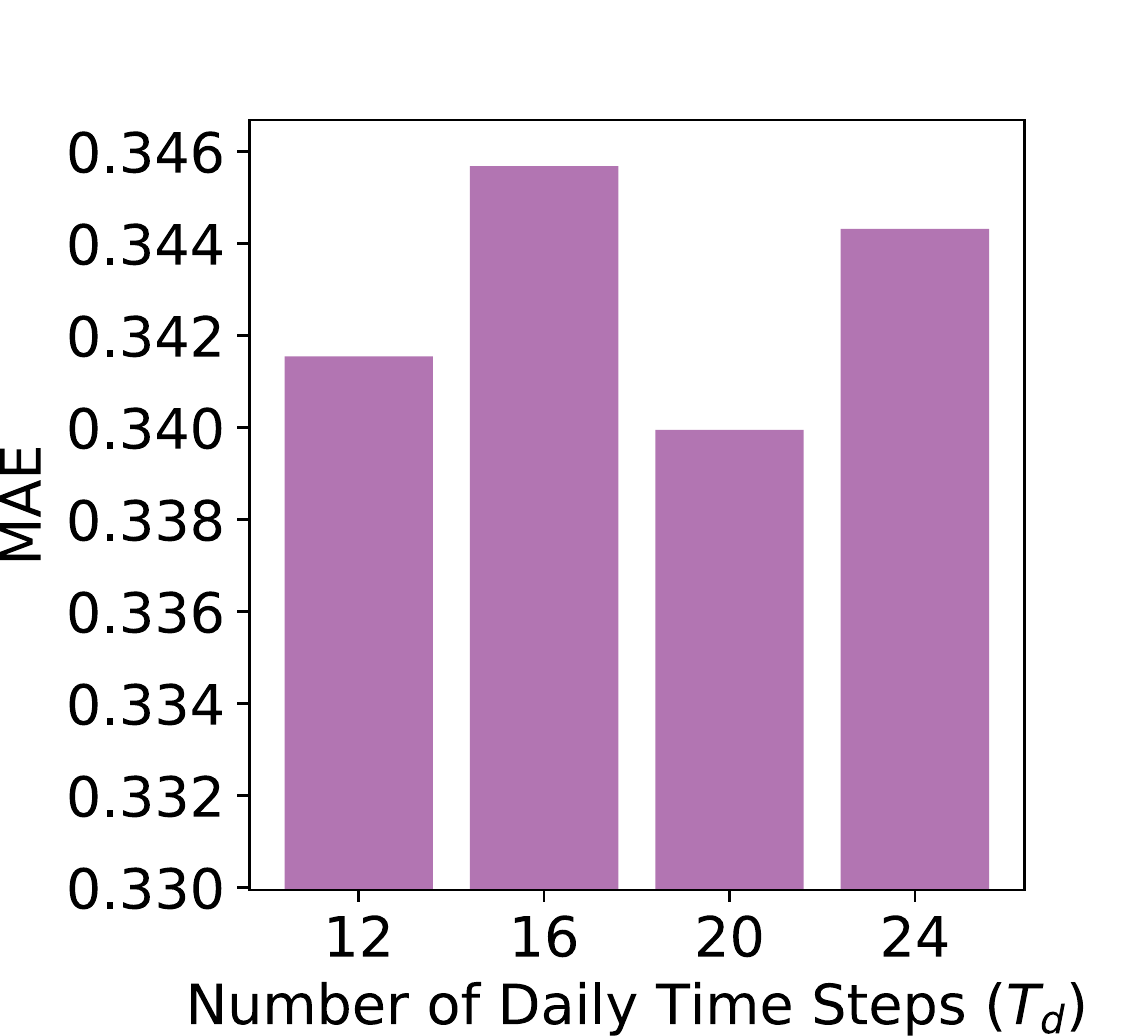
		\caption{C4: Narcotics}
		\label{Fig:param_TD7}
	\end{subfigure}
	\caption{Effect of the Number of Daily Time Steps, $T_d$}
	\label{Fig:param_TD}
	\vspace{-1em}
\end{figure}

\begin{figure}[htbp]
	\begin{subfigure}{0.245\columnwidth}
	    \def\svgwidth{\columnwidth}
	    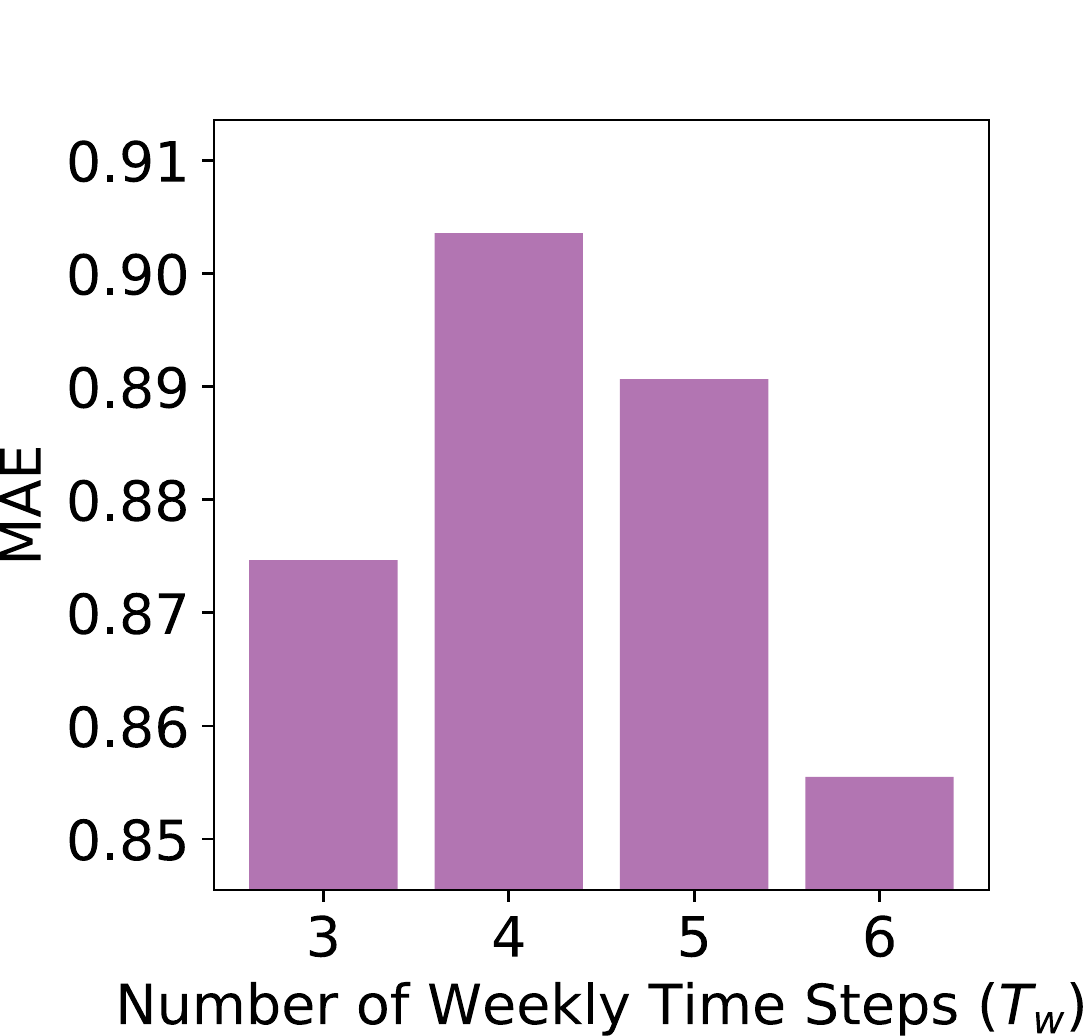
		\caption{C1: Theft}
		\label{Fig:param_TW1}
	\end{subfigure}
	\begin{subfigure}{0.245\columnwidth}
	    \def\svgwidth{\columnwidth}
	    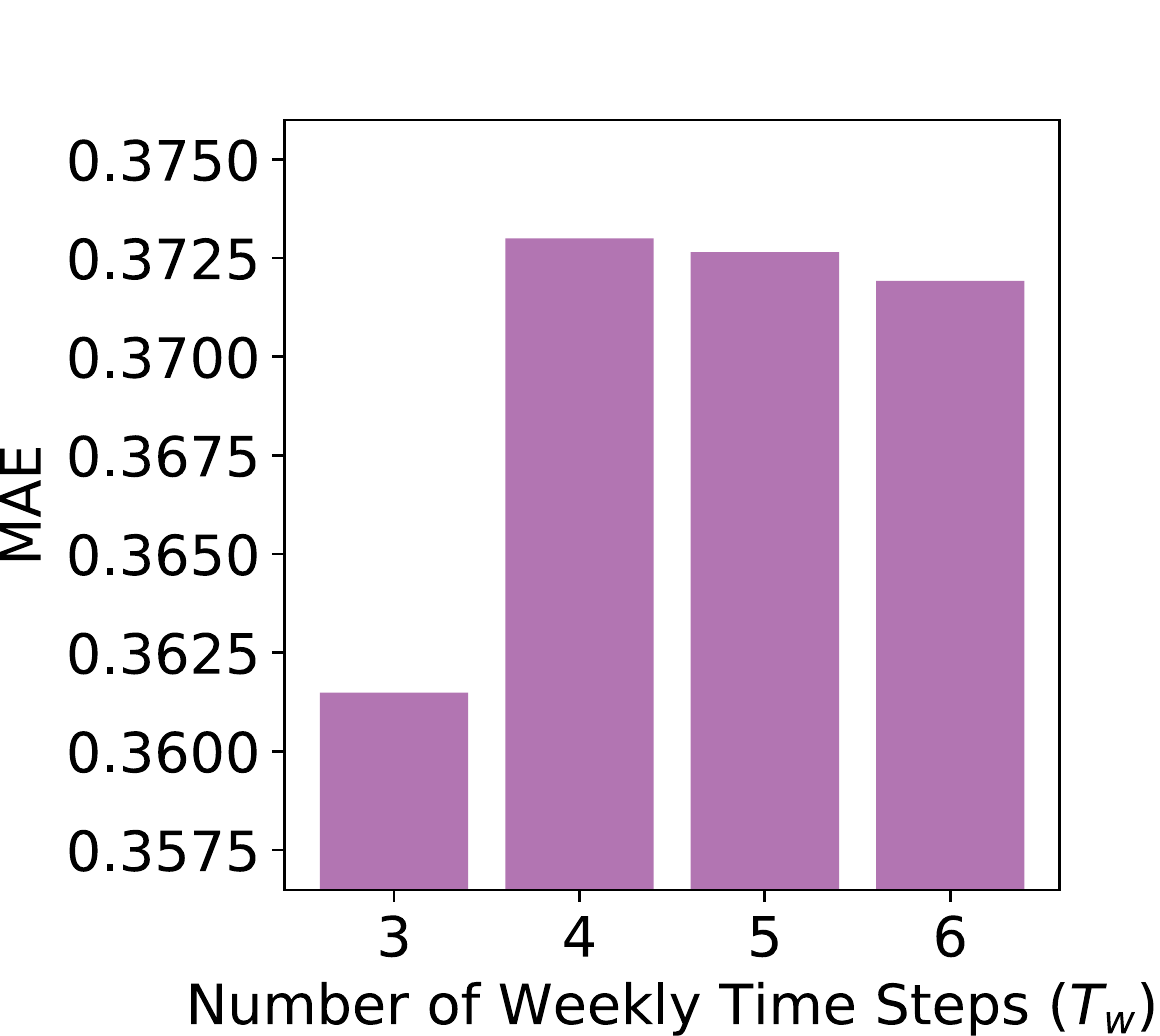
		\caption{C2: Criminal Damage}
		\label{Fig:param_TW2}
	\end{subfigure}
	\begin{subfigure}{0.245\columnwidth}
	    \def\svgwidth{\columnwidth}
	    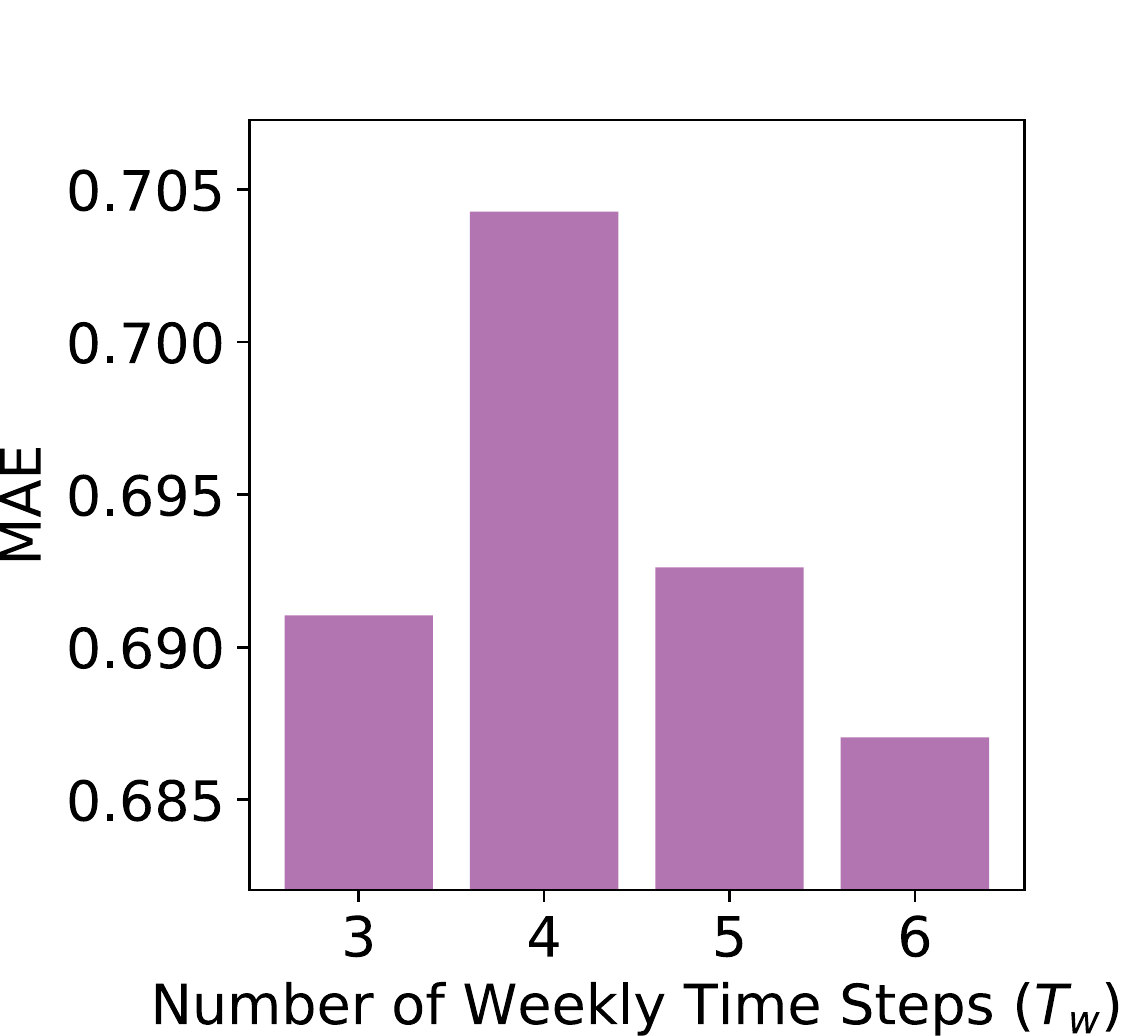
		\caption{C3: Battery}
		\label{Fig:param_TW3}
	\end{subfigure}
	\begin{subfigure}{0.245\columnwidth}
	    \def\svgwidth{\columnwidth}
	    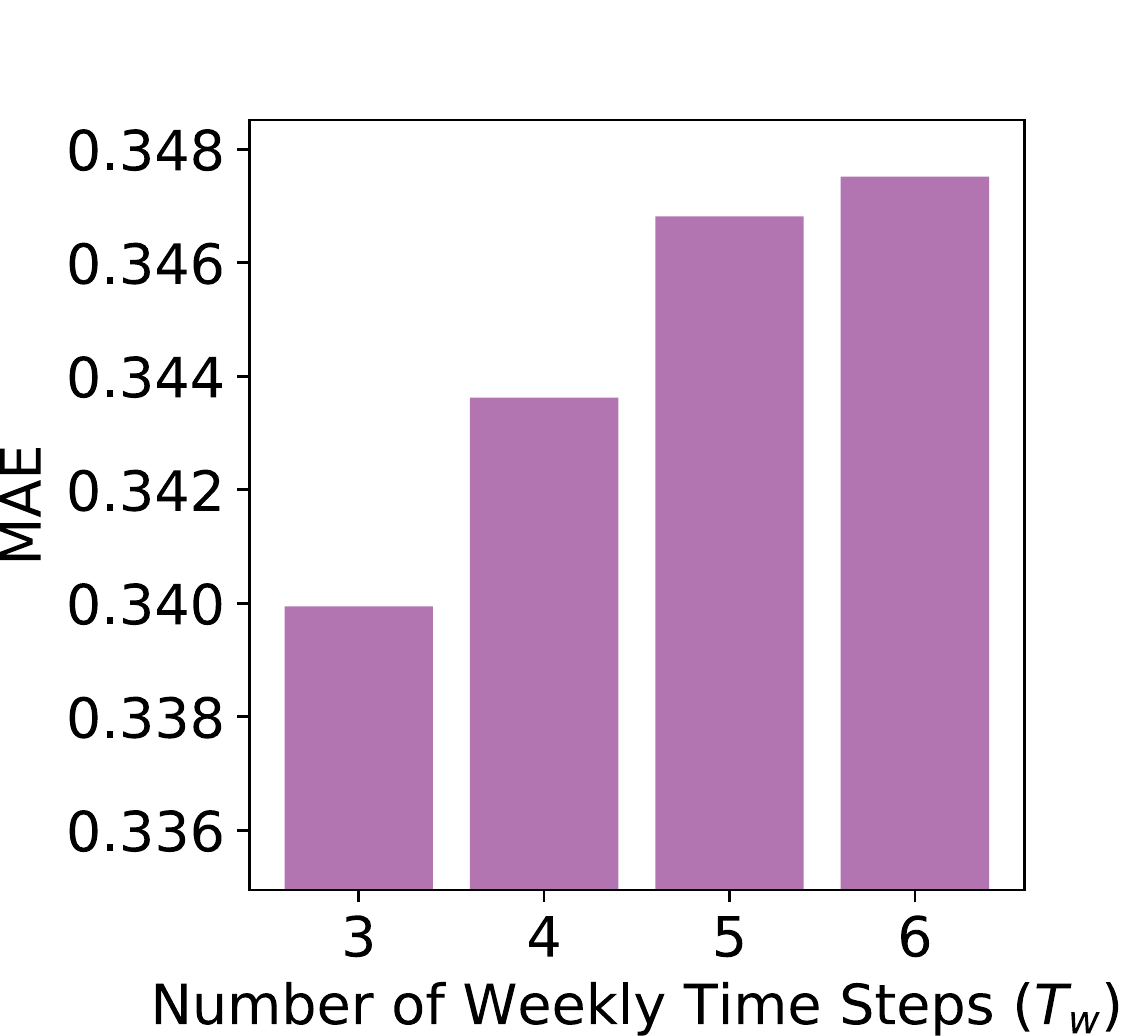
		\caption{C4: Narcotics}
		\label{Fig:param_TW4}
	\end{subfigure}
	\caption{Effect of the Number of Weekly Time Steps, $T_w$}
	\label{Fig:param_TW}
    \vspace{-1em}
\end{figure}

\begin{figure}[htbp]
	\begin{subfigure}{0.245\columnwidth}
	    \def\svgwidth{\columnwidth}
	    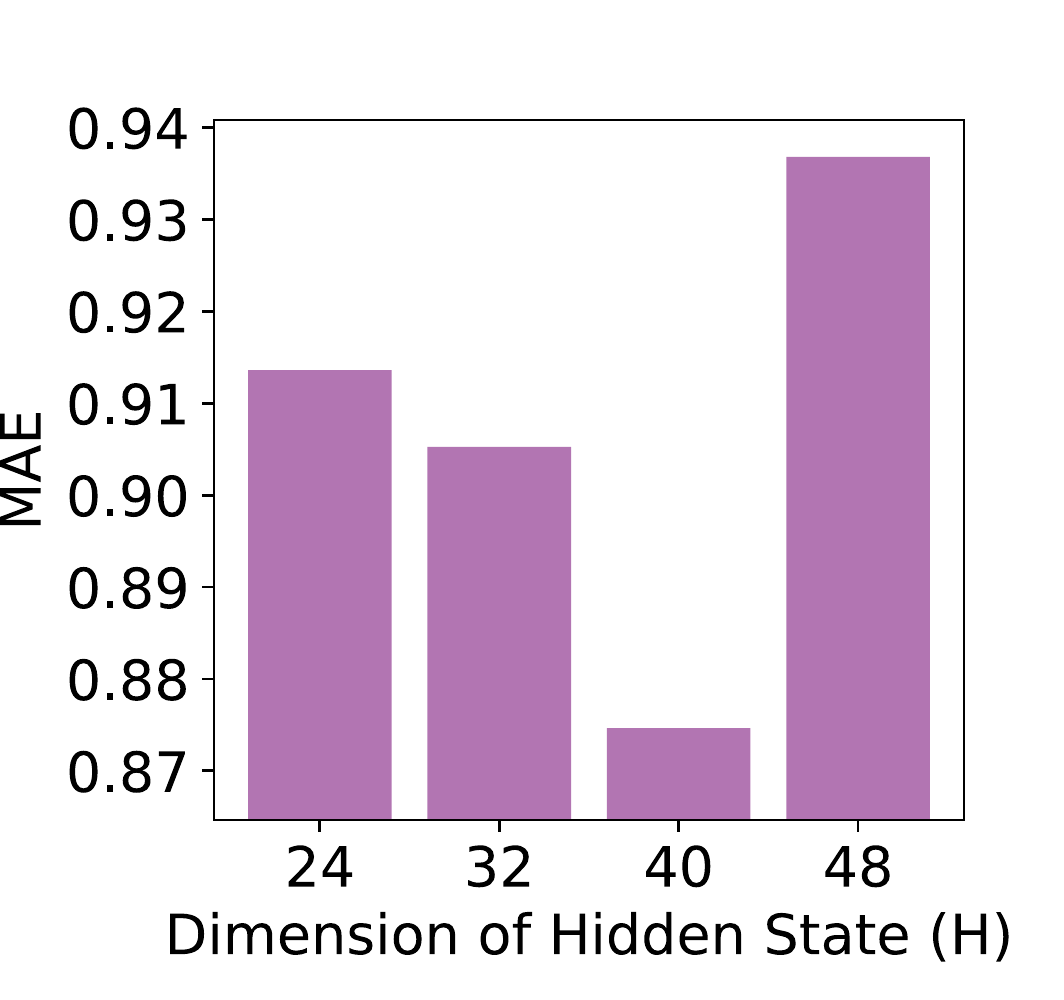
		\caption{C1: Theft}
		\label{Fig:param_H1}
	\end{subfigure}
	\begin{subfigure}{0.245\columnwidth}
	    \def\svgwidth{\columnwidth}
	    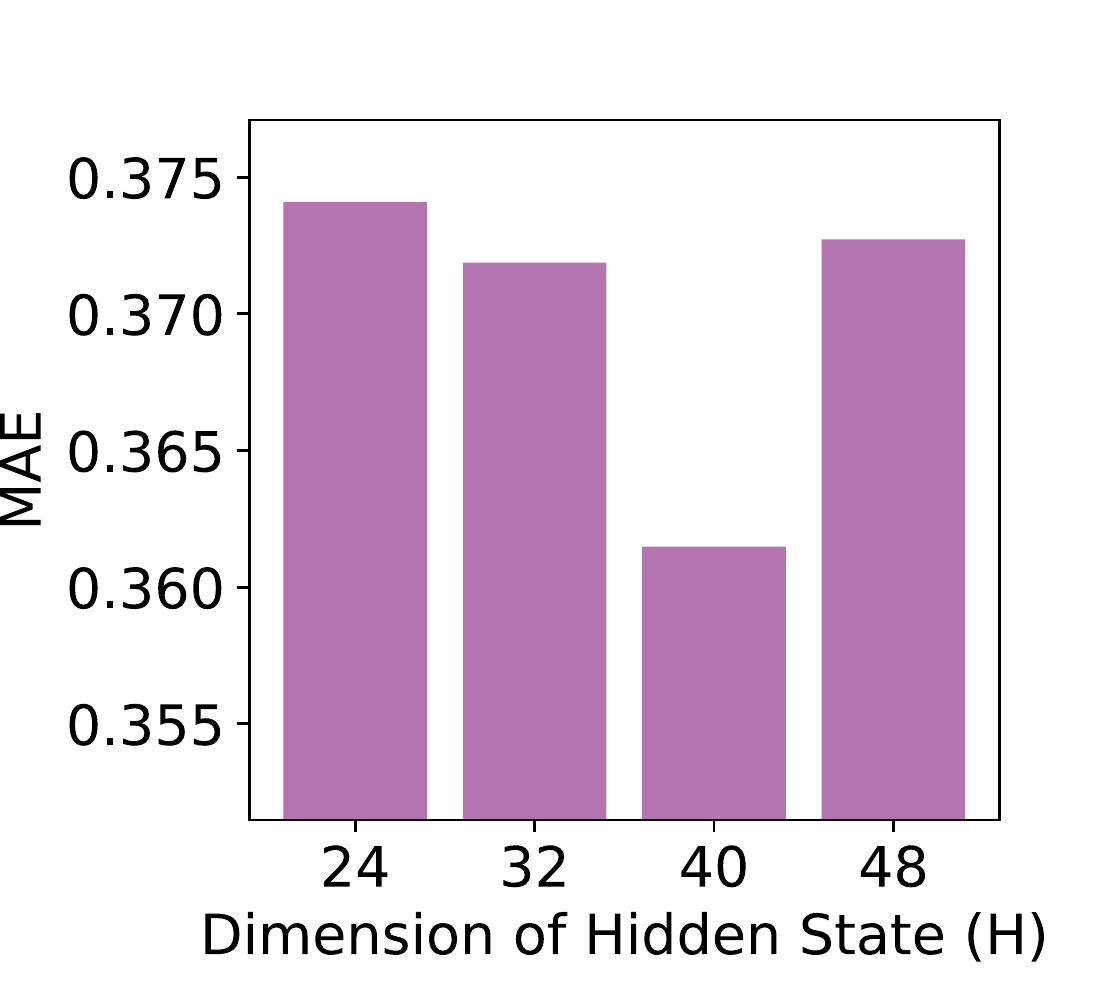
		\caption{C2: Criminal Damage}
		\label{Fig:param_H2}
	\end{subfigure}
	\begin{subfigure}{0.245\columnwidth}
	    \def\svgwidth{\columnwidth}
	    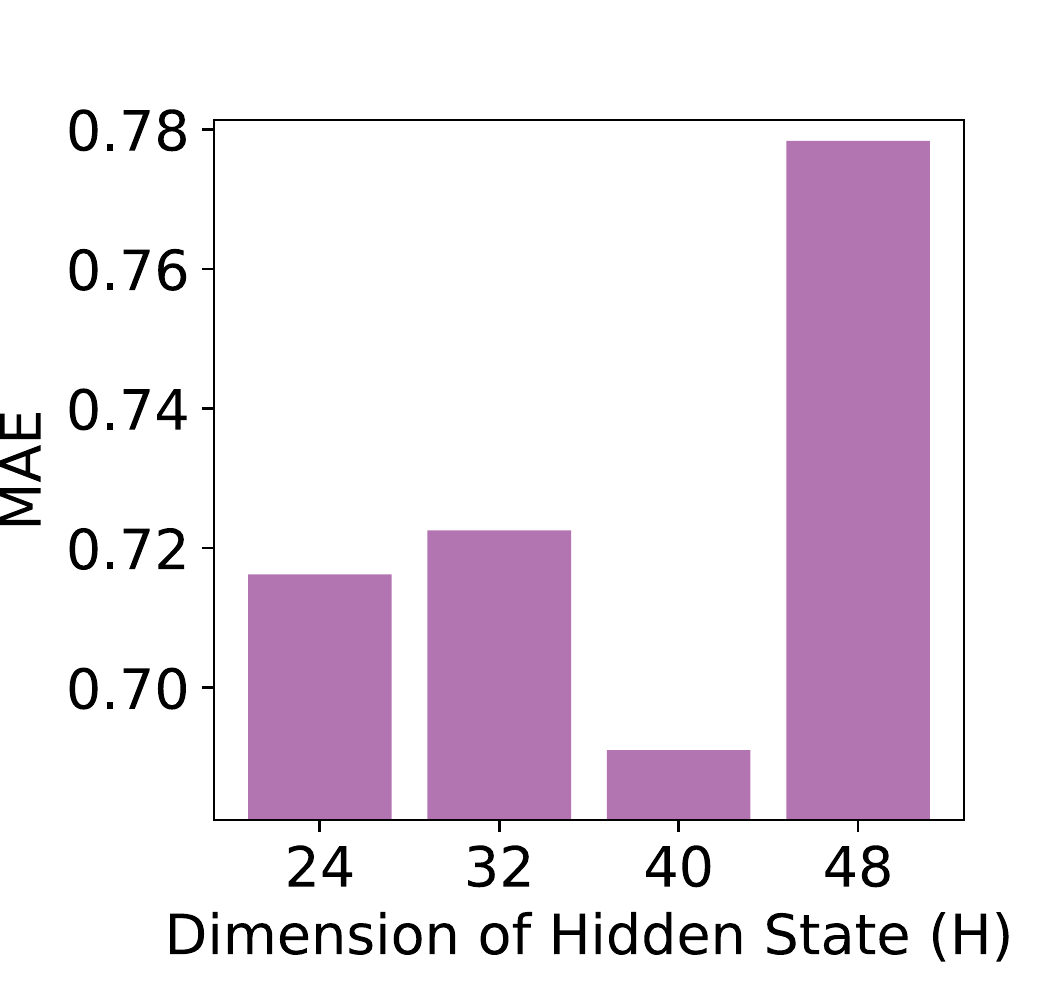
		\caption{C3: Battery}
		\label{Fig:param_H3}
	\end{subfigure}
	\begin{subfigure}{0.245\columnwidth}
	    \def\svgwidth{\columnwidth}
	    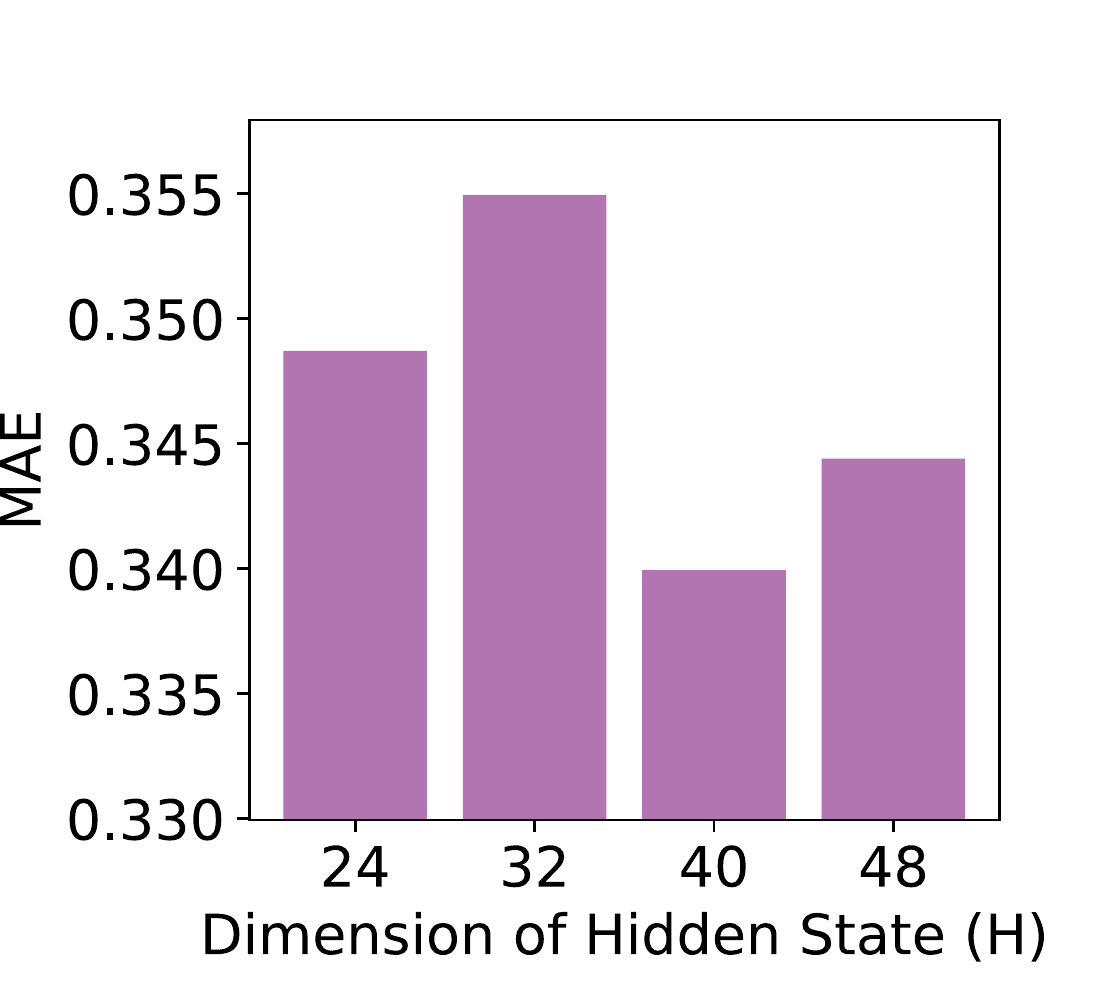
		\caption{C4: Narcotics}
		\label{Fig:param_H7}
	\end{subfigure}
	\caption{Effect of the Dimension of Hidden State, $H$}
	\label{Fig:param_H}
	\vspace{-1em}
\end{figure}

\begin{figure}[htbp]
	\begin{subfigure}{0.245\columnwidth}
	    \def\svgwidth{\columnwidth}
	    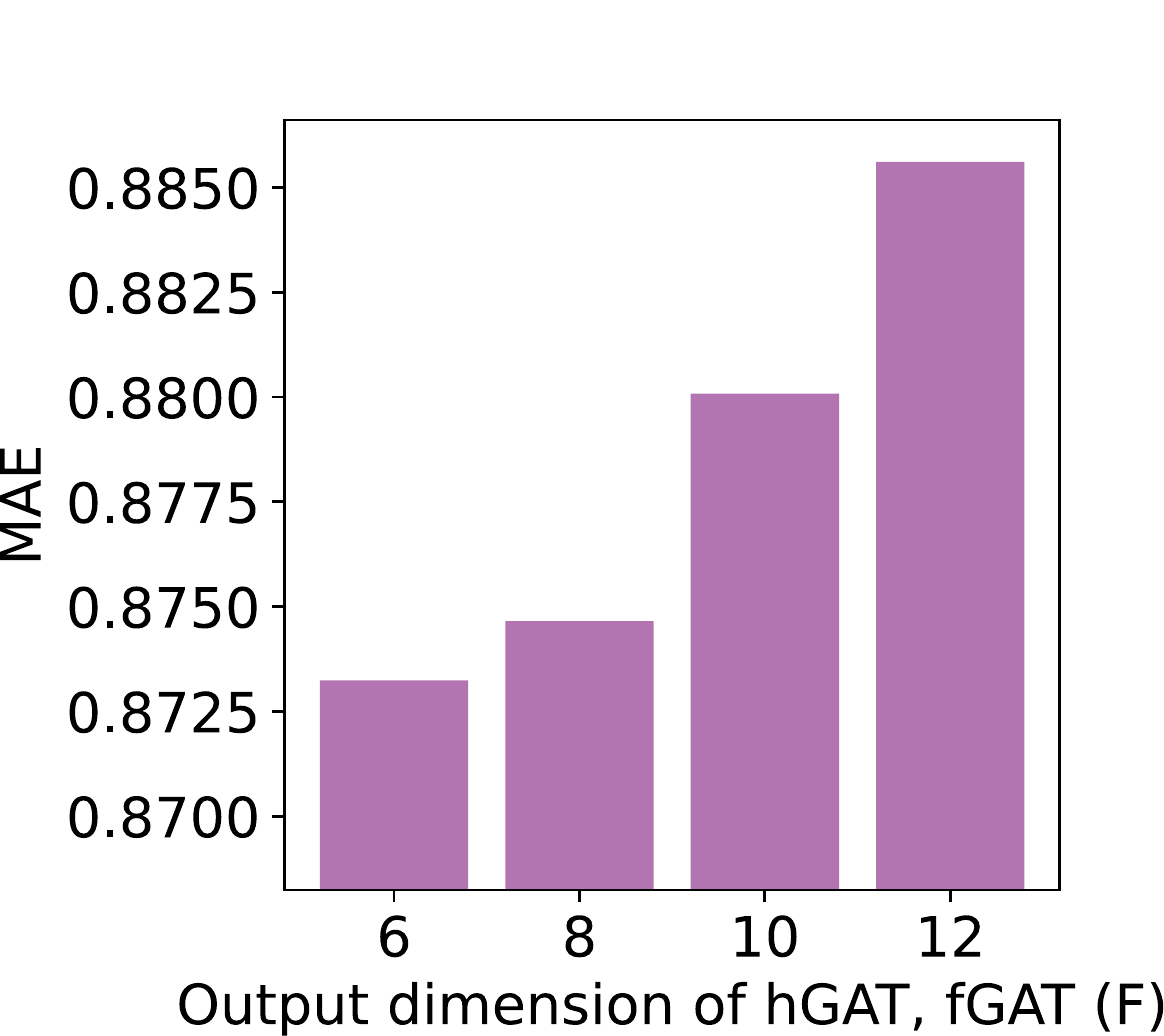
		\caption{C1: Theft}
		\label{Fig:param_F1}
	\end{subfigure}
	\begin{subfigure}{0.245\columnwidth}
	    \def\svgwidth{\columnwidth}
	    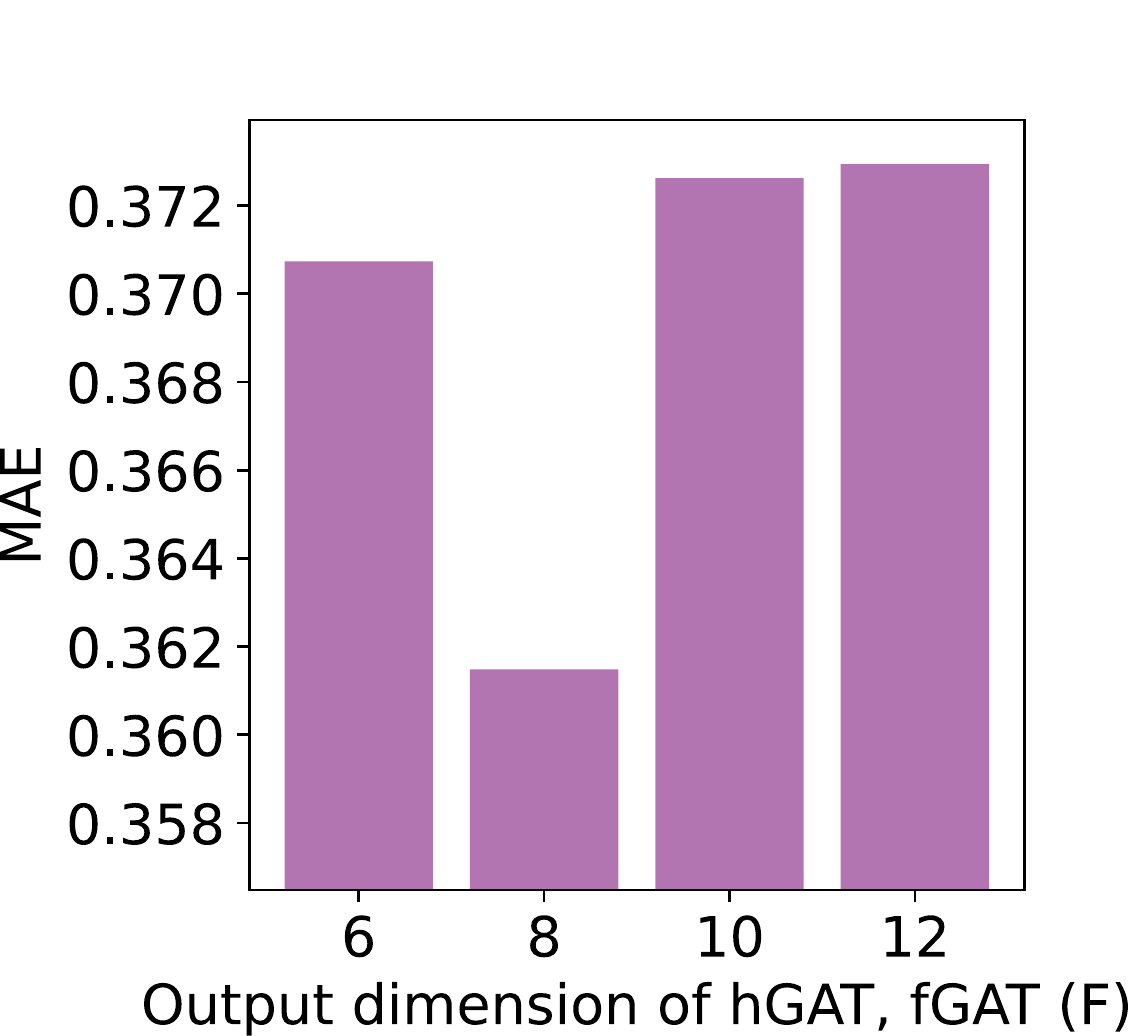
		\caption{C2: Criminal Damage}
		\label{Fig:param_F2}
	\end{subfigure}
	\begin{subfigure}{0.245\columnwidth}
	    \def\svgwidth{\columnwidth}
	    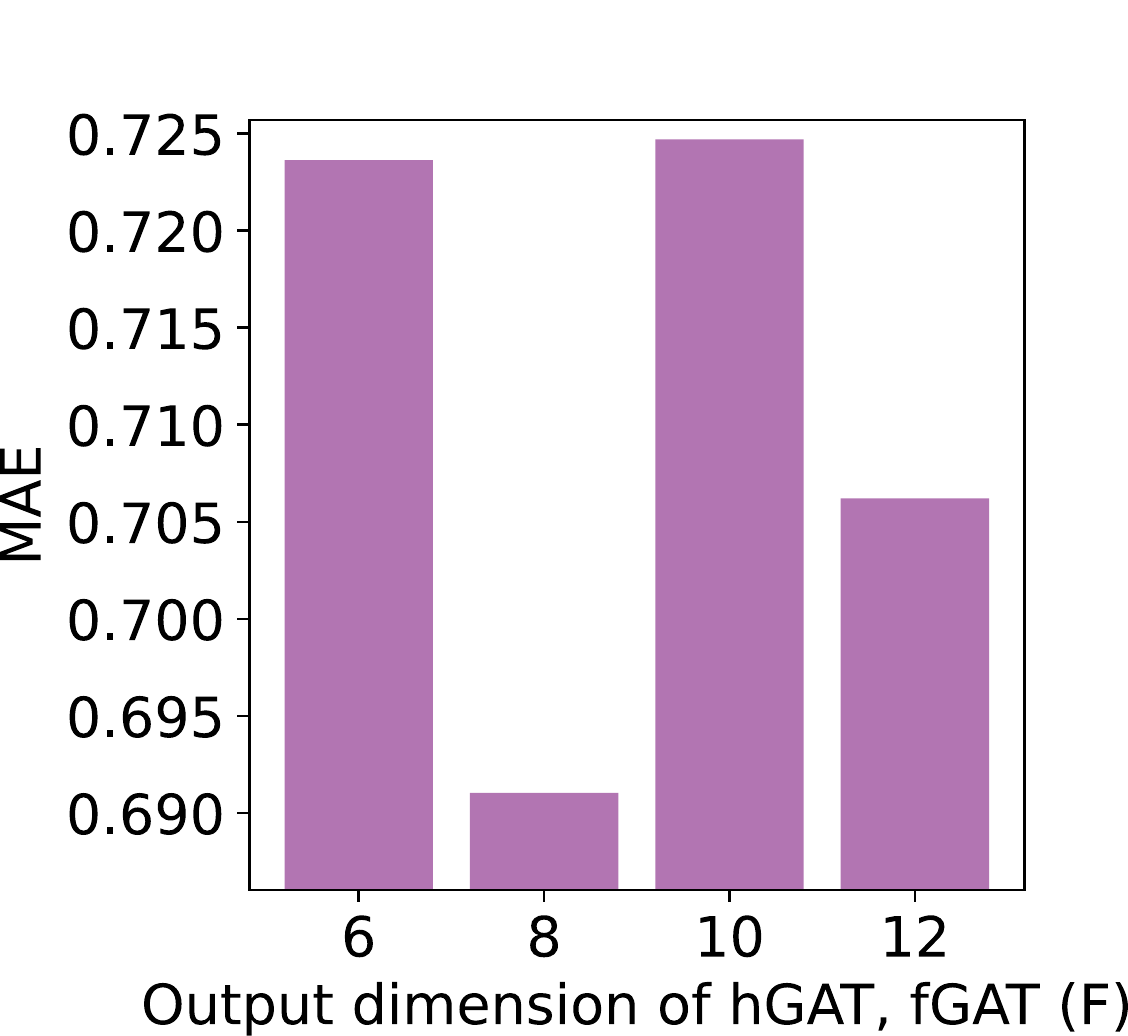
		\caption{C3: Battery}
		\label{Fig:param_F3}
	\end{subfigure}
	\begin{subfigure}{0.245\columnwidth}
	    \def\svgwidth{\columnwidth}
	    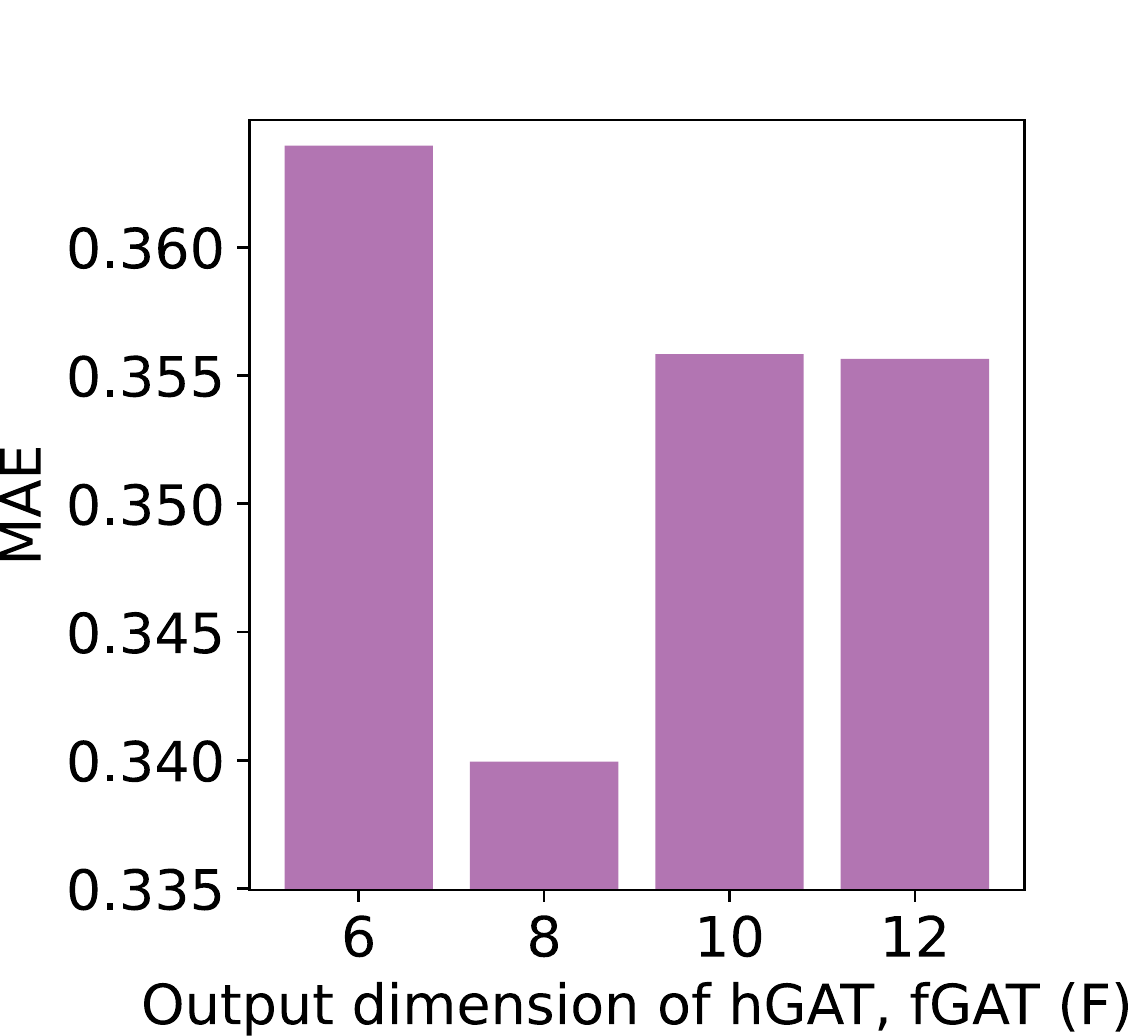
		\caption{C4: Narcotics}
		\label{Fig:param_F7}
	\end{subfigure}
	\caption{Effect of the Output Size of hGAT and fGAT, $F$}
	\label{Fig:param_F}
\end{figure}

\begin{figure}[htbp]
\begin{subfigure}{0.245\columnwidth}
	\def\svgwidth{\columnwidth}
	    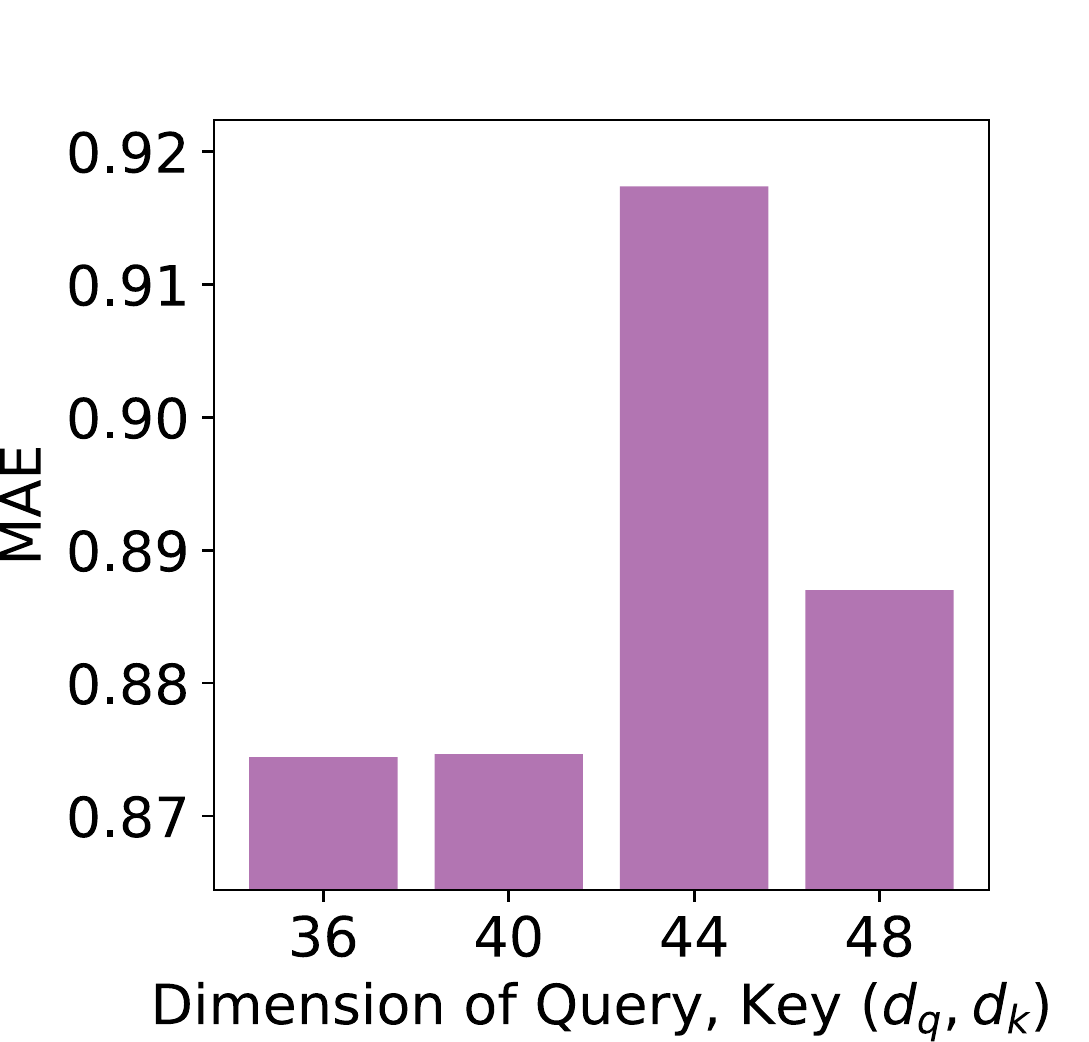
		\caption{C1: Theft}
		\label{Fig:param_Q1}
    \end{subfigure}
	\begin{subfigure}{0.245\columnwidth}
	\def\svgwidth{\columnwidth}
	    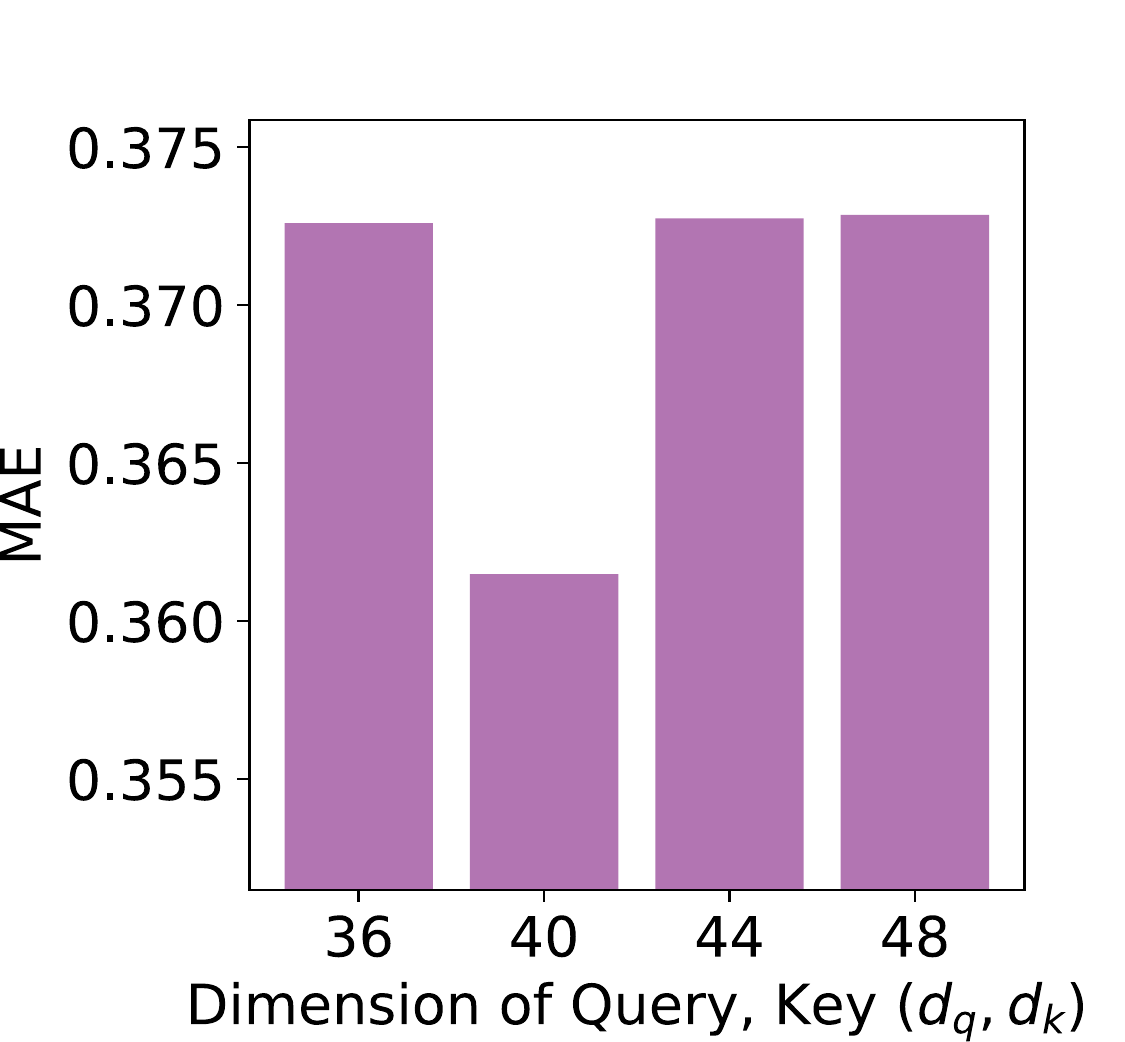
		\caption{C2: Criminal Damage}
		\label{Fig:param_Q2}
  \end{subfigure}
  \begin{subfigure}{0.245\columnwidth}
	\def\svgwidth{\columnwidth}
	    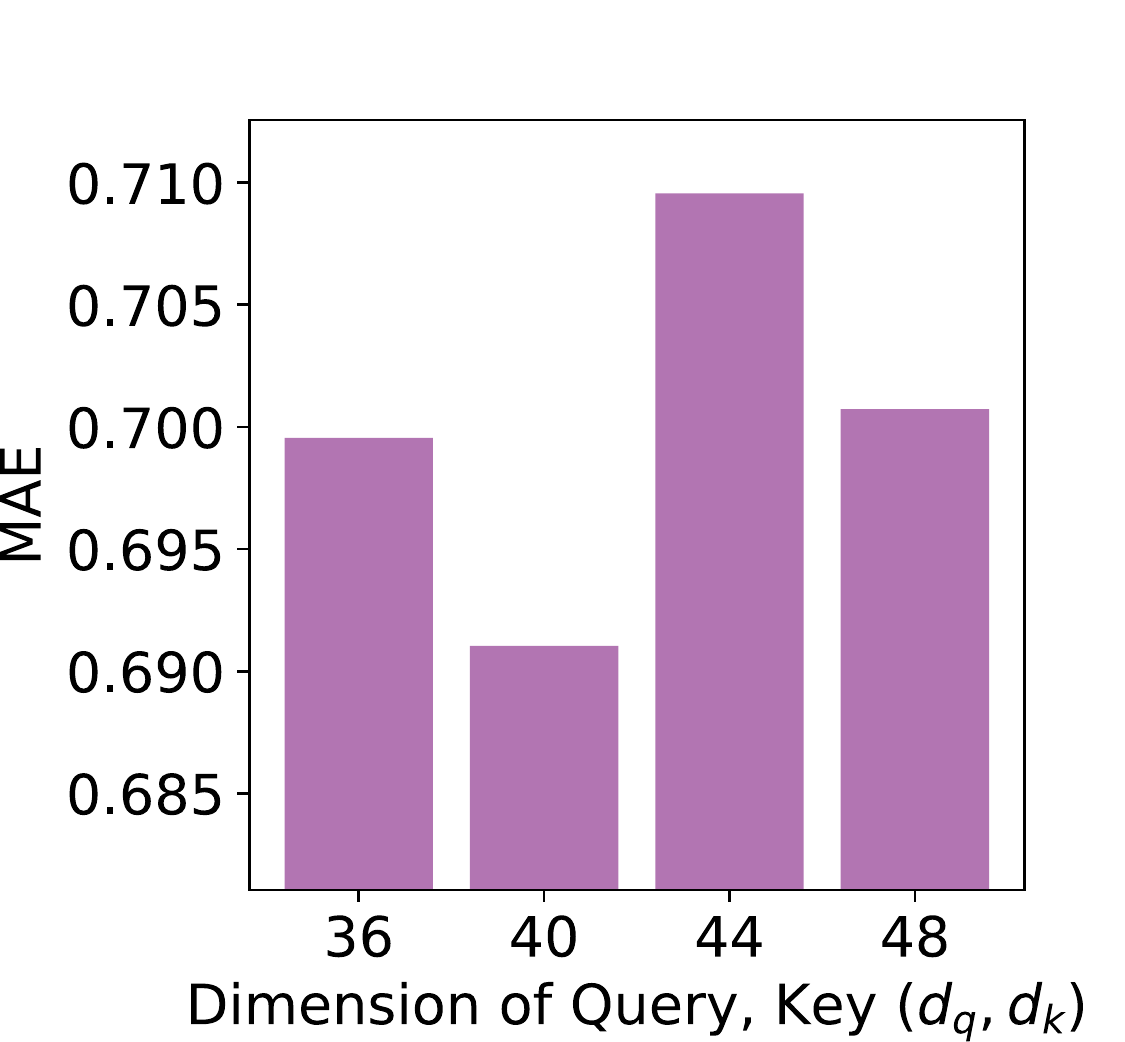
		\caption{C3: Battery}
		\label{Fig:param_Q3}
  \end{subfigure}
  \begin{subfigure}{0.245\columnwidth}
	\def\svgwidth{\columnwidth}
	    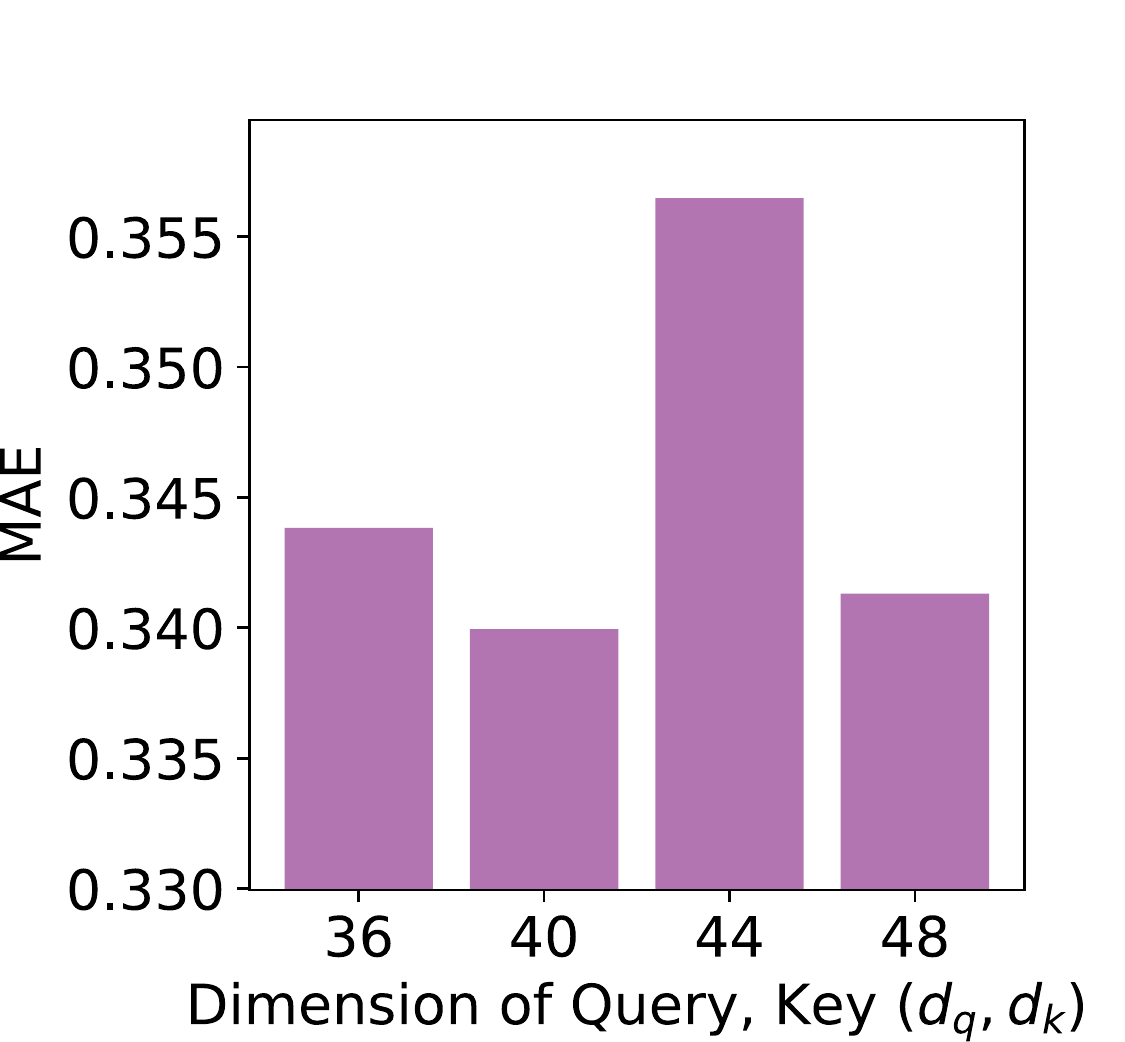
		\caption{C4: Narcotics}
		\label{Fig:param_Q7}
  \end{subfigure}
\caption{Effect of the Dimension of Query and Key, $d_q, d_k$}
\label{Fig:param_Q}
\end{figure}

\begin{figure}[htbp]
	\begin{subfigure}{0.245\columnwidth}
	    \def\svgwidth{\columnwidth}
	    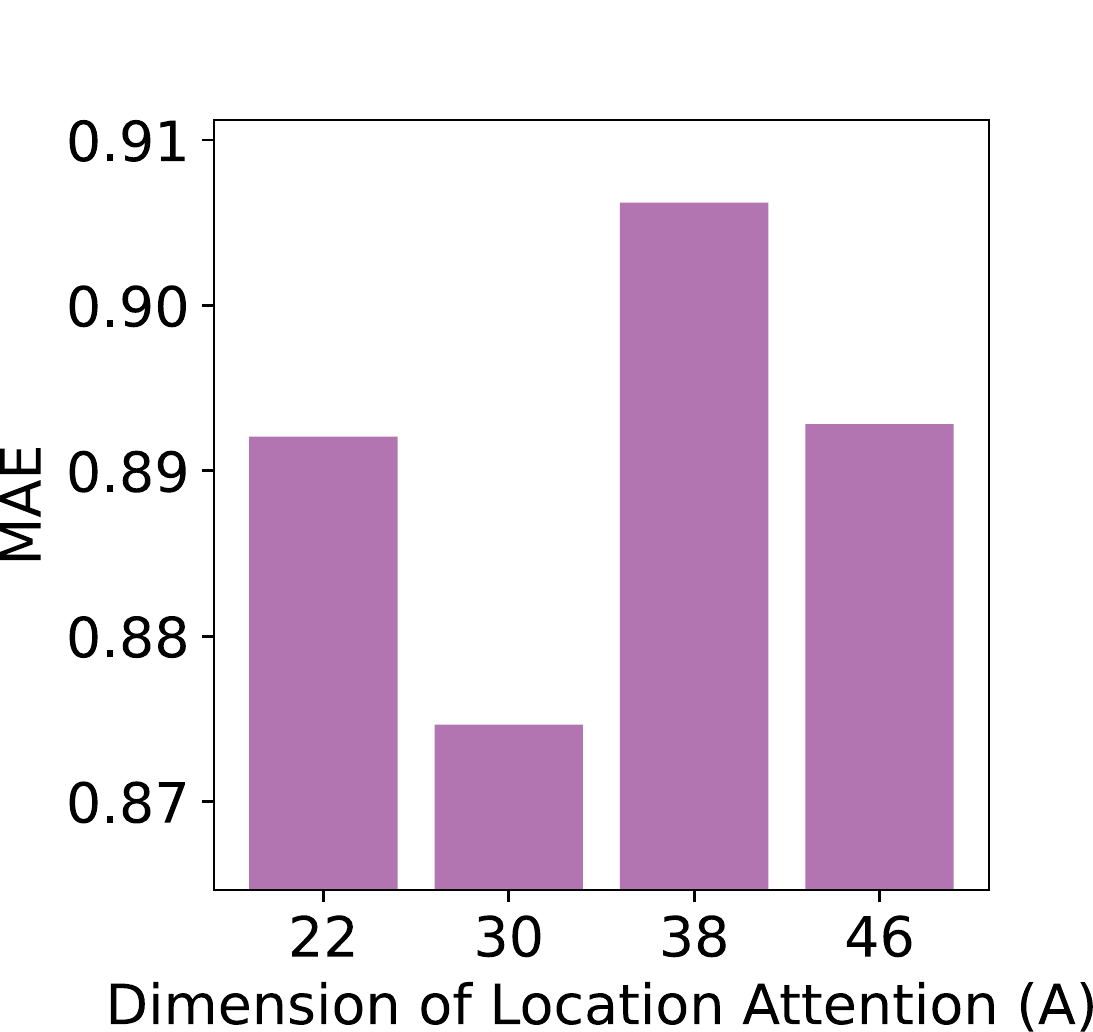
		\caption{C1: Theft}
		\label{Fig:param_A1}
	\end{subfigure}
	\begin{subfigure}{0.245\columnwidth}
	    \def\svgwidth{\columnwidth}
	    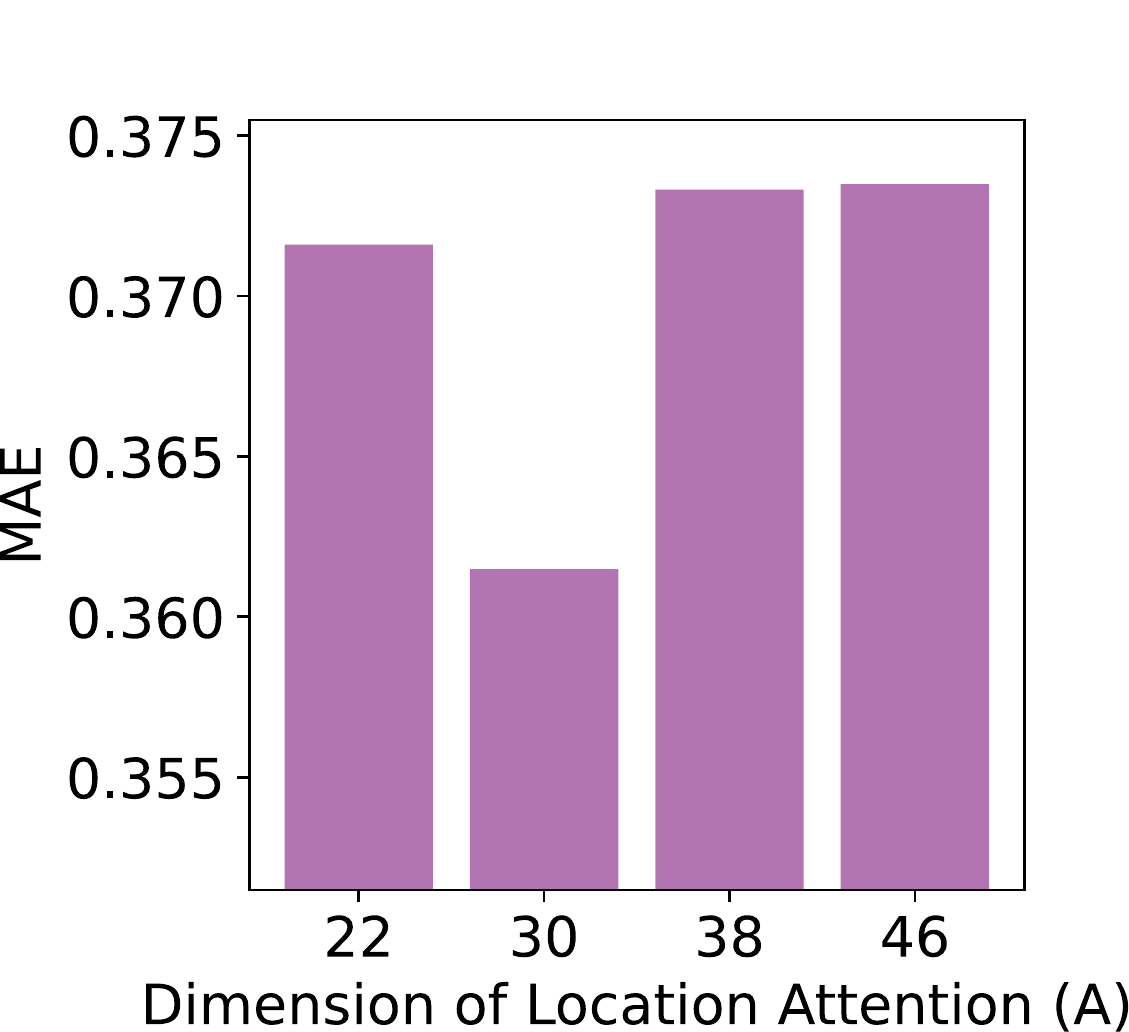
		\caption{C2: Criminal Damage}
		\label{Fig:param_A2}
	\end{subfigure}
	\begin{subfigure}{0.245\columnwidth}
	    \def\svgwidth{\columnwidth}
	    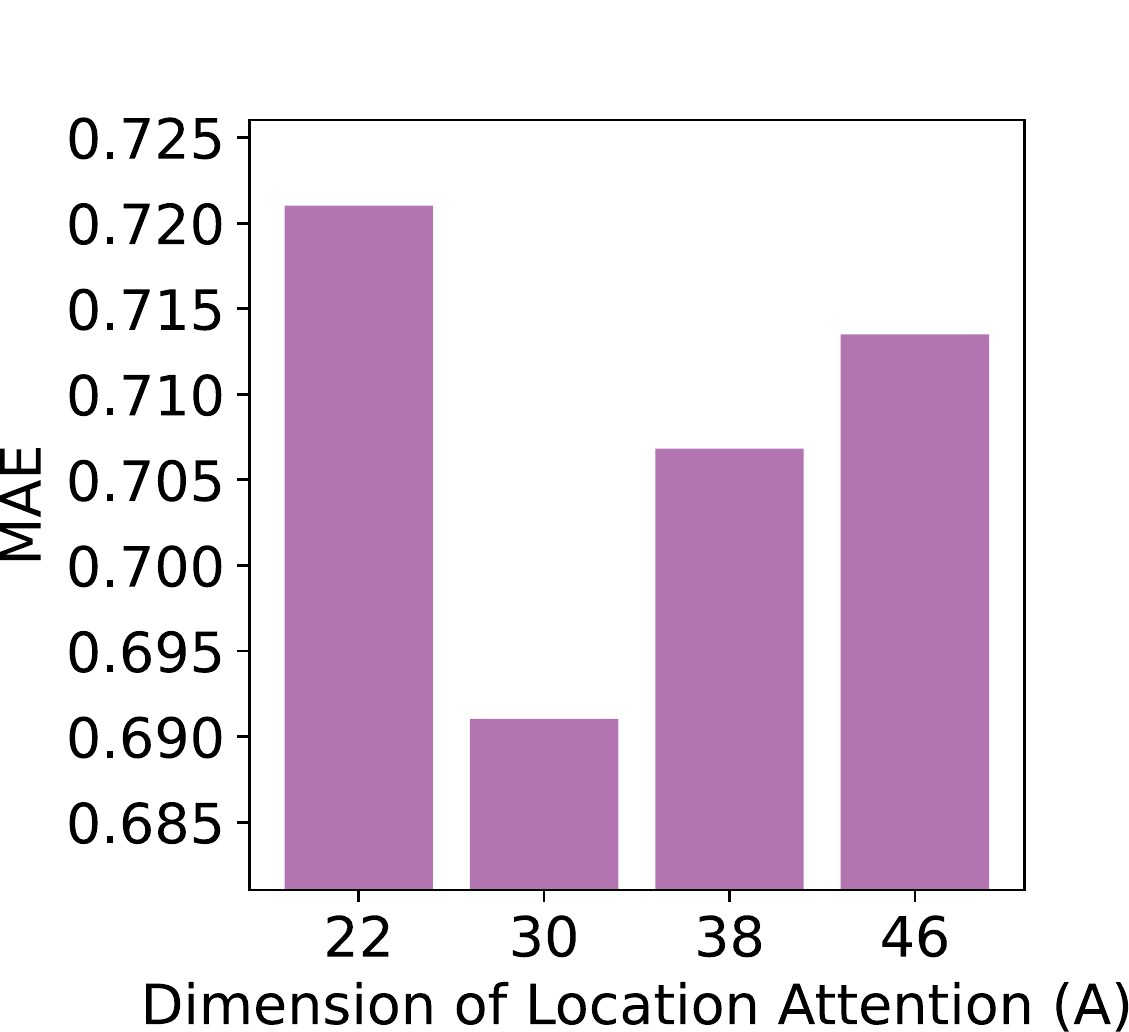
		\caption{C3: Battery}
		\label{Fig:param_A3}
	\end{subfigure}
	\begin{subfigure}{0.245\columnwidth}
	    \def\svgwidth{\columnwidth}
	    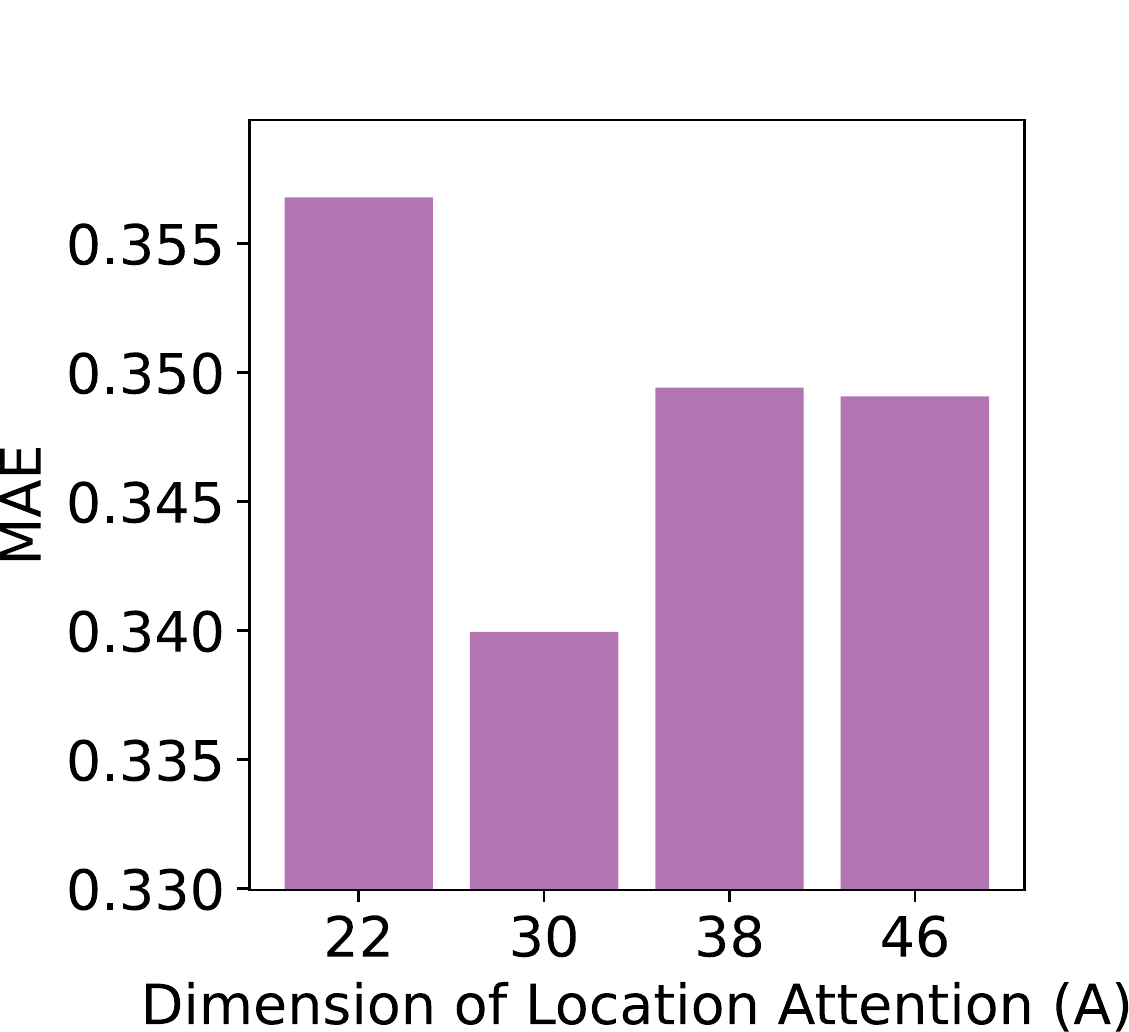
		\caption{C4: Narcotics}
		\label{Fig:param_A7}
	\end{subfigure}
	\caption{Effect of the Dimension of Location Attention, $A$}
	\label{Fig:param_A}
\end{figure}

Different parameters like number of recent ($T$), daily ($T_d$) and weekly ($T_w$) time steps, dimension of SAB-LSTM hidden states ($H$), output size of hGAT and fGAT ($F$), dimension of query ($d_q$), key ($d_k$) vectors and location attention ($A$) have impact on the performance of AIST. To better understand the crucial parameters of AIST and their effect on its prediction performance, we run several experiments and present the results in ~\Cref{Fig:param_T,Fig:param_TD,Fig:param_TW,Fig:param_H,Fig:param_F,Fig:param_Q,Fig:param_A}. To observe the effect of a parameter, the value of the parameter is varied within its range, and other parameters are set to their default values as presented in Table~\ref{table:exp_hyperparameter}.

Figures~\ref{Fig:param_T},~\ref{Fig:param_TD}, and~\ref{Fig:param_TW} suggest that AIST is sensitive to the number of recent ($T$), daily ($T_d$) and weekly ($T_w$) time steps, which are being fed to the three SAB-LSTMs as input. AIST shows poor performance for both smaller and larger $T, T_d$ due to the lack of data for learning temporal dependencies and the absence of long temporal correlation, respectively. Somewhere in between, when $T, T_d = 20$, AIST in general performs best by capturing the recent and periodic properties of the crime. On the contrary, AIST performs well when the number of weekly time steps $(T_w)$ is relatively small (Figure~\ref{Fig:param_TW}). However, unlike other categories, AIST performs best for category Theft (C1), when the number of recent and weekly time steps are comparatively larger (Figure~\ref{Fig:param_T1},~\ref{Fig:param_TW1}). This signifies the existence of long term temporal dependencies for category Theft.

AIST is also sensitive to the dimension $(H)$ of the hidden states of SAB-LSTMs (Figure~\ref{Fig:param_H}) and output size $(F)$ of hGAT and fGAT (Figure~\ref{Fig:param_F}). Limited spatial and temporal information make the training hard for AIST. As a result, the performance of AIST deteriorates. On the other hand, a larger output size and dimension of the hidden state make it easier for AIST to overfit the data. Hence, we set $F=8$ and $H=40$ so that AIST can generalize well by learning sufficient spatial and temporal information. 

It is evident from Figures~\ref{Fig:param_Q} and~\ref{Fig:param_A} that the query and key dimensions $(d_q, d_k)$ and location attention dimension $(A)$ follow the same trend as the other hyperparameters discussed above. Based on the performance of AIST across different crime categories, we set $d_q, d_k = 40$ and $A = 30$.


\subsubsection{Effect of Train/Test Ratio}
We run several experiments to learn the effect of the train-test ratio on the prediction performance of AIST. We consider the first $n \in \{6, 7, 8, 9, 10\}$ months of Chicago crime data (2019) as training set. We take 10\% of the remaining data of last $(12-n)\in \{6, 5, 4, 3, 12\}$ months for the validation set and use the rest as the test set. Table~\ref{table:tt_ratio} shows the prediction performance of AIST across all crime categories against different train-test ratio.

Similar to the most deep learning models, fewer training samples cause AIST to overfit the data and as a result AIST shows poor prediction performance. This is evident from the reported MAE and MSE scores of AIST when only 6 months of data is used for training.  Once we gradually increase the size of the training data, the prediction performance of AIST improves significantly and reaches its peak when the size of the training data is 8 months. Adding additional training data beyond 8 months again deteriorates the prediction performance of AIST. 


\begin{small}
\begin{table*}[htbp]
	\caption{Effect of Train-Test Ratio on Crime Prediction Performance of AIST}
	\centering
		\begin{tabular}[t]{|E|D|D|D|D|D|D|D|D|}
		    \hline
            Train / Test Ratio 
            & \multicolumn{2}{c|}{\centering Theft (C1)} 
            & \multicolumn{2}{c|}{\begin{tabular}{@{}c@{}} Criminal \\ Damage (C2)\end{tabular}}
            & \multicolumn{2}{c|}{Battery (C3)}
            & \multicolumn{2}{c|}{Narcotics (C4)}
            \\
			\hline
			In Percent  & MAE & MSE & MAE & MSE & MAE & MSE & MAE & MSE\\
			\hline
			
			0.50 / 0.45 & 1.2576 & 4.4737 & 0.3869 & 0.5121 & 0.7745 & 1.2920 & 0.4179 & 0.7457\\
			\hline
			
			0.58 / 0.38 & 0.9445 & 2.0771 & 0.3765 & 0.4972 & 0.7229 & 1.0653 & 0.3518 & 0.5619\\
			\hline
			
    		0.67 / 0.30 & \textbf{0.8747} & \textbf{1.6986} & \textbf{0.3615} & 0.4837 & \textbf{0.6910} & \textbf{0.9568} & \textbf{0.3399} & \textbf{0.5609}\\
			\hline
			
			0.75 / 0.23 & 1.3403 & 4.1698 & 0.3709 & 0.4748 & 0.7915 & 1.2591 & 0.4208 & 0.7391\\
			\hline
		    
		    0.83 / 0.15 & 1.2807 & 3.8667 & 0.3646 & 0.\textbf{4650} & 0.7988 & 1.1857 & 0.4240 & 0.7518\\
			\hline
	\end{tabular}

	\label{table:tt_ratio}
\end{table*}
\end{small}

\subsection{Evaluation of Interpretability}
The notion of interpretability mainly comes down to two points: i) plausibility: how understandable it is to humans, and ii) faithfulness: how accurately it refers to the true reasoning process of a model. 

Besides human evaluations~\cite{DBLP:conf/chi/KaurNJCWV20, 
DBLP:conf/iui/EhsanTCHR19, DBLP:conf/naacl/MullenbachWDSE18}, explanations that align directly with the input have been considered as plausible explanations~\cite{DBLP:conf/emnlp/LeiBJ16}. Attentions are plausible explanations because they assign importance weights to the inputs while making a prediction ~\cite{DBLP:conf/emnlp/WiegreffeP19}. Since AIST interprets the importance of different regions, features, time steps and trends on the crime prediction based on four attention modules, the interpretation of AIST is \textit{plausible}.

A recent study~\cite{DBLP:conf/emnlp/WiegreffeP19} shows the conditions under which attentions can be regarded as faithful explanations, and nullifies the claim~\cite{DBLP:conf/acl/SerranoS19,DBLP:conf/naacl/JainW19} that criticizes attention as a form of faithful explanation due to its weak correlation with other feature importance metrics and the existence of alternate adversarial attention weights. Specifically,~\cite{DBLP:conf/emnlp/WiegreffeP19} proposes a series of extensive experiments based on dataset and model properties: i) train on uniform attention weights: the attention distribution is frozen to uniform weights to validate whether the attention is actually necessary for a better performance, ii) calibration of variance: the model is trained with different initializing seeds to generate base variance for attention distributions, iii) train an MLP (multilayer perceptron): the LSTM cells are replaced by MLP and are trained separately and iv) train an adversary: the model is trained to provide similar predictions as the base model while keeping the attention distributions distant from the actual ones for ascertaining exclusivity. We evaluate the attention weights generated by AIST by performing these experiments (except iii since the attention modules used in AIST are either feed-forward neural networks or sparse which do not comply with the experimental settings) to validate their \textit{faithfulness}. 

Since the faithfulness varies across model, tasks and input space, both~\cite{DBLP:conf/emnlp/WiegreffeP19, DBLP:conf/acl/JacoviG20} emphasize that the faithfulness should be evaluated in grayscale instead of a binary term, i.e., faithful or not faithful. Following~\cite{DBLP:conf/emnlp/WiegreffeP19, DBLP:conf/acl/JacoviG20}, we consider the degree of faithfulness as it allows to identify the interpretation that is sufficiently faithful to be useful in
practice.

\begin{figure}[t]
	\begin{subfigure}{0.32\columnwidth}
		\includegraphics[width=\columnwidth]{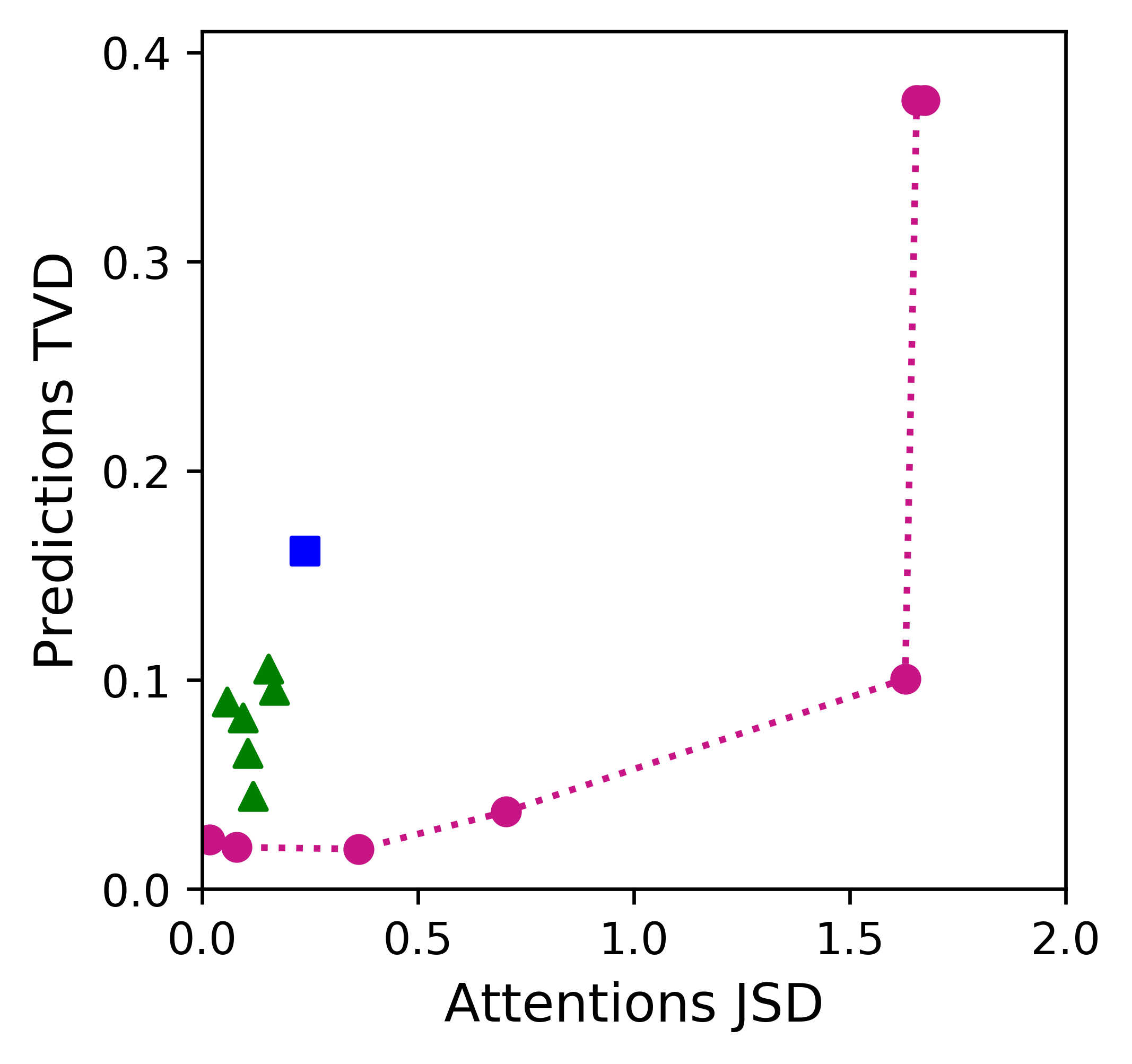}
		\caption{R24 C1}
		\label{Fig:eval_int_adv_24_1}
	\end{subfigure}
	\begin{subfigure}{0.32\columnwidth}
		\includegraphics[width=\columnwidth]{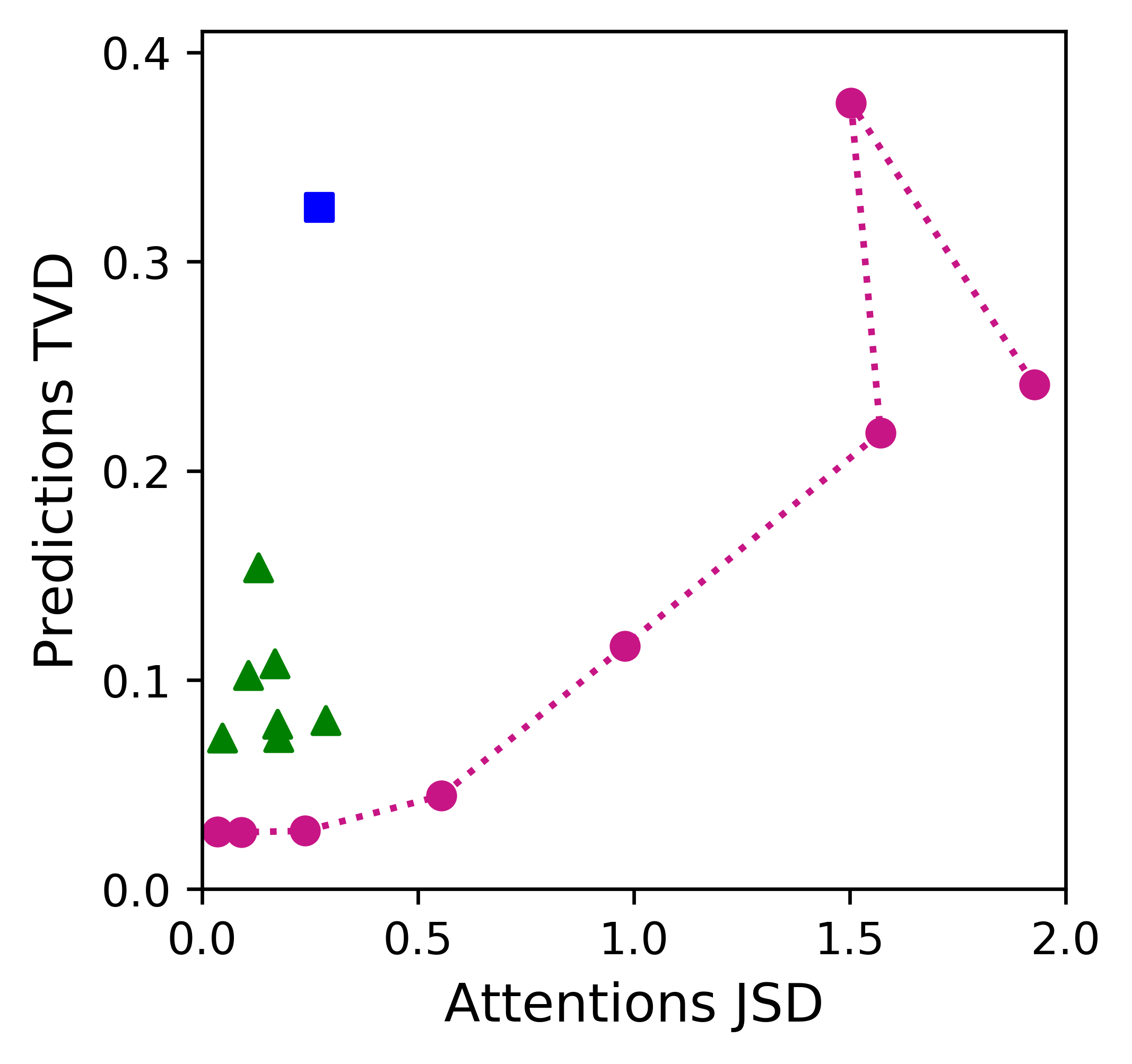}
		\caption{R27 C1}
		\label{Fig:eval_int_adv_27_1}
	\end{subfigure}
	\begin{subfigure}{0.32\columnwidth}
		\includegraphics[width=\columnwidth]{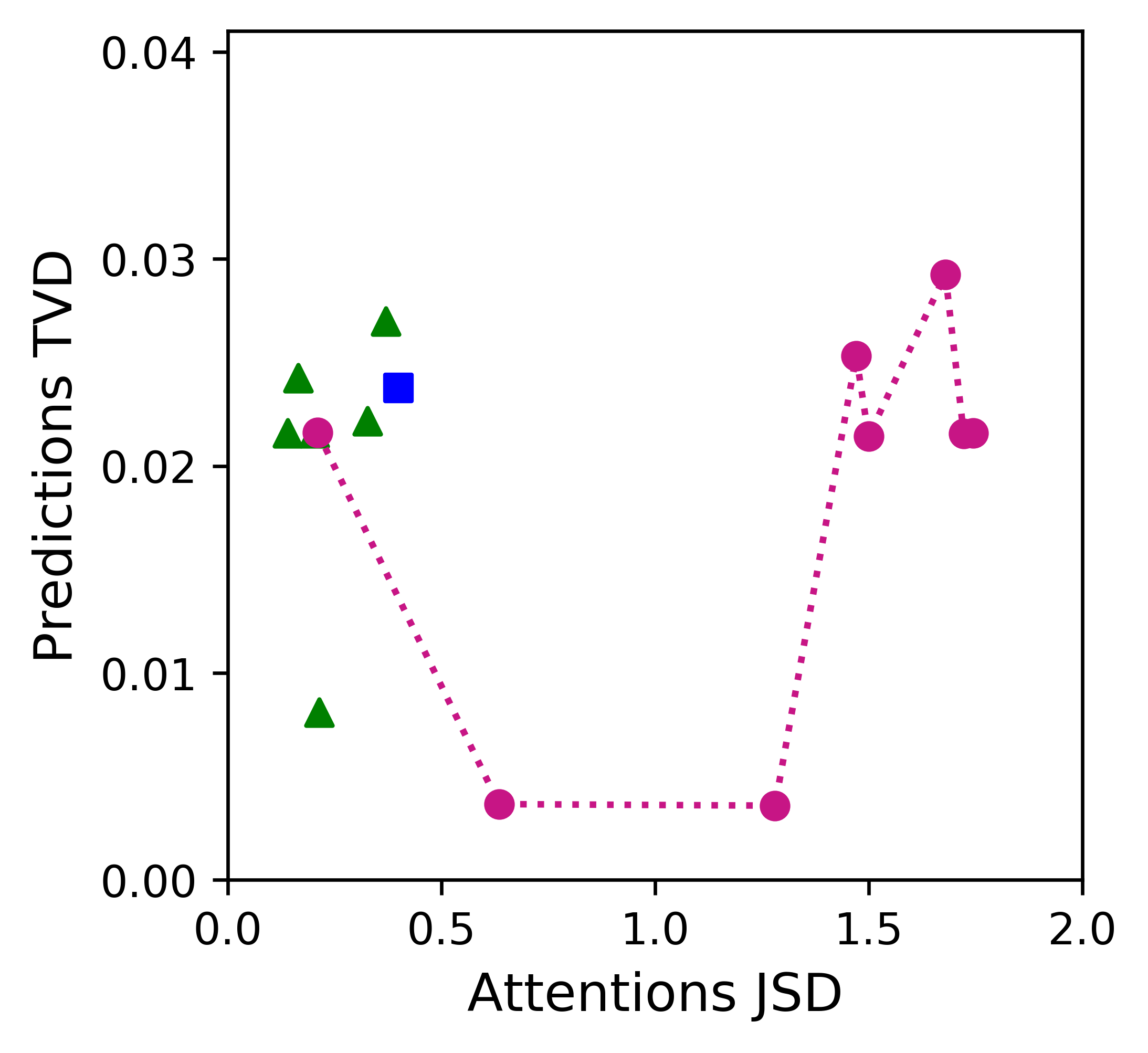}
		\caption{R0 C1}
		\label{Fig:eval_int_adv_0_1}
	\end{subfigure}
	\begin{subfigure}{0.32\columnwidth}
		\includegraphics[width=\columnwidth]{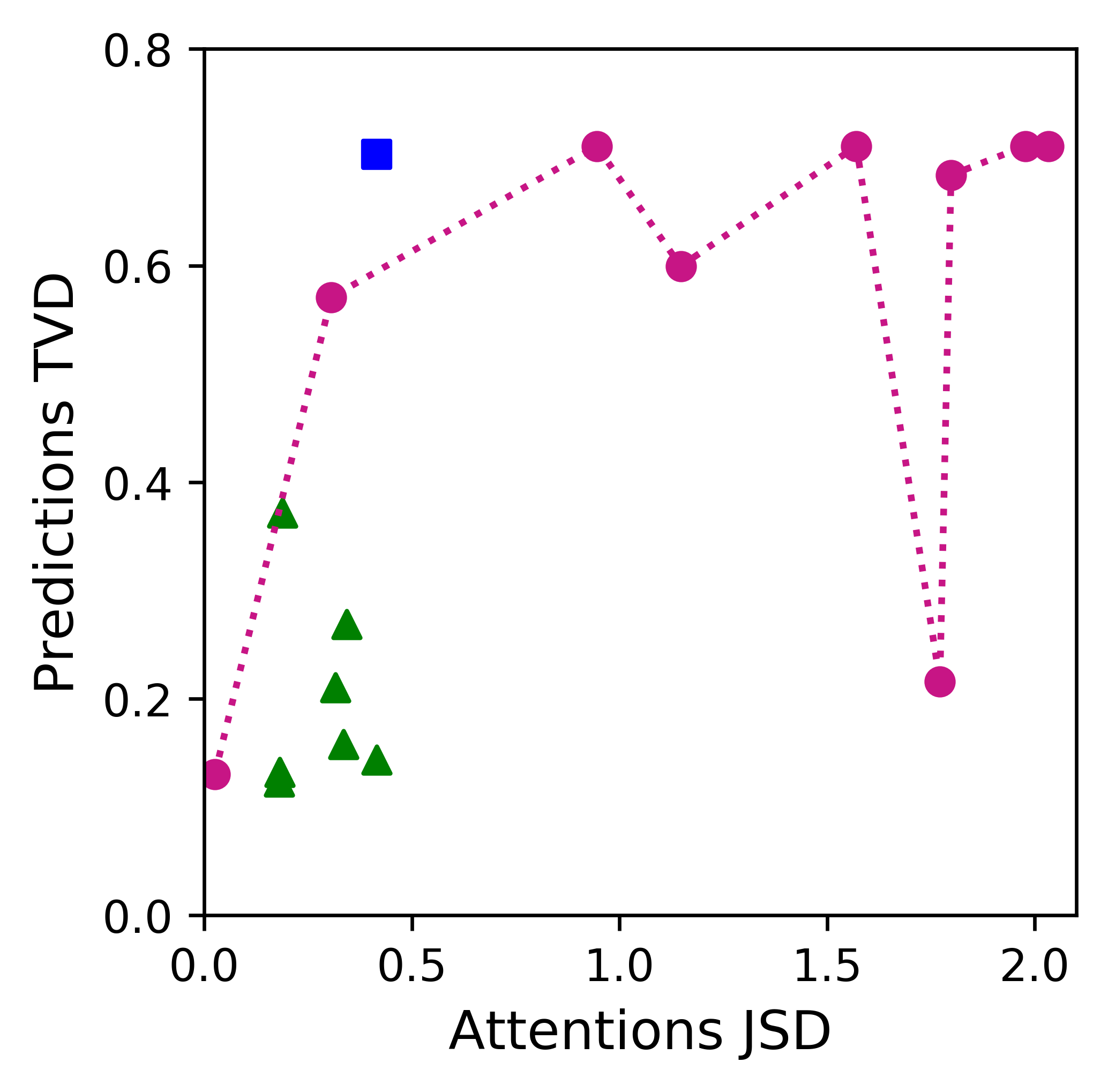}
		\caption{R31 C1}
		\label{Fig:eval_int_adv_31_1}
	\end{subfigure}
	\begin{subfigure}{0.32\columnwidth}
		\includegraphics[width=\columnwidth]{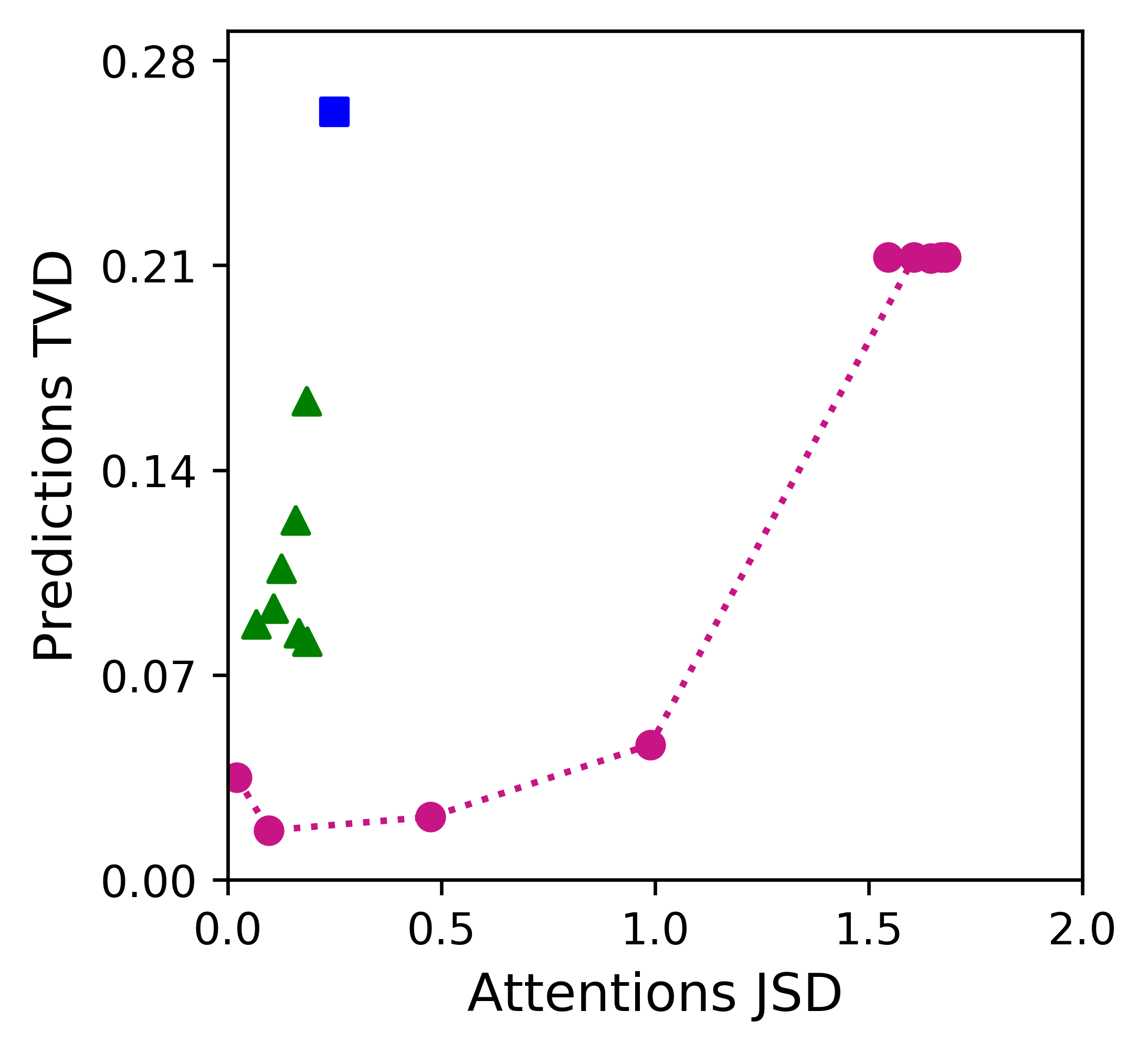}
		\caption{R70 C3}
		\label{Fig:eval_int_adv_70_3}
	\end{subfigure}
	\begin{subfigure}{0.32\columnwidth}
		\includegraphics[width=\columnwidth]{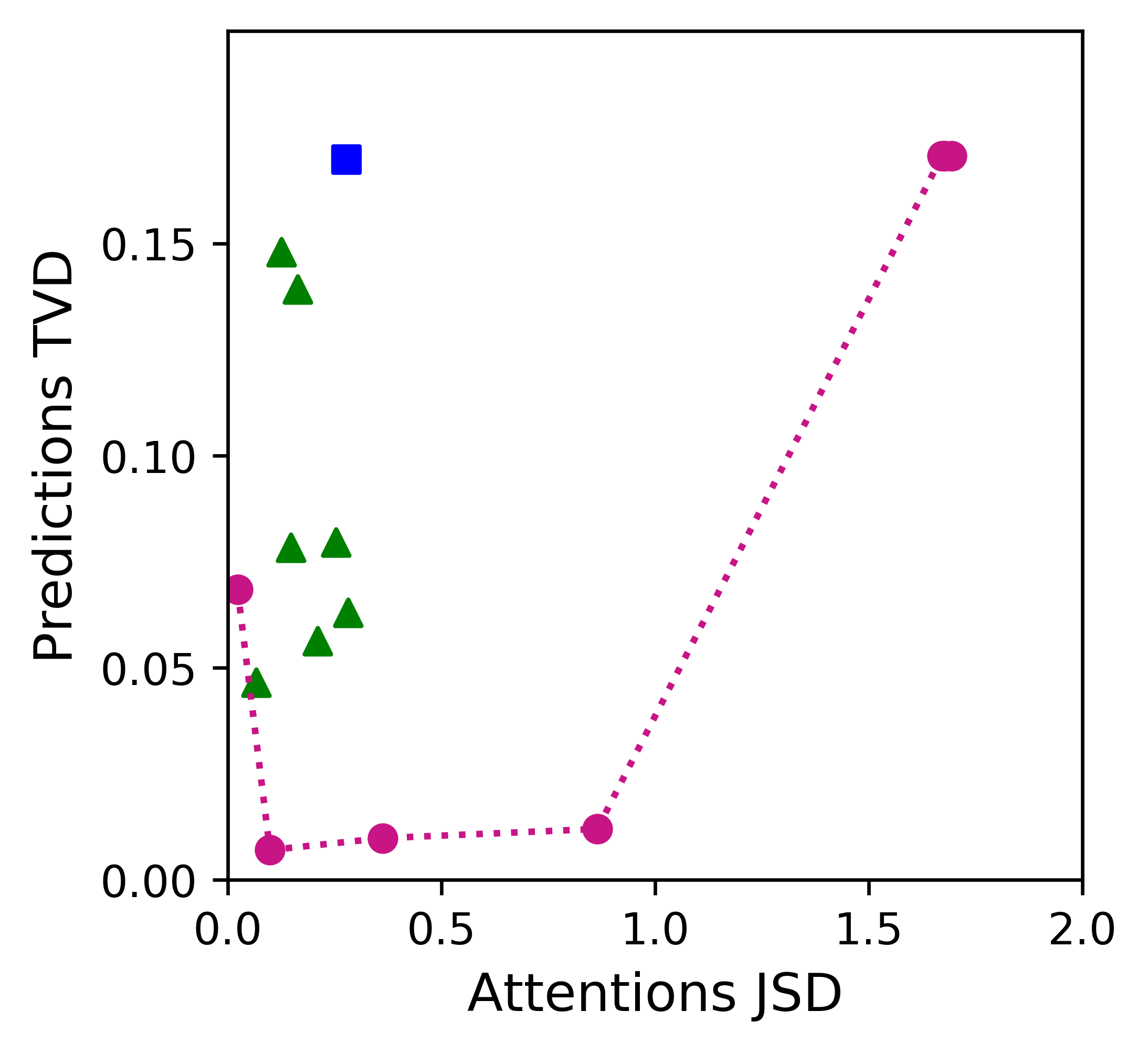}
		\caption{R43 C3}
		\label{Fig:eval_int_adv_43_3}
	\end{subfigure}
	\caption{Evaluation of interpretability (Averaged per-instance test set JSD and TVD from base model for each model variant. JSD is bounded at $\sim2.07$; \mytriangle{black!60!green}: random seed; \mysquare{blue} uniform weights; dotted line: our adversarial setup as $\lambda$ is varied)}
	\vspace{-1em}
	\label{Fig:eval_int_adv}
\end{figure}

Our process to generate the adversarial attention weights to establish the exclusivity, hence the faithfulness of the model is as follows. We train an adversarial model ($\mathcal{M}_{adv}$) with the objective of minimizing the prediction differences from our AIST model ($\mathcal{M}_{AIST}$) along with a divergent attention distribution for an instance $i$.
$$\mathcal{L}(\mathcal{M}_{AIST}, \mathcal{M}_{adv}) = \text{TVD}(\hat{y}_{AIST}^{(i)}, \hat{y}_{adv}^{(i)}) - \lambda \; \text{KL}(\alpha_{AIST}^{(i)}, \alpha_{adv}^{(i)})$$
Here, $\lambda$ is a hyperparameter that controls the tradeoff between TVD and JSD, where TVD is the levels of prediction variance and JSD (Jensen-Shannon Divergence) quantifies the difference between two attention distributions. In Figure~\ref{Fig:eval_int_adv}, we present the TVD between the predictions of the adversarial and AIST model against the increasing JSD between their attention distributions for a selected number of regions on specific crime categories.  We believe the graphs in Figure~\ref{Fig:eval_int_adv} to be representative of all $4$ crime categories across $77$ regions as they show all of the possible three cases: not faithful, moderately faithful and concretely faithful. Fast increase in the prediction difference concurs that the attention scores are not easily manipulable and exclusive. Hence, they can be used as faithful explanations. We also include the scores of uniform model variant (\mysquare{blue}) and random seed initialization (\mytriangle{black!60!green}) in these TVD vs JSD graphs. Figure~\ref{Fig:eval_int_adv_24_1}, ~\ref{Fig:eval_int_adv_27_1}, ~\ref{Fig:eval_int_adv_31_1}, ~\ref{Fig:eval_int_adv_70_3}, ~\ref{Fig:eval_int_adv_43_3} establish attentions as faithful explanations for the specified regions and crime categories as the increase in JSD comes at a high price of the increased TVD (at different rates). However, Figure~\ref{Fig:eval_int_adv_0_1} shows an example where the attention distributions generated by AIST can not be deemed faithful as it is easy to manipulate the attentions without losing much of the prediction performance. We include the predictions of the best adversarial models with instance-average JSD > 1 in Table~\ref{table:exp_result_int}.

Table~\ref{table:exp_result_int} shows the superiority of the base model (AIST) across all four crime categories over its uniform variant and adversarial models. Substantial increase of MAE scores of the uniform model suggests that attention is indeed a necessary component for better performance in the crime prediction. Similarly, a higher MAE score of the adversarial models ascertain the exclusivity of the predictions generated by AIST model.

\begin{small}
\begin{table*}[htbp]
	\caption{Comparison of AIST with its uniform variant and adversarial models on MAE}
	\centering
		\begin{tabular}[t]{|c|c|c|c|}
			\hline
			Crime Category & Uniform & Base Model & Adversarial\\
			\hline
			Theft (C1) & 0.9776 & \textbf{0.8747} & 1.1807\\
			
			\hline
			Criminal Damage (C2) & 0.3738 & \textbf{0.3615} & 0.3734 \\
			\hline
			Battery (C3) & 0.7206 & \textbf{0.6910} & 0.8029  \\
			\hline
			
			Narcotics (C4) & 0.3634 & \textbf{0.3399} & 0.5537 \\
			\hline
			
		\end{tabular}
	\label{table:exp_result_int}
\end{table*}
\end{small}

\subsection{Case Study} 
\begin{figure*}[t]
    \captionsetup[subfigure]{aboveskip=-5pt, belowskip=1pt}
	\begin{subfigure}{\textwidth}
	\def\svgwidth{0.9\textwidth}
	    \centering
	    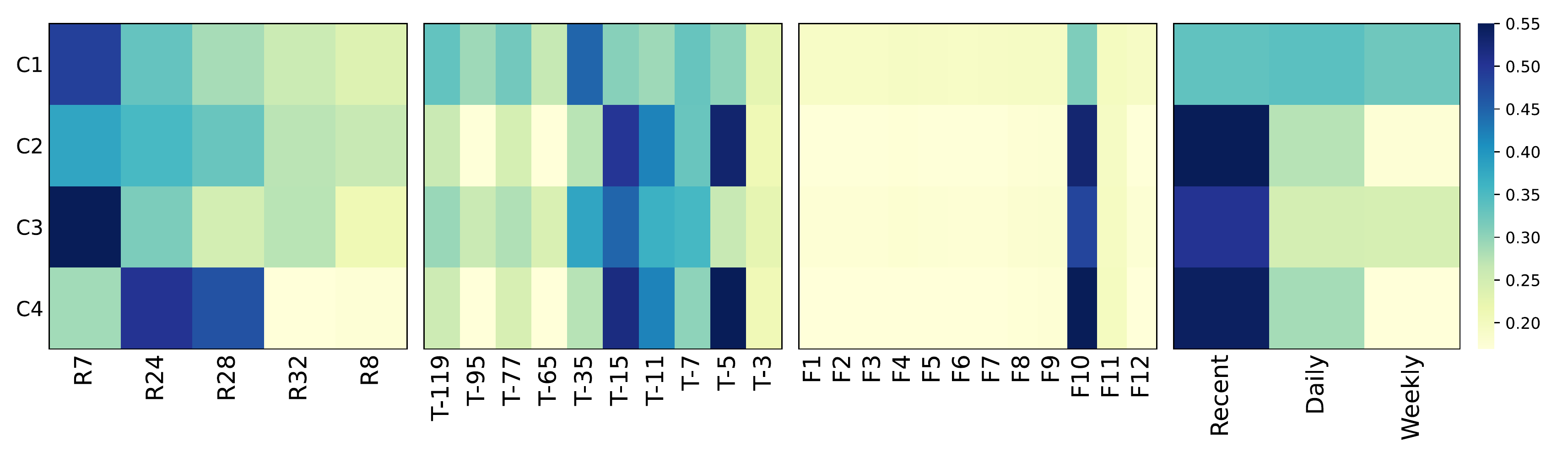
		\caption{R8 (Near North Side)}
        \label{Fig:case_8}
	\end{subfigure}
	\begin{subfigure}{\textwidth}
	\def\svgwidth{0.9\textwidth}
	    \centering
	    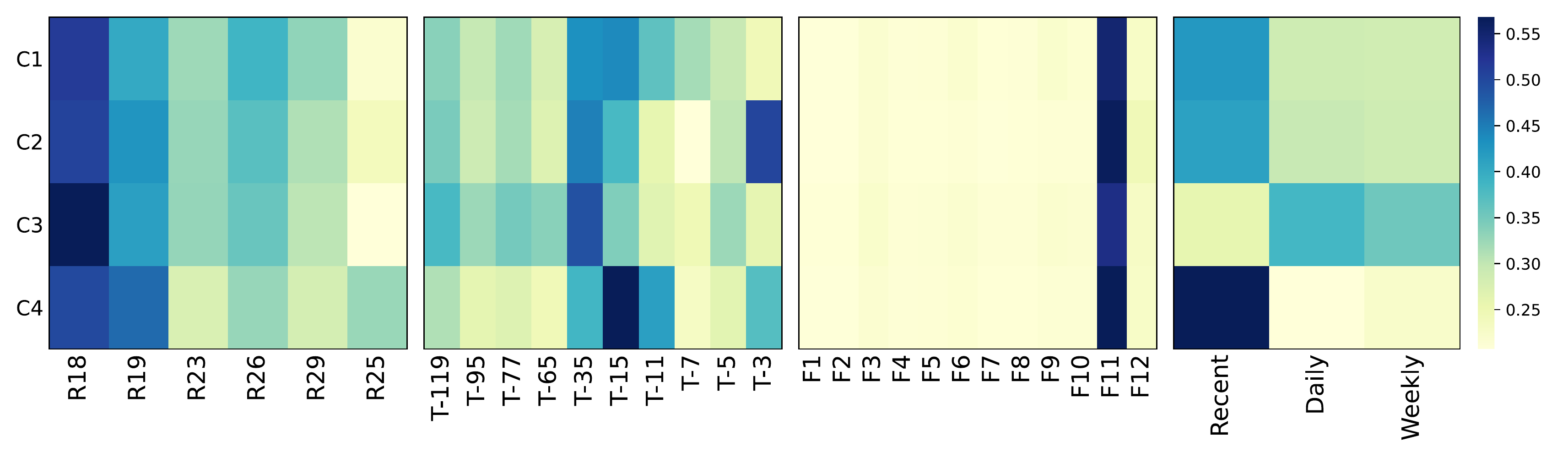
		\caption{R25 (Austin)}
        \label{Fig:case_25}
	\end{subfigure}
	\begin{subfigure}{\textwidth}
	 \def\svgwidth{0.9\textwidth}
	    \centering
	    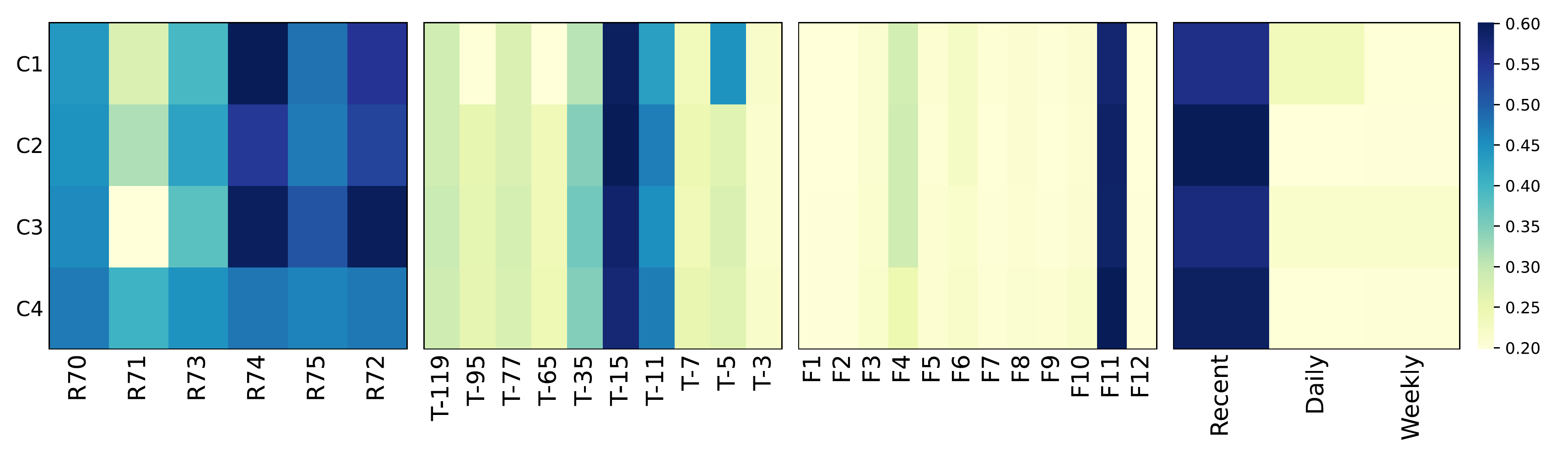
        \caption{R72 (Beverly)}
        \label{Fig:case_72}
    \end{subfigure}
	\caption {Case analysis of Region 8, 25 and 72}
	\label{Fig:case_8_25}
	\vspace{-3mm}
\end{figure*}

We select three communities for exploration: (i) R8 (Near North Side): situated in downtown Central Chicago and experiences high crime distribution, (ii) R25 (Austin): situated on the Western side of Chicago and is not as busy as R8, but has a high crime distribution and (iii) R72 (Beverly): located in Southern Chicago and is a quiet residential community with low crime rate. For each community, we randomly select 200 samples from test set and present the contribution of neighbor regions, POI and taxi flow features, trends and important time steps as a heat map in Figure~\ref{Fig:case_8_25}. We denote crimes category Theft, Criminal Damage, Battery and Narcotics with C1, C2, C3 and C4, respectively.
From Equation~\ref{eq:1}, the contribution coefficient of the crime occurrences of region $r_i'\in\mathcal{N}_i$ to the crime embedding of target region $r_i$ during time step $t$ can be calculated as, $\phi(\mathbf{{c}}_{i, t}^k, x_{i', t}^k) = \alpha_{ii'}\mathbf{w_x}{x_{i', t}^k}$. From Equation~\ref{eq:2}, the contribution coefficient of feature $j$ on target region can be calculated as $\phi(\mathbf{{e}}_{i, t}^k, f_{t}^j) = \beta_{ii'}^{j} \sum_{i' \in \mathcal{N}_i}\alpha_{ii'}  \mathbf{w_v}f_{i', t}^j$. Similarly, $\phi(\hat{y}_{i, T+1}^k, {h}_{T+1}^a) =\alpha_a \mathbf{w}\mathbf{h}_{T+1}^a$ denotes the contribution coefficient of recent, daily and periodic trends (Equations~\ref{eq:4},~\ref{eq:5}).


For R8, professional POIs (F10) made the highest contribution. R8 is a business region with thousands of jobs and has a large number of professional POIs. Thus it is expected that those POIs have large impact on it's crime embedding. On the other hand, R25 and R72 are residential regions. POI Shop (F11) contributed most for R25 and R72. Besides F11, Residence (F4) POIs also contributed for R72 for all categories except C4. Hence, our model learns both region and category specific influential features. An interesting observation for R8 is that though R8 has a large number of Food and Shop POIs, their contribution is almost none which signifies the quality of our model's prediction.
C1 shows strong long term temporal correlation for R8, whereas none of the crime category shows long term temporal correlation for R72 as the crime number for R72 is low. C3 in R25 shows a strong periodic correlation and unlike R8 and R72, it does not depend on recent crimes. C4 hardly present any long term correlation, which is intuitive. C2 shows daily periodicity in R8 and R25.



R7 and R24 have the most similar crime distribution as R8. However, these similarities vary with crime categories, e.g. for C1 and C3, R7 is given more attention whereas for C4, the attention shifts to R24 and R28. This is because both R24 and R28 experience large number of C4 crimes and have greater influence than R7. For R25, R18 is the most influential region across all crime categories. Unlike R8, the contribution of its neighboring regions are almost same except for R23 which is given less importance compared to other neighbors. The fact that R23 shares its boundary with different regions of different districts/sides makes their crime distribution less similar. R72 gives equal importance to each of its neighbors except R71. R71 has a higher number of crime occurrences than R72 and their crime distribution is quite different for all crime categories except C4. Thus, our model is able to capture diverse spatial correlation.
\section{Conclusion}
\label{sec:conclusion}
We propose AIST, a novel interpretable deep learning framework for crime prediction. AIST captures the dynamic spatio-temporal correlations based on the past crime occurrences, external features (e.g., traffic flow and POI information) and the recent and periodic crime trends. We develop two novel variants of GAT, $hGAT$ and $fGAT$ that allows AIST to improve prediction accuracy and provide the insights behind a prediction. Experiments and case studies on real-world Chicago crime data show that AIST outperforms the baseline models in terms of prediction accuracy and we can exploit attention weights associated with different parts of the model to interpret its prediction. On average, AIST shows a decrease of $8.3$\% on MAE and $20.98$\% on MSE over the state-of-the-art for crime prediction tasks. 
AIST also outperforms the high-performance spatio-temporal models~\cite{DBLP:conf/ijcai/LiangKZYZ18, DBLP:conf/ijcai/YuYZ18, DBLP:journals/TKDE/9139357} developed for solving different domain of tasks (e.g., geo-sensory time series, traffic or crowd flow prediction).  On average, AIST shows a decrease of $4.1\%$ on MAE and $7.45\%$ on MSE, when we customize these models for the crime prediction task.

Though we evaluate AIST for the crime prediction problem, AIST has the ability to learn an arbitrary function over the spatio-temporal-semantic space and can be adapted for any other spatio-temporal problem (e.g. traffic, citywide passenger demand, taxi demand prediction) that can benefit from incorporating semantically relevant information and knowing the interpretation of the prediction.


\bibliographystyle{ACM-Reference-Format}
\bibliography{sample-manuscript}
\end{document}

%% file: Figures/Community_Structure_v2.pdf_tex
\begingroup%
  \makeatletter%
  \providecommand\color[2][]{%
    \errmessage{(Inkscape) Color is used for the text in Inkscape, but the package 'color.sty' is not loaded}%
    \renewcommand\color[2][]{}%
  }%
  \providecommand\transparent[1]{%
    \errmessage{(Inkscape) Transparency is used (non-zero) for the text in Inkscape, but the package 'transparent.sty' is not loaded}%
    \renewcommand\transparent[1]{}%
  }%
  \providecommand\rotatebox[2]{#2}%
  \newcommand*\fsize{\dimexpr\f@size pt\relax}%
  \newcommand*\lineheight[1]{\fontsize{\fsize}{#1\fsize}\selectfont}%
  \ifx\svgwidth\undefined%
    \setlength{\unitlength}{96.09452376bp}%
    \ifx\svgscale\undefined%
      \relax%
    \else%
      \setlength{\unitlength}{\unitlength * \real{\svgscale}}%
    \fi%
  \else%
    \setlength{\unitlength}{\svgwidth}%
  \fi%
  \global\let\svgwidth\undefined%
  \global\let\svgscale\undefined%
  \makeatother%
  \begin{picture}(1,0.79598188)%
    \lineheight{1}%
    \setlength\tabcolsep{0pt}%
    \put(0,0){\includegraphics[width=\unitlength,page=1]{Community_Structure_v2.pdf}}%
    \put(0.47744466,0.3665179){\color[rgb]{0,0,0}\makebox(0,0)[lt]{\lineheight{1.25}\smash{\begin{tabular}[t]{l}\small{$8$}\end{tabular}}}}%
    \put(0.05718627,0.65083691){\color[rgb]{0,0,0}\makebox(0,0)[lt]{\lineheight{1.25}\smash{\begin{tabular}[t]{l}\small{$24$}\end{tabular}}}}%
    \put(0.05933038,0.07619649){\color[rgb]{0,0,0}\makebox(0,0)[lt]{\lineheight{1.25}\smash{\begin{tabular}[t]{l}\small{$28$}\end{tabular}}}}%
    \put(0.87924407,0.66257066){\color[rgb]{0,0,0}\makebox(0,0)[lt]{\lineheight{1.25}\smash{\begin{tabular}[t]{l}\small{$7$}\end{tabular}}}}%
    \put(0.85282922,0.07305649){\color[rgb]{0,0,0}\makebox(0,0)[lt]{\lineheight{1.25}\smash{\begin{tabular}[t]{l}\small{$32$}\end{tabular}}}}%
  \end{picture}%
\endgroup%

%% file: sample-manuscript.bbl

\begin{thebibliography}{68}


\ifx \showCODEN    \undefined \def \showCODEN     #1{\unskip}     \fi
\ifx \showDOI      \undefined \def \showDOI       #1{#1}\fi
\ifx \showISBNx    \undefined \def \showISBNx     #1{\unskip}     \fi
\ifx \showISBNxiii \undefined \def \showISBNxiii  #1{\unskip}     \fi
\ifx \showISSN     \undefined \def \showISSN      #1{\unskip}     \fi
\ifx \showLCCN     \undefined \def \showLCCN      #1{\unskip}     \fi
\ifx \shownote     \undefined \def \shownote      #1{#1}          \fi
\ifx \showarticletitle \undefined \def \showarticletitle #1{#1}   \fi
\ifx \showURL      \undefined \def \showURL       {\relax}        \fi
\providecommand\bibfield[2]{#2}
\providecommand\bibinfo[2]{#2}
\providecommand\natexlab[1]{#1}
\providecommand\showeprint[2][]{arXiv:#2}

\bibitem[\protect\citeauthoryear{Ba, Mnih, and Kavukcuoglu}{Ba
  et~al\mbox{.}}{2015}]%
        {DBLP:journals/corr/BaMK14}
\bibfield{author}{\bibinfo{person}{Jimmy Ba}, \bibinfo{person}{Volodymyr Mnih},
  {and} \bibinfo{person}{Koray Kavukcuoglu}.} \bibinfo{year}{2015}\natexlab{}.
\newblock \showarticletitle{Multiple Object Recognition with Visual Attention}.
  In \bibinfo{booktitle}{\emph{{ICLR}}}.
\newblock


\bibitem[\protect\citeauthoryear{Bahdanau, Cho, and Bengio}{Bahdanau
  et~al\mbox{.}}{2015}]%
        {DBLP:journals/corr/BahdanauCB14}
\bibfield{author}{\bibinfo{person}{Dzmitry Bahdanau},
  \bibinfo{person}{Kyunghyun Cho}, {and} \bibinfo{person}{Yoshua Bengio}.}
  \bibinfo{year}{2015}\natexlab{}.
\newblock \showarticletitle{Neural Machine Translation by Jointly Learning to
  Align and Translate}. In \bibinfo{booktitle}{\emph{{ICLR}}}.
\newblock


\bibitem[\protect\citeauthoryear{Bai, Zhang, Egleston, and Vucetic}{Bai
  et~al\mbox{.}}{2018}]%
        {DBLP:conf/kdd/BaiZEV18}
\bibfield{author}{\bibinfo{person}{Tian Bai}, \bibinfo{person}{Shanshan Zhang},
  \bibinfo{person}{Brian~L. Egleston}, {and} \bibinfo{person}{Slobodan
  Vucetic}.} \bibinfo{year}{2018}\natexlab{}.
\newblock \showarticletitle{Interpretable Representation Learning for
  Healthcare via Capturing Disease Progression through Time}. In
  \bibinfo{booktitle}{\emph{{SIGKDD}}}. \bibinfo{publisher}{{ACM}},
  \bibinfo{pages}{43--51}.
\newblock


\bibitem[\protect\citeauthoryear{Breiman, Friedman, Olshen, and Stone}{Breiman
  et~al\mbox{.}}{1984}]%
        {DBLP:books/wa/BreimanFOS84}
\bibfield{author}{\bibinfo{person}{Leo Breiman}, \bibinfo{person}{J.~H.
  Friedman}, \bibinfo{person}{R.~A. Olshen}, {and} \bibinfo{person}{C.~J.
  Stone}.} \bibinfo{year}{1984}\natexlab{}.
\newblock \bibinfo{booktitle}{\emph{Classification and Regression Trees}}.
\newblock \bibinfo{publisher}{Wadsworth}.
\newblock


\bibitem[\protect\citeauthoryear{Buczak and Gifford}{Buczak and
  Gifford}{2010}]%
        {10.1145/1938606.1938608}
\bibfield{author}{\bibinfo{person}{Anna~L. Buczak} {and}
  \bibinfo{person}{Christopher~M. Gifford}.} \bibinfo{year}{2010}\natexlab{}.
\newblock \showarticletitle{Fuzzy Association Rule Mining for Community Crime
  Pattern Discovery}. In \bibinfo{booktitle}{\emph{ACM SIGKDD Workshop on
  Intelligence and Security Informatics}}. \bibinfo{publisher}{Association for
  Computing Machinery}.
\newblock


\bibitem[\protect\citeauthoryear{Chainey, Tompson, and Uhlig}{Chainey
  et~al\mbox{.}}{2008}]%
        {article}
\bibfield{author}{\bibinfo{person}{Spencer Chainey}, \bibinfo{person}{Lisa
  Tompson}, {and} \bibinfo{person}{Sebastian Uhlig}.}
  \bibinfo{year}{2008}\natexlab{}.
\newblock \showarticletitle{The Utility of Hotspot Mapping for Predicting
  Spatial Patterns of Crime}.
\newblock \bibinfo{journal}{\emph{Security Journal}}  \bibinfo{volume}{21}
  (\bibinfo{year}{2008}), \bibinfo{pages}{4--28}.
\newblock


\bibitem[\protect\citeauthoryear{Chen, Li, Teo, Chen, Zou, Yang, Vijay, Feng,
  and Zeng}{Chen et~al\mbox{.}}{2018a}]%
        {DBLP:conf/icdm/ChenLTCZYVFZ18}
\bibfield{author}{\bibinfo{person}{Cen Chen}, \bibinfo{person}{Kenli Li},
  \bibinfo{person}{Sin~G. Teo}, \bibinfo{person}{Guizi Chen},
  \bibinfo{person}{Xiaofeng Zou}, \bibinfo{person}{Xulei Yang},
  \bibinfo{person}{Ramaseshan~C. Vijay}, \bibinfo{person}{Jiashi Feng}, {and}
  \bibinfo{person}{Zeng Zeng}.} \bibinfo{year}{2018}\natexlab{a}.
\newblock \showarticletitle{Exploiting Spatio-Temporal Correlations with
  Multiple 3D Convolutional Neural Networks for Citywide Vehicle Flow
  Prediction}. In \bibinfo{booktitle}{\emph{{ICDM}}}.
  \bibinfo{publisher}{{IEEE} Computer Society}, \bibinfo{pages}{893--898}.
\newblock


\bibitem[\protect\citeauthoryear{Chen, Li, Tao, Barnett, Rudin, and Su}{Chen
  et~al\mbox{.}}{2019}]%
        {DBLP:conf/nips/ChenLTBRS19}
\bibfield{author}{\bibinfo{person}{Chaofan Chen}, \bibinfo{person}{Oscar Li},
  \bibinfo{person}{Daniel Tao}, \bibinfo{person}{Alina Barnett},
  \bibinfo{person}{Cynthia Rudin}, {and} \bibinfo{person}{Jonathan Su}.}
  \bibinfo{year}{2019}\natexlab{}.
\newblock \showarticletitle{This Looks Like That: Deep Learning for
  Interpretable Image Recognition}. In \bibinfo{booktitle}{\emph{{NeurIPS}}}.
  \bibinfo{pages}{8928--8939}.
\newblock


\bibitem[\protect\citeauthoryear{Chen, Yu, and Liu}{Chen
  et~al\mbox{.}}{2018b}]%
        {DBLP:journals/tits/ChenYL18}
\bibfield{author}{\bibinfo{person}{Meng Chen}, \bibinfo{person}{Xiaohui Yu},
  {and} \bibinfo{person}{Yang Liu}.} \bibinfo{year}{2018}\natexlab{b}.
\newblock \showarticletitle{{PCNN:} Deep Convolutional Networks for Short-Term
  Traffic Congestion Prediction}.
\newblock \bibinfo{journal}{\emph{{IEEE} Trans. Intell. Transp. Syst.}}
  \bibinfo{volume}{19}, \bibinfo{number}{11} (\bibinfo{year}{2018}),
  \bibinfo{pages}{3550--3559}.
\newblock


\bibitem[\protect\citeauthoryear{{Chen}, {Yuan}, and {Shu}}{{Chen}
  et~al\mbox{.}}{2008}]%
        {4666600}
\bibfield{author}{\bibinfo{person}{P. {Chen}}, \bibinfo{person}{H. {Yuan}},
  {and} \bibinfo{person}{X. {Shu}}.} \bibinfo{year}{2008}\natexlab{}.
\newblock \showarticletitle{Forecasting Crime Using the ARIMA Model}. In
  \bibinfo{booktitle}{\emph{FSKD}}. \bibinfo{pages}{627--630}.
\newblock


\bibitem[\protect\citeauthoryear{Choi, Bahadori, Song, Stewart, and Sun}{Choi
  et~al\mbox{.}}{2017}]%
        {DBLP:conf/kdd/ChoiBSSS17}
\bibfield{author}{\bibinfo{person}{Edward Choi}, \bibinfo{person}{Mohammad~Taha
  Bahadori}, \bibinfo{person}{Le Song}, \bibinfo{person}{Walter~F. Stewart},
  {and} \bibinfo{person}{Jimeng Sun}.} \bibinfo{year}{2017}\natexlab{}.
\newblock \showarticletitle{{GRAM:} Graph-based Attention Model for Healthcare
  Representation Learning}. In \bibinfo{booktitle}{\emph{{SIGKDD}}}.
  \bibinfo{publisher}{{ACM}}, \bibinfo{pages}{787--795}.
\newblock


\bibitem[\protect\citeauthoryear{Choi, Bahadori, Sun, Kulas, Schuetz, and
  Stewart}{Choi et~al\mbox{.}}{2016}]%
        {DBLP:conf/nips/ChoiBSKSS16}
\bibfield{author}{\bibinfo{person}{Edward Choi}, \bibinfo{person}{Mohammad~Taha
  Bahadori}, \bibinfo{person}{Jimeng Sun}, \bibinfo{person}{Joshua Kulas},
  \bibinfo{person}{Andy Schuetz}, {and} \bibinfo{person}{Walter~F. Stewart}.}
  \bibinfo{year}{2016}\natexlab{}.
\newblock \showarticletitle{{RETAIN:} An Interpretable Predictive Model for
  Healthcare using Reverse Time Attention Mechanism}. In
  \bibinfo{booktitle}{\emph{{NeurIPS}}}. \bibinfo{pages}{3504--3512}.
\newblock


\bibitem[\protect\citeauthoryear{Crimes}{Crimes}{2019}]%
        {chicago_crime2019}
\bibfield{author}{\bibinfo{person}{Chicago Crimes}.}
  \bibinfo{year}{2019}\natexlab{}.
\newblock \bibinfo{title}{City of Chicago Data Portal}.
\newblock
  \bibinfo{howpublished}{\url{https://data.cityofchicago.org/Public-Safety/Crimes-2019/w98m-zvie}}.
\newblock


\bibitem[\protect\citeauthoryear{Cui, Ke, and Wang}{Cui et~al\mbox{.}}{2016}]%
        {DBLP:journals/corr/abs-1801-02143}
\bibfield{author}{\bibinfo{person}{Zhiyong Cui}, \bibinfo{person}{Ruimin Ke},
  {and} \bibinfo{person}{Yinhai Wang}.} \bibinfo{year}{2016}\natexlab{}.
\newblock \showarticletitle{Deep Stacked Bidirectional and Unidirectional LSTM
  Recurrent Neural Network for Network-wide Traffic Speed Prediction}.
\newblock  (\bibinfo{year}{2016}).
\newblock


\bibitem[\protect\citeauthoryear{Dabkowski and Gal}{Dabkowski and Gal}{2017}]%
        {DBLP:conf/nips/DabkowskiG17}
\bibfield{author}{\bibinfo{person}{Piotr Dabkowski} {and}
  \bibinfo{person}{Yarin Gal}.} \bibinfo{year}{2017}\natexlab{}.
\newblock \showarticletitle{Real Time Image Saliency for Black Box
  Classifiers}. In \bibinfo{booktitle}{\emph{{NeurIPS}}}.
  \bibinfo{pages}{6967--6976}.
\newblock


\bibitem[\protect\citeauthoryear{de~Queiroz~Neto, dos Santos, and
  Vidal}{de~Queiroz~Neto et~al\mbox{.}}{2016}]%
        {DBLP:conf/sibgrapi/NetoSV16}
\bibfield{author}{\bibinfo{person}{Jose~Florencio de Queiroz~Neto},
  \bibinfo{person}{Emanuele~Marques dos Santos}, {and}
  \bibinfo{person}{Creto~Augusto Vidal}.} \bibinfo{year}{2016}\natexlab{}.
\newblock \showarticletitle{{MSKDE} - Using Marching Squares to Quickly Make
  High Quality Crime Hotspot Maps}. In \bibinfo{booktitle}{\emph{{SIBGRAPI}}}.
  \bibinfo{publisher}{{IEEE} Computer Society}, \bibinfo{pages}{305--312}.
\newblock


\bibitem[\protect\citeauthoryear{Eck, Chainey, Cameron, and Wilson}{Eck
  et~al\mbox{.}}{2005}]%
        {eck2005mapping}
\bibfield{author}{\bibinfo{person}{John Eck}, \bibinfo{person}{Spencer
  Chainey}, \bibinfo{person}{James Cameron}, {and} \bibinfo{person}{Ronald
  Wilson}.} \bibinfo{year}{2005}\natexlab{}.
\newblock \showarticletitle{Mapping crime: Understanding hotspots}.
\newblock  (\bibinfo{year}{2005}).
\newblock


\bibitem[\protect\citeauthoryear{Ehsan, Tambwekar, Chan, Harrison, and
  Riedl}{Ehsan et~al\mbox{.}}{2019}]%
        {DBLP:conf/iui/EhsanTCHR19}
\bibfield{author}{\bibinfo{person}{Upol Ehsan}, \bibinfo{person}{Pradyumna
  Tambwekar}, \bibinfo{person}{Larry Chan}, \bibinfo{person}{Brent Harrison},
  {and} \bibinfo{person}{Mark~O. Riedl}.} \bibinfo{year}{2019}\natexlab{}.
\newblock \showarticletitle{Automated rationale generation: a technique for
  explainable {AI} and its effects on human perceptions}. In
  \bibinfo{booktitle}{\emph{{IUI}}}. \bibinfo{publisher}{{ACM}},
  \bibinfo{pages}{263--274}.
\newblock


\bibitem[\protect\citeauthoryear{Fu, Zheng, and Mei}{Fu et~al\mbox{.}}{2017}]%
        {DBLP:conf/cvpr/FuZM17}
\bibfield{author}{\bibinfo{person}{Jianlong Fu}, \bibinfo{person}{Heliang
  Zheng}, {and} \bibinfo{person}{Tao Mei}.} \bibinfo{year}{2017}\natexlab{}.
\newblock \showarticletitle{Look Closer to See Better: Recurrent Attention
  Convolutional Neural Network for Fine-Grained Image Recognition}. In
  \bibinfo{booktitle}{\emph{{CVPR}}}. \bibinfo{publisher}{{IEEE} Computer
  Society}, \bibinfo{pages}{4476--4484}.
\newblock


\bibitem[\protect\citeauthoryear{Gerber}{Gerber}{2014}]%
        {DBLP:journals/dss/Gerber14}
\bibfield{author}{\bibinfo{person}{Matthew~S. Gerber}.}
  \bibinfo{year}{2014}\natexlab{}.
\newblock \showarticletitle{Predicting crime using Twitter and kernel density
  estimation}.
\newblock \bibinfo{journal}{\emph{Decis. Support Syst.}}  \bibinfo{volume}{61}
  (\bibinfo{year}{2014}), \bibinfo{pages}{115--125}.
\newblock


\bibitem[\protect\citeauthoryear{Guo, Lin, Feng, Song, and Wan}{Guo
  et~al\mbox{.}}{2019}]%
        {DBLP:conf/aaai/GuoLFSW19}
\bibfield{author}{\bibinfo{person}{Shengnan Guo}, \bibinfo{person}{Youfang
  Lin}, \bibinfo{person}{Ning Feng}, \bibinfo{person}{Chao Song}, {and}
  \bibinfo{person}{Huaiyu Wan}.} \bibinfo{year}{2019}\natexlab{}.
\newblock \showarticletitle{Attention Based Spatial-Temporal Graph
  Convolutional Networks for Traffic Flow Forecasting}. In
  \bibinfo{booktitle}{\emph{{AAAI}}}. \bibinfo{pages}{922--929}.
\newblock


\bibitem[\protect\citeauthoryear{Hart and Zandbergen}{Hart and
  Zandbergen}{2014}]%
        {DBLP:journals/Hart}
\bibfield{author}{\bibinfo{person}{Timothy Hart} {and} \bibinfo{person}{Paul
  Zandbergen}.} \bibinfo{year}{2014}\natexlab{}.
\newblock \showarticletitle{Kernel density estimation and hotspot mapping:
  Examining the influence of interpolation method, grid cell size, and
  bandwidth on crime forecasting}.
\newblock \bibinfo{journal}{\emph{Policing: An International Journal of Police
  Strategies and Management}}  \bibinfo{volume}{37} (\bibinfo{date}{05}
  \bibinfo{year}{2014}).
\newblock


\bibitem[\protect\citeauthoryear{Hong, Lin, Yang, Li, Fu, Wang, Qie, and
  Ye}{Hong et~al\mbox{.}}{2020}]%
        {DBLP:conf/kdd/HongLYLFWQY20}
\bibfield{author}{\bibinfo{person}{Huiting Hong}, \bibinfo{person}{Yucheng
  Lin}, \bibinfo{person}{Xiaoqing Yang}, \bibinfo{person}{Zang Li},
  \bibinfo{person}{Kung Fu}, \bibinfo{person}{Zheng Wang},
  \bibinfo{person}{Xiaohu Qie}, {and} \bibinfo{person}{Jieping Ye}.}
  \bibinfo{year}{2020}\natexlab{}.
\newblock \showarticletitle{HetETA: Heterogeneous Information Network Embedding
  for Estimating Time of Arrival}. In \bibinfo{booktitle}{\emph{{KDD}}}.
  \bibinfo{publisher}{{ACM}}, \bibinfo{pages}{2444--2454}.
\newblock


\bibitem[\protect\citeauthoryear{Huang, Zhang, Zhao, Wu, Chawla, and Yin}{Huang
  et~al\mbox{.}}{2019}]%
        {DBLP:conf/www/HuangZZWCY19}
\bibfield{author}{\bibinfo{person}{Chao Huang}, \bibinfo{person}{Chuxu Zhang},
  \bibinfo{person}{Jiashu Zhao}, \bibinfo{person}{Xian Wu},
  \bibinfo{person}{Nitesh~V. Chawla}, {and} \bibinfo{person}{Dawei Yin}.}
  \bibinfo{year}{2019}\natexlab{}.
\newblock \showarticletitle{MiST: {A} Multiview and Multimodal Spatial-Temporal
  Learning Framework for Citywide Abnormal Event Forecasting}. In
  \bibinfo{booktitle}{\emph{{WWW}}}. \bibinfo{publisher}{{ACM}},
  \bibinfo{pages}{717--728}.
\newblock


\bibitem[\protect\citeauthoryear{Huang, Zhang, Zheng, and Chawla}{Huang
  et~al\mbox{.}}{2018}]%
        {DBLP:conf/cikm/HuangZZC18}
\bibfield{author}{\bibinfo{person}{Chao Huang}, \bibinfo{person}{Junbo Zhang},
  \bibinfo{person}{Yu Zheng}, {and} \bibinfo{person}{Nitesh~V. Chawla}.}
  \bibinfo{year}{2018}\natexlab{}.
\newblock \showarticletitle{DeepCrime: Attentive Hierarchical Recurrent
  Networks for Crime Prediction}. In \bibinfo{booktitle}{\emph{{CIKM}}}.
  \bibinfo{publisher}{{ACM}}, \bibinfo{pages}{1423--1432}.
\newblock


\bibitem[\protect\citeauthoryear{Jacovi and Goldberg}{Jacovi and
  Goldberg}{2020}]%
        {DBLP:conf/acl/JacoviG20}
\bibfield{author}{\bibinfo{person}{Alon Jacovi} {and} \bibinfo{person}{Yoav
  Goldberg}.} \bibinfo{year}{2020}\natexlab{}.
\newblock \showarticletitle{Towards Faithfully Interpretable {NLP} Systems: How
  Should We Define and Evaluate Faithfulness?}. In
  \bibinfo{booktitle}{\emph{{ACL}}}. \bibinfo{publisher}{Association for
  Computational Linguistics}, \bibinfo{pages}{4198--4205}.
\newblock


\bibitem[\protect\citeauthoryear{Jain and Wallace}{Jain and Wallace}{2019}]%
        {DBLP:conf/naacl/JainW19}
\bibfield{author}{\bibinfo{person}{Sarthak Jain} {and}
  \bibinfo{person}{Byron~C. Wallace}.} \bibinfo{year}{2019}\natexlab{}.
\newblock \showarticletitle{Attention is not Explanation}. In
  \bibinfo{booktitle}{\emph{{NAACL-HLT}}}. \bibinfo{publisher}{Association for
  Computational Linguistics}, \bibinfo{pages}{3543--3556}.
\newblock


\bibitem[\protect\citeauthoryear{Kaur, Nori, Jenkins, Caruana, Wallach, and
  Vaughan}{Kaur et~al\mbox{.}}{2020}]%
        {DBLP:conf/chi/KaurNJCWV20}
\bibfield{author}{\bibinfo{person}{Harmanpreet Kaur}, \bibinfo{person}{Harsha
  Nori}, \bibinfo{person}{Samuel Jenkins}, \bibinfo{person}{Rich Caruana},
  \bibinfo{person}{Hanna~M. Wallach}, {and} \bibinfo{person}{Jennifer~Wortman
  Vaughan}.} \bibinfo{year}{2020}\natexlab{}.
\newblock \showarticletitle{Interpreting Interpretability: Understanding Data
  Scientists' Use of Interpretability Tools for Machine Learning}. In
  \bibinfo{booktitle}{\emph{{CHI}}}. \bibinfo{publisher}{{ACM}},
  \bibinfo{pages}{1--14}.
\newblock


\bibitem[\protect\citeauthoryear{Ke, Goyal, Bilaniuk, Binas, Mozer, Pal, and
  Bengio}{Ke et~al\mbox{.}}{2018}]%
        {DBLP:conf/nips/KeGBBMPB18}
\bibfield{author}{\bibinfo{person}{Nan~Rosemary Ke}, \bibinfo{person}{Anirudh
  Goyal}, \bibinfo{person}{Olexa Bilaniuk}, \bibinfo{person}{Jonathan Binas},
  \bibinfo{person}{Michael~C. Mozer}, \bibinfo{person}{Chris Pal}, {and}
  \bibinfo{person}{Yoshua Bengio}.} \bibinfo{year}{2018}\natexlab{}.
\newblock \showarticletitle{Sparse Attentive Backtracking: Temporal Credit
  Assignment Through Reminding}. In \bibinfo{booktitle}{\emph{{NeurIPS}}}.
  \bibinfo{pages}{7651--7662}.
\newblock


\bibitem[\protect\citeauthoryear{Kipf and Welling}{Kipf and Welling}{2017}]%
        {DBLP:conf/iclr/KipfW17}
\bibfield{author}{\bibinfo{person}{Thomas~N. Kipf} {and} \bibinfo{person}{Max
  Welling}.} \bibinfo{year}{2017}\natexlab{}.
\newblock \showarticletitle{Semi-Supervised Classification with Graph
  Convolutional Networks}. In \bibinfo{booktitle}{\emph{{ICLR}}}.
  \bibinfo{publisher}{OpenReview.net}.
\newblock


\bibitem[\protect\citeauthoryear{Lei, Barzilay, and Jaakkola}{Lei
  et~al\mbox{.}}{2016}]%
        {DBLP:conf/emnlp/LeiBJ16}
\bibfield{author}{\bibinfo{person}{Tao Lei}, \bibinfo{person}{Regina Barzilay},
  {and} \bibinfo{person}{Tommi~S. Jaakkola}.} \bibinfo{year}{2016}\natexlab{}.
\newblock \showarticletitle{Rationalizing Neural Predictions}. In
  \bibinfo{booktitle}{\emph{{EMNLP}}}. \bibinfo{publisher}{The Association for
  Computational Linguistics}, \bibinfo{pages}{107--117}.
\newblock


\bibitem[\protect\citeauthoryear{Li, Zhu, Kong, Xu, and Zhao}{Li
  et~al\mbox{.}}{2019}]%
        {DBLP:conf/aaai/LiZKXZ19}
\bibfield{author}{\bibinfo{person}{Youru Li}, \bibinfo{person}{Zhenfeng Zhu},
  \bibinfo{person}{Deqiang Kong}, \bibinfo{person}{Meixiang Xu}, {and}
  \bibinfo{person}{Yao Zhao}.} \bibinfo{year}{2019}\natexlab{}.
\newblock \showarticletitle{Learning Heterogeneous Spatial-Temporal
  Representation for Bike-Sharing Demand Prediction}. In
  \bibinfo{booktitle}{\emph{{AAAI}}}. \bibinfo{publisher}{{AAAI} Press},
  \bibinfo{pages}{1004--1011}.
\newblock


\bibitem[\protect\citeauthoryear{Liang, Ke, Zhang, Yi, and Zheng}{Liang
  et~al\mbox{.}}{2018}]%
        {DBLP:conf/ijcai/LiangKZYZ18}
\bibfield{author}{\bibinfo{person}{Yuxuan Liang}, \bibinfo{person}{Songyu Ke},
  \bibinfo{person}{Junbo Zhang}, \bibinfo{person}{Xiuwen Yi}, {and}
  \bibinfo{person}{Yu Zheng}.} \bibinfo{year}{2018}\natexlab{}.
\newblock \showarticletitle{GeoMAN: Multi-level Attention Networks for
  Geo-sensory Time Series Prediction}. In \bibinfo{booktitle}{\emph{{IJCAI}}}.
  \bibinfo{publisher}{ijcai.org}, \bibinfo{pages}{3428--3434}.
\newblock


\bibitem[\protect\citeauthoryear{Lundberg and Lee}{Lundberg and Lee}{2017}]%
        {DBLP:conf/nips/LundbergL17}
\bibfield{author}{\bibinfo{person}{Scott~M. Lundberg} {and}
  \bibinfo{person}{Su{-}In Lee}.} \bibinfo{year}{2017}\natexlab{}.
\newblock \showarticletitle{A Unified Approach to Interpreting Model
  Predictions}. In \bibinfo{booktitle}{\emph{{NeurIPS}}}.
  \bibinfo{pages}{4765--4774}.
\newblock


\bibitem[\protect\citeauthoryear{Luong, Pham, and Manning}{Luong
  et~al\mbox{.}}{2015}]%
        {DBLP:conf/emnlp/LuongPM15}
\bibfield{author}{\bibinfo{person}{Thang Luong}, \bibinfo{person}{Hieu Pham},
  {and} \bibinfo{person}{Christopher~D. Manning}.}
  \bibinfo{year}{2015}\natexlab{}.
\newblock \showarticletitle{Effective Approaches to Attention-based Neural
  Machine Translation}. In \bibinfo{booktitle}{\emph{{EMNLP}}}.
  \bibinfo{publisher}{ACL}, \bibinfo{pages}{1412--1421}.
\newblock


\bibitem[\protect\citeauthoryear{Ma, Chitta, Zhou, You, Sun, and Gao}{Ma
  et~al\mbox{.}}{2017}]%
        {DBLP:conf/kdd/MaCZYSG17}
\bibfield{author}{\bibinfo{person}{Fenglong Ma}, \bibinfo{person}{Radha
  Chitta}, \bibinfo{person}{Jing Zhou}, \bibinfo{person}{Quanzeng You},
  \bibinfo{person}{Tong Sun}, {and} \bibinfo{person}{Jing Gao}.}
  \bibinfo{year}{2017}\natexlab{}.
\newblock \showarticletitle{Dipole: Diagnosis Prediction in Healthcare via
  Attention-based Bidirectional Recurrent Neural Networks}. In
  \bibinfo{booktitle}{\emph{{SIGKDD}}}. \bibinfo{publisher}{{ACM}},
  \bibinfo{pages}{1903--1911}.
\newblock


\bibitem[\protect\citeauthoryear{Mnih, Heess, Graves, and Kavukcuoglu}{Mnih
  et~al\mbox{.}}{2014}]%
        {DBLP:conf/nips/MnihHGK14}
\bibfield{author}{\bibinfo{person}{Volodymyr Mnih}, \bibinfo{person}{Nicolas
  Heess}, \bibinfo{person}{Alex Graves}, {and} \bibinfo{person}{Koray
  Kavukcuoglu}.} \bibinfo{year}{2014}\natexlab{}.
\newblock \showarticletitle{Recurrent Models of Visual Attention}. In
  \bibinfo{booktitle}{\emph{{NeurIPS}}}. \bibinfo{pages}{2204--2212}.
\newblock


\bibitem[\protect\citeauthoryear{Mohler, Short, Brantingham, Schoenberg, and
  Tita}{Mohler et~al\mbox{.}}{2011}]%
        {doi:10.1198/jasa.2011.ap09546}
\bibfield{author}{\bibinfo{person}{G.~O. Mohler}, \bibinfo{person}{M.~B.
  Short}, \bibinfo{person}{P.~J. Brantingham}, \bibinfo{person}{F.~P.
  Schoenberg}, {and} \bibinfo{person}{G.~E. Tita}.}
  \bibinfo{year}{2011}\natexlab{}.
\newblock \showarticletitle{Self-Exciting Point Process Modeling of Crime}.
\newblock \bibinfo{journal}{\emph{J. Amer. Statist. Assoc.}}
  \bibinfo{volume}{106}, \bibinfo{number}{493} (\bibinfo{year}{2011}),
  \bibinfo{pages}{100--108}.
\newblock


\bibitem[\protect\citeauthoryear{Mullenbach, Wiegreffe, Duke, Sun, and
  Eisenstein}{Mullenbach et~al\mbox{.}}{2018}]%
        {DBLP:conf/naacl/MullenbachWDSE18}
\bibfield{author}{\bibinfo{person}{James Mullenbach}, \bibinfo{person}{Sarah
  Wiegreffe}, \bibinfo{person}{Jon Duke}, \bibinfo{person}{Jimeng Sun}, {and}
  \bibinfo{person}{Jacob Eisenstein}.} \bibinfo{year}{2018}\natexlab{}.
\newblock \showarticletitle{Explainable Prediction of Medical Codes from
  Clinical Text}. In \bibinfo{booktitle}{\emph{{NAACL-HLT}}}.
  \bibinfo{publisher}{Association for Computational Linguistics},
  \bibinfo{pages}{1101--1111}.
\newblock


\bibitem[\protect\citeauthoryear{Nakaya and Yano}{Nakaya and Yano}{2010}]%
        {DBLP:journals/tgis/NakayaY10}
\bibfield{author}{\bibinfo{person}{Tomoki Nakaya} {and} \bibinfo{person}{Keiji
  Yano}.} \bibinfo{year}{2010}\natexlab{}.
\newblock \showarticletitle{Visualising Crime Clusters in a Space-time Cube: An
  Exploratory Data-analysis Approach Using Space-time Kernel Density Estimation
  and Scan Statistics}.
\newblock \bibinfo{journal}{\emph{Trans. {GIS}}} \bibinfo{volume}{14},
  \bibinfo{number}{3} (\bibinfo{year}{2010}), \bibinfo{pages}{223--239}.
\newblock


\bibitem[\protect\citeauthoryear{Ribeiro, Singh, and Guestrin}{Ribeiro
  et~al\mbox{.}}{2016}]%
        {DBLP:conf/kdd/Ribeiro0G16}
\bibfield{author}{\bibinfo{person}{Marco~T{\'{u}}lio Ribeiro},
  \bibinfo{person}{Sameer Singh}, {and} \bibinfo{person}{Carlos Guestrin}.}
  \bibinfo{year}{2016}\natexlab{}.
\newblock \showarticletitle{"Why Should {I} Trust You?": Explaining the
  Predictions of Any Classifier}. In \bibinfo{booktitle}{\emph{{SIGKDD}}}.
  \bibinfo{publisher}{{ACM}}, \bibinfo{pages}{1135--1144}.
\newblock


\bibitem[\protect\citeauthoryear{Rong, Xu, Yan, and Ma}{Rong
  et~al\mbox{.}}{2018}]%
        {DBLP:conf/kdd/RongXYM18}
\bibfield{author}{\bibinfo{person}{Yuecheng Rong}, \bibinfo{person}{Zhimian
  Xu}, \bibinfo{person}{Ruibo Yan}, {and} \bibinfo{person}{Xu Ma}.}
  \bibinfo{year}{2018}\natexlab{}.
\newblock \showarticletitle{Du-Parking: Spatio-Temporal Big Data Tells You
  Realtime Parking Availability}. In \bibinfo{booktitle}{\emph{{SIGKDD}}}.
  \bibinfo{publisher}{{ACM}}, \bibinfo{pages}{646--654}.
\newblock


\bibitem[\protect\citeauthoryear{Rudin}{Rudin}{2019}]%
        {10.1038/s42256-019-0048-x}
\bibfield{author}{\bibinfo{person}{Cynthia Rudin}.}
  \bibinfo{year}{2019}\natexlab{}.
\newblock \showarticletitle{{Stop explaining black box machine learning models
  for high stakes decisions and use interpretable models instead}}.
\newblock \bibinfo{journal}{\emph{Nature Machine Intelligence}}
  \bibinfo{volume}{1}, \bibinfo{number}{5} (\bibinfo{year}{2019}),
  \bibinfo{pages}{206--215}.
\newblock


\bibitem[\protect\citeauthoryear{Rumi, Luong, and Salim}{Rumi
  et~al\mbox{.}}{2019}]%
        {DBLP:journals/corr/abs-1908-02570}
\bibfield{author}{\bibinfo{person}{Shakila~Khan Rumi}, \bibinfo{person}{Phillip
  Luong}, {and} \bibinfo{person}{Flora~D. Salim}.}
  \bibinfo{year}{2019}\natexlab{}.
\newblock \showarticletitle{Crime Rate Prediction with Region Risk and Movement
  Patterns}.
\newblock \bibinfo{journal}{\emph{CoRR}}  \bibinfo{volume}{abs/1908.02570}
  (\bibinfo{year}{2019}).
\newblock


\bibitem[\protect\citeauthoryear{Serrano and Smith}{Serrano and Smith}{2019}]%
        {DBLP:conf/acl/SerranoS19}
\bibfield{author}{\bibinfo{person}{Sofia Serrano} {and}
  \bibinfo{person}{Noah~A. Smith}.} \bibinfo{year}{2019}\natexlab{}.
\newblock \showarticletitle{Is Attention Interpretable?}. In
  \bibinfo{booktitle}{\emph{{ACL}}}. \bibinfo{publisher}{Association for
  Computational Linguistics}, \bibinfo{pages}{2931--2951}.
\newblock


\bibitem[\protect\citeauthoryear{Su, Wei, Varshney, and Malioutov}{Su
  et~al\mbox{.}}{2016}]%
        {DBLP:journals/corr/SuWVM16}
\bibfield{author}{\bibinfo{person}{Guolong Su}, \bibinfo{person}{Dennis Wei},
  \bibinfo{person}{Kush~R. Varshney}, {and} \bibinfo{person}{Dmitry~M.
  Malioutov}.} \bibinfo{year}{2016}\natexlab{}.
\newblock \showarticletitle{Interpretable Two-level Boolean Rule Learning for
  Classification}.
\newblock \bibinfo{journal}{\emph{CoRR}}  \bibinfo{volume}{abs/1606.05798}
  (\bibinfo{year}{2016}).
\newblock


\bibitem[\protect\citeauthoryear{{Sun}, {Zhang}, {Li}, {Yi}, {Liang}, and
  {Zheng}}{{Sun} et~al\mbox{.}}{2020}]%
        {DBLP:journals/TKDE/9139357}
\bibfield{author}{\bibinfo{person}{J. {Sun}}, \bibinfo{person}{J. {Zhang}},
  \bibinfo{person}{Q. {Li}}, \bibinfo{person}{X. {Yi}}, \bibinfo{person}{Y.
  {Liang}}, {and} \bibinfo{person}{Y. {Zheng}}.}
  \bibinfo{year}{2020}\natexlab{}.
\newblock \showarticletitle{Predicting Citywide Crowd Flows in Irregular
  Regions Using Multi-View Graph Convolutional Networks}.
\newblock \bibinfo{journal}{\emph{IEEE Transactions on Knowledge and Data
  Engineering}} (\bibinfo{year}{2020}), \bibinfo{pages}{1--1}.
\newblock


\bibitem[\protect\citeauthoryear{Toole, Eagle, and Plotkin}{Toole
  et~al\mbox{.}}{2011}]%
        {DBLP:journals/tist/TooleEP11}
\bibfield{author}{\bibinfo{person}{Jameson~L. Toole}, \bibinfo{person}{Nathan
  Eagle}, {and} \bibinfo{person}{Joshua~B. Plotkin}.}
  \bibinfo{year}{2011}\natexlab{}.
\newblock \showarticletitle{Spatiotemporal correlations in criminal offense
  records}.
\newblock \bibinfo{journal}{\emph{{TIST}}} \bibinfo{volume}{2},
  \bibinfo{number}{4} (\bibinfo{year}{2011}), \bibinfo{pages}{38:1--38:18}.
\newblock


\bibitem[\protect\citeauthoryear{Trips}{Trips}{2019}]%
        {chicago_taxi2019}
\bibfield{author}{\bibinfo{person}{Chicago~Taxi Trips}.}
  \bibinfo{year}{2019}\natexlab{}.
\newblock \bibinfo{title}{City of Chicago Data Portal}.
\newblock
  \bibinfo{howpublished}{\url{https://data.cityofchicago.org/Transportation/Taxi-Trips-2019/h4cq-z3dy}}.
\newblock


\bibitem[\protect\citeauthoryear{Vaswani, Shazeer, Parmar, Uszkoreit, Jones,
  Gomez, Kaiser, and Polosukhin}{Vaswani et~al\mbox{.}}{2017}]%
        {DBLP:conf/nips/VaswaniSPUJGKP17}
\bibfield{author}{\bibinfo{person}{Ashish Vaswani}, \bibinfo{person}{Noam
  Shazeer}, \bibinfo{person}{Niki Parmar}, \bibinfo{person}{Jakob Uszkoreit},
  \bibinfo{person}{Llion Jones}, \bibinfo{person}{Aidan~N. Gomez},
  \bibinfo{person}{Lukasz Kaiser}, {and} \bibinfo{person}{Illia Polosukhin}.}
  \bibinfo{year}{2017}\natexlab{}.
\newblock \showarticletitle{Attention is All you Need}. In
  \bibinfo{booktitle}{\emph{{NeurIPS}}}. \bibinfo{pages}{5998--6008}.
\newblock


\bibitem[\protect\citeauthoryear{Velickovic, Cucurull, Casanova, Romero,
  Li{\`{o}}, and Bengio}{Velickovic et~al\mbox{.}}{2018}]%
        {DBLP:conf/iclr/VelickovicCCRLB18}
\bibfield{author}{\bibinfo{person}{Petar Velickovic}, \bibinfo{person}{Guillem
  Cucurull}, \bibinfo{person}{Arantxa Casanova}, \bibinfo{person}{Adriana
  Romero}, \bibinfo{person}{Pietro Li{\`{o}}}, {and} \bibinfo{person}{Yoshua
  Bengio}.} \bibinfo{year}{2018}\natexlab{}.
\newblock \showarticletitle{Graph Attention Networks}. In
  \bibinfo{booktitle}{\emph{{ICLR}}}. \bibinfo{publisher}{OpenReview.net}.
\newblock


\bibitem[\protect\citeauthoryear{Wang, Zhang, Zhang, Brantingham, and
  Bertozzi}{Wang et~al\mbox{.}}{2019c}]%
        {DBLP:journals/corr/WangZZBB17}
\bibfield{author}{\bibinfo{person}{Bao Wang}, \bibinfo{person}{Duo Zhang},
  \bibinfo{person}{Duanhao Zhang}, \bibinfo{person}{P.~Jeffery Brantingham},
  {and} \bibinfo{person}{Andrea~L. Bertozzi}.}
  \bibinfo{year}{2019}\natexlab{c}.
\newblock \showarticletitle{Deep Learning for Real-Time Crime Forecasting and
  Its Ternarization}.
\newblock \bibinfo{journal}{\emph{Chinese Annals of Mathematics, Series B}}
  \bibinfo{volume}{40} (\bibinfo{year}{2019}), \bibinfo{pages}{949–966}.
\newblock


\bibitem[\protect\citeauthoryear{Wang, Jenkins, Wei, Wu, and Li}{Wang
  et~al\mbox{.}}{2019a}]%
        {DBLP:conf/www/WangJWWL19}
\bibfield{author}{\bibinfo{person}{Hongjian Wang}, \bibinfo{person}{Porter
  Jenkins}, \bibinfo{person}{Hua Wei}, \bibinfo{person}{Fei Wu}, {and}
  \bibinfo{person}{Zhenhui Li}.} \bibinfo{year}{2019}\natexlab{a}.
\newblock \showarticletitle{Learning Task-Specific City Region Partition}. In
  \bibinfo{booktitle}{\emph{{WWW}}}. \bibinfo{publisher}{{ACM}},
  \bibinfo{pages}{3300--3306}.
\newblock


\bibitem[\protect\citeauthoryear{Wang, Kifer, Graif, and Li}{Wang
  et~al\mbox{.}}{2016}]%
        {DBLP:conf/kdd/WangKGL16}
\bibfield{author}{\bibinfo{person}{Hongjian Wang}, \bibinfo{person}{Daniel
  Kifer}, \bibinfo{person}{Corina Graif}, {and} \bibinfo{person}{Zhenhui Li}.}
  \bibinfo{year}{2016}\natexlab{}.
\newblock \showarticletitle{Crime Rate Inference with Big Data}. In
  \bibinfo{booktitle}{\emph{{SIGKDD}}}. \bibinfo{publisher}{{ACM}},
  \bibinfo{pages}{635--644}.
\newblock


\bibitem[\protect\citeauthoryear{Wang, Yao, Kifer, Graif, and Li}{Wang
  et~al\mbox{.}}{2019b}]%
        {DBLP:journals/tbd/WangYKGL19}
\bibfield{author}{\bibinfo{person}{Hongjian Wang}, \bibinfo{person}{Huaxiu
  Yao}, \bibinfo{person}{Daniel Kifer}, \bibinfo{person}{Corina Graif}, {and}
  \bibinfo{person}{Zhenhui Li}.} \bibinfo{year}{2019}\natexlab{b}.
\newblock \showarticletitle{Non-Stationary Model for Crime Rate Inference Using
  Modern Urban Data}.
\newblock \bibinfo{journal}{\emph{{IEEE} Trans. Big Data}} \bibinfo{volume}{5},
  \bibinfo{number}{2} (\bibinfo{year}{2019}), \bibinfo{pages}{180--194}.
\newblock


\bibitem[\protect\citeauthoryear{Wang, Ma, Wang, Jin, Wang, Tang, Jia, and
  Yu}{Wang et~al\mbox{.}}{2020}]%
        {DBLP:conf/www/Wang0WJWTJY20}
\bibfield{author}{\bibinfo{person}{Xiaoyang Wang}, \bibinfo{person}{Yao Ma},
  \bibinfo{person}{Yiqi Wang}, \bibinfo{person}{Wei Jin}, \bibinfo{person}{Xin
  Wang}, \bibinfo{person}{Jiliang Tang}, \bibinfo{person}{Caiyan Jia}, {and}
  \bibinfo{person}{Jian Yu}.} \bibinfo{year}{2020}\natexlab{}.
\newblock \showarticletitle{Traffic Flow Prediction via Spatial Temporal Graph
  Neural Network}. In \bibinfo{booktitle}{\emph{{WWW}}}.
  \bibinfo{publisher}{{ACM}}, \bibinfo{pages}{1082--1092}.
\newblock


\bibitem[\protect\citeauthoryear{Wiegreffe and Pinter}{Wiegreffe and
  Pinter}{2019}]%
        {DBLP:conf/emnlp/WiegreffeP19}
\bibfield{author}{\bibinfo{person}{Sarah Wiegreffe} {and}
  \bibinfo{person}{Yuval Pinter}.} \bibinfo{year}{2019}\natexlab{}.
\newblock \showarticletitle{Attention is not not Explanation}. In
  \bibinfo{booktitle}{\emph{{EMNLP-IJCNLP}}}. \bibinfo{publisher}{Association
  for Computational Linguistics}, \bibinfo{pages}{11--20}.
\newblock


\bibitem[\protect\citeauthoryear{Xie, Guo, Chen, Xiao, Wang, and Zhao}{Xie
  et~al\mbox{.}}{2020}]%
        {DBLP:conf/cikm/XieG0X0Z20}
\bibfield{author}{\bibinfo{person}{Qinge Xie}, \bibinfo{person}{Tiancheng Guo},
  \bibinfo{person}{Yang Chen}, \bibinfo{person}{Yu Xiao}, \bibinfo{person}{Xin
  Wang}, {and} \bibinfo{person}{Ben~Y. Zhao}.} \bibinfo{year}{2020}\natexlab{}.
\newblock \showarticletitle{Deep Graph Convolutional Networks for
  Incident-Driven Traffic Speed Prediction}. In
  \bibinfo{booktitle}{\emph{{CIKM}}}. \bibinfo{publisher}{{ACM}},
  \bibinfo{pages}{1665--1674}.
\newblock


\bibitem[\protect\citeauthoryear{Xiong, Srivastava, Kannan, Damle, Prasanna,
  and Southers}{Xiong et~al\mbox{.}}{2019}]%
        {DBLP:conf/gis/XiongSKDPS19}
\bibfield{author}{\bibinfo{person}{Chuanxiu Xiong}, \bibinfo{person}{Ajitesh
  Srivastava}, \bibinfo{person}{Rajgopal Kannan}, \bibinfo{person}{Omkar
  Damle}, \bibinfo{person}{Viktor~K. Prasanna}, {and} \bibinfo{person}{Erroll
  Southers}.} \bibinfo{year}{2019}\natexlab{}.
\newblock \showarticletitle{On Predicting Crime with Heterogeneous Spatial
  Patterns: Methods and Evaluation}. In
  \bibinfo{booktitle}{\emph{{SIGSPATIAL}}}. \bibinfo{publisher}{{ACM}},
  \bibinfo{pages}{43--51}.
\newblock


\bibitem[\protect\citeauthoryear{Xu, Ba, Kiros, Cho, Courville, Salakhutdinov,
  Zemel, and Bengio}{Xu et~al\mbox{.}}{2015}]%
        {DBLP:conf/icml/XuBKCCSZB15}
\bibfield{author}{\bibinfo{person}{Kelvin Xu}, \bibinfo{person}{Jimmy Ba},
  \bibinfo{person}{Ryan Kiros}, \bibinfo{person}{Kyunghyun Cho},
  \bibinfo{person}{Aaron~C. Courville}, \bibinfo{person}{Ruslan Salakhutdinov},
  \bibinfo{person}{Richard~S. Zemel}, {and} \bibinfo{person}{Yoshua Bengio}.}
  \bibinfo{year}{2015}\natexlab{}.
\newblock \showarticletitle{Show, Attend and Tell: Neural Image Caption
  Generation with Visual Attention}. In \bibinfo{booktitle}{\emph{{ICML}}}.
  \bibinfo{publisher}{JMLR.org}, \bibinfo{pages}{2048--2057}.
\newblock


\bibitem[\protect\citeauthoryear{Yao, Tang, Wei, Zheng, and Li}{Yao
  et~al\mbox{.}}{2019}]%
        {DBLP:conf/aaai/YaoTWZL19}
\bibfield{author}{\bibinfo{person}{Huaxiu Yao}, \bibinfo{person}{Xianfeng
  Tang}, \bibinfo{person}{Hua Wei}, \bibinfo{person}{Guanjie Zheng}, {and}
  \bibinfo{person}{Zhenhui Li}.} \bibinfo{year}{2019}\natexlab{}.
\newblock \showarticletitle{Revisiting Spatial-Temporal Similarity: {A} Deep
  Learning Framework for Traffic Prediction}. In
  \bibinfo{booktitle}{\emph{{AAAI}}}. \bibinfo{publisher}{{AAAI} Press},
  \bibinfo{pages}{5668--5675}.
\newblock


\bibitem[\protect\citeauthoryear{Yao, Wu, Ke, Tang, Jia, Lu, Gong, Ye, and
  Li}{Yao et~al\mbox{.}}{2018}]%
        {DBLP:conf/aaai/Yao0KTJLGYL18}
\bibfield{author}{\bibinfo{person}{Huaxiu Yao}, \bibinfo{person}{Fei Wu},
  \bibinfo{person}{Jintao Ke}, \bibinfo{person}{Xianfeng Tang},
  \bibinfo{person}{Yitian Jia}, \bibinfo{person}{Siyu Lu},
  \bibinfo{person}{Pinghua Gong}, \bibinfo{person}{Jieping Ye}, {and}
  \bibinfo{person}{Zhenhui Li}.} \bibinfo{year}{2018}\natexlab{}.
\newblock \showarticletitle{Deep Multi-View Spatial-Temporal Network for Taxi
  Demand Prediction}. In \bibinfo{booktitle}{\emph{{AAAI}}}.
  \bibinfo{publisher}{{AAAI} Press}, \bibinfo{pages}{2588--2595}.
\newblock


\bibitem[\protect\citeauthoryear{Yu, Yin, and Zhu}{Yu et~al\mbox{.}}{2018}]%
        {DBLP:conf/ijcai/YuYZ18}
\bibfield{author}{\bibinfo{person}{Bing Yu}, \bibinfo{person}{Haoteng Yin},
  {and} \bibinfo{person}{Zhanxing Zhu}.} \bibinfo{year}{2018}\natexlab{}.
\newblock \showarticletitle{Spatio-Temporal Graph Convolutional Networks: {A}
  Deep Learning Framework for Traffic Forecasting}. In
  \bibinfo{booktitle}{\emph{{IJCAI}}}. \bibinfo{publisher}{ijcai.org},
  \bibinfo{pages}{3634--3640}.
\newblock


\bibitem[\protect\citeauthoryear{Yu, Ding, Chen, and Morabito}{Yu
  et~al\mbox{.}}{2014}]%
        {DBLP:conf/pakdd/Yu0CM14}
\bibfield{author}{\bibinfo{person}{Chung{-}Hsien Yu}, \bibinfo{person}{Wei
  Ding}, \bibinfo{person}{Ping Chen}, {and} \bibinfo{person}{Melissa
  Morabito}.} \bibinfo{year}{2014}\natexlab{}.
\newblock \showarticletitle{Crime Forecasting Using Spatio-temporal Pattern
  with Ensemble Learning}. In \bibinfo{booktitle}{\emph{{PAKDD}}}.
  \bibinfo{publisher}{Springer}, \bibinfo{pages}{174--185}.
\newblock


\bibitem[\protect\citeauthoryear{Zhang, Zheng, and Qi}{Zhang
  et~al\mbox{.}}{2017}]%
        {DBLP:conf/aaai/ZhangZQ17}
\bibfield{author}{\bibinfo{person}{Junbo Zhang}, \bibinfo{person}{Yu Zheng},
  {and} \bibinfo{person}{Dekang Qi}.} \bibinfo{year}{2017}\natexlab{}.
\newblock \showarticletitle{Deep Spatio-Temporal Residual Networks for Citywide
  Crowd Flows Prediction}. In \bibinfo{booktitle}{\emph{{AAAI}}}.
  \bibinfo{publisher}{{AAAI} Press}, \bibinfo{pages}{1655--1661}.
\newblock


\bibitem[\protect\citeauthoryear{Zhang, Zheng, Qi, Li, and Yi}{Zhang
  et~al\mbox{.}}{2016}]%
        {DBLP:conf/gis/ZhangZQLY16}
\bibfield{author}{\bibinfo{person}{Junbo Zhang}, \bibinfo{person}{Yu Zheng},
  \bibinfo{person}{Dekang Qi}, \bibinfo{person}{Ruiyuan Li}, {and}
  \bibinfo{person}{Xiuwen Yi}.} \bibinfo{year}{2016}\natexlab{}.
\newblock \showarticletitle{DNN-based prediction model for spatio-temporal
  data}. In \bibinfo{booktitle}{\emph{{SIGSPATIAL}}}.
  \bibinfo{publisher}{{ACM}}, \bibinfo{pages}{92:1--92:4}.
\newblock


\bibitem[\protect\citeauthoryear{Zhang, Huang, Xu, and Xia}{Zhang
  et~al\mbox{.}}{2020}]%
        {DBLP:conf/cikm/ZhangHXX20}
\bibfield{author}{\bibinfo{person}{Xiyue Zhang}, \bibinfo{person}{Chao Huang},
  \bibinfo{person}{Yong Xu}, {and} \bibinfo{person}{Lianghao Xia}.}
  \bibinfo{year}{2020}\natexlab{}.
\newblock \showarticletitle{Spatial-Temporal Convolutional Graph Attention
  Networks for Citywide Traffic Flow Forecasting}. In
  \bibinfo{booktitle}{\emph{{CIKM}}}. \bibinfo{publisher}{{ACM}},
  \bibinfo{pages}{1853--1862}.
\newblock


\bibitem[\protect\citeauthoryear{Zhao and Tang}{Zhao and Tang}{2017}]%
        {DBLP:conf/cikm/ZhaoT17}
\bibfield{author}{\bibinfo{person}{Xiangyu Zhao} {and} \bibinfo{person}{Jiliang
  Tang}.} \bibinfo{year}{2017}\natexlab{}.
\newblock \showarticletitle{Modeling Temporal-Spatial Correlations for Crime
  Prediction}. In \bibinfo{booktitle}{\emph{{CIKM}}}.
  \bibinfo{publisher}{{ACM}}, \bibinfo{pages}{497--506}.
\newblock


\end{thebibliography}
